\documentclass{article}

\usepackage[accepted]{icml2021}

\usepackage[utf8]{inputenc} 
\usepackage[T1]{fontenc}    
\usepackage[colorlinks=true,linkcolor=red]{hyperref}
\usepackage{xcolor}
\usepackage{url}            
\usepackage{booktabs}       
\usepackage{nicefrac}       
\usepackage{microtype} 
\usepackage{color}
\usepackage{soul}

\usepackage{subfig}     

\usepackage[auth-lg]{authblk}
\usepackage{tabularx}
\usepackage{graphicx}
\usepackage[normalem]{ulem}
\usepackage{xspace}
\usepackage{multirow}
\usepackage{float}
\usepackage{amsmath,amsthm,amssymb,bm}
\usepackage{wrapfig}
\usepackage{multicol}
\usepackage{mwe}
\usepackage[inline, shortlabels]{enumitem}
\setlist[enumerate]{nosep}
\usepackage{cleveref}
\usepackage{dsfont}
\usepackage{array}
\usepackage[hang,flushmargin]{footmisc}
\usepackage{rotating}       
\usepackage{verbatim}

\DeclareGraphicsExtensions{.pdf, .PDF, .png, .PNG, .jpg, .JPG, .jpeg, .JPEG}
\newcommand*{\proposed}{{\tt JTLA}\@\xspace}
\newcommand*{\dknn}{{\tt DKNN}\@\xspace}
\newcommand*{\trust}{{\tt Trust}\@\xspace}
\newcommand*{\odds}{{\tt Odds}\@\xspace}
\newcommand*{\dm}{{\tt Mahalanobis}\@\xspace}
\newcommand*{\lid}{{\tt LID}\@\xspace}

\newcommand{\BigO}[1]{\ensuremath{\operatorname{O}\left(#1\right)}}

\newcommand{\argmax}{\operatornamewithlimits{arg\,max}}
\newcommand{\ben}{\begin{enumerate}}
\newcommand{\een}{\end{enumerate}}
\newcommand{\beq}{\begin{equation}}
\newcommand{\eeq}{\end{equation}}
\newcommand{\beqa}{\begin{eqnarray}}
\newcommand{\eeqa}{\end{eqnarray}}
\newcommand{\bit}{\begin{itemize}}
\newcommand{\eit}{\end{itemize}}
\newcommand{\btab}{\begin{tabular}}
\newcommand{\etab}{\end{tabular}}

\newcommand{\ceil}[1]{\left\lceil #1 \right\rceil}

\newcommand{\cond}{{\,|\,}}

\newcommand{\noprint}[1]{}

\newcommand\NoDo{\renewcommand\algorithmicdo{}}
\newcommand\OptDo{\renewcommand\algorithmicdo{\textbf{[optional]}}}
\newcommand\NoThen{\renewcommand\algorithmicthen{}}

\def \ie {{\em i.e.},~}

\def \eg {{\em e.g.},~}

\def \reals {\mathds{R}}

\def \proba {\mathds{P}}
\def \indicator {\mathds{1}}
\def \mysum {\displaystyle\sum\limits}
\def \myprod {\displaystyle\prod\limits}

\def \bfdelta {\bm{\delta}}

\def \bft {\mathbf{t}}

\def \bfw {\mathbf{w}}
\def \bfx {\mathbf{x}}
\def \bfy {\mathbf{y}}

\icmltitlerunning{A General Framework For Detecting Anomalous Inputs to DNN Classifiers}

\begin{document}


\twocolumn[
\icmltitle{A General Framework For Detecting Anomalous Inputs to DNN Classifiers} 



\icmlsetsymbol{equal}{*}

\begin{icmlauthorlist}
\icmlauthor{Jayaram Raghuram}{equal,to}
\icmlauthor{Varun Chandrasekaran}{equal,to}
\icmlauthor{Somesh Jha}{to,too}
\icmlauthor{Suman Banerjee}{to}
\end{icmlauthorlist}

\icmlaffiliation{to}{Computer Sciences, University of Wisconsin, Madison, USA.}
\icmlaffiliation{too}{XaiPient Inc., Princeton, NJ, USA}

\icmlcorrespondingauthor{Jayaram Raghuram}{\texttt{jayaramr@cs.wisc.edu}}

\icmlkeywords{adversarial detection, OOD detection, adversarial attacks, DNN layer representations, anomaly detection, p-values}

\vskip 0.3in
]



\printAffiliationsAndNotice{\icmlEqualContribution} 


\begin{abstract}
Detecting anomalous inputs, such as adversarial and out-of-distribution (OOD) inputs, is critical for classifiers (including deep neural networks or DNNs) deployed in real-world applications. 
While prior works have proposed various methods to detect such anomalous samples using information from the internal layer representations of a DNN, there is a lack of consensus on a principled approach for the different components of such a detection method. 
As a result, often heuristic and one-off methods are applied for different aspects of this problem. 
We propose an unsupervised anomaly detection framework based on the internal DNN layer representations in the form of a meta-algorithm with configurable components. 
We proceed to propose specific instantiations for each component of the meta-algorithm based on ideas grounded in statistical testing and anomaly detection. 
We evaluate the proposed methods on well-known image classification datasets with strong adversarial attacks and OOD inputs, including an {\em adaptive attack} that uses the internal layer representations of the DNN (often not considered in prior work). 
Comparisons with five recently-proposed competing detection methods demonstrates the effectiveness of our method in detecting adversarial and OOD inputs.

\end{abstract}

\section{Introduction}
\label{sec:intro}
Deep neural networks (DNNs) have achieved impressive performance on a variety of challenging machine learning (ML) problems such as image classification, object detection, speech recognition, and natural language processing~\cite{he2015delving, krizhevsky2017imagenet}. However, it is well-known that DNN classifiers can be highly inaccurate (sometimes with high confidence) on test inputs from outside the training distribution~\cite{nguyen2015deep, szegedy2013intriguing, hendrycks2017baseline, hein2019relu}. Such anomalous inputs can arise in real-world settings either unintentionally due to external factors, or due to malicious adversaries that intend to cause prediction errors in the DNN and disrupt the system~\cite{barreno2006can, biggio2018wild}. Therefore, it is critical to have a defense mechanism that can detect such anomalous inputs, and take suitable corrective action (\eg abstain from predicting~\cite{tax2008growing} or provide a more reliable class prediction). 

In this work, we propose \proposed (Joint statistical Testing across DNN Layers for Anomalies), a general unsupervised framework for detecting anomalous inputs (including adversarial and OOD) to a DNN classifier using its layer representations.
%
%
\proposed utilizes the rich information at the intermediate layer representations of a DNN to obtain a better understanding of the patterns produced by anomalous inputs for detection. Our method is {\em unsupervised}, \ie it does not utilize any specific class(es) of known anomalous samples for learning or tuning its parameters. 
While a number of prior works have addressed the problem of adversarial and OOD detection~\cite{feinman2017detecting, xu2017feature1, li2017adversarial, lee2018simple, ma2018characterizing_iclr, roth2019odds}, including ones that utilize intermediate layer representations of a DNN~\cite{li2017adversarial, meng2017magnet, xu2017feature1, lee2018simple, zheng2018robust, ma2018characterizing_iclr, papernot2018deep, miller2019ada, yang2019ml, sastry2019gram}, some key limitations persist, that we propose to address in this work.

\noindent{\bf Limitations of prior work.}
First, a number of existing detection methods~\cite{feinman2017detecting, lee2018simple, ma2018characterizing_iclr, yang2019ml} being supervised, have to be presented with a broad sampling of known anomalous samples for training (\eg different adversarial attacks of varying strength). Such methods typically also need to configure hyper-parameters based on the known anomalous samples from the training set (\eg using cross-validation). As a result, they often do not generalize well to {\em unknown} anomalies (\eg novel or adaptive attacks). It has been shown that a majority of the current detection methods fail to handle unseen and adaptive adversaries that are aware of the defense mechanism~\cite{carlini2017bypassing, tramer2020adaptive}.
{\bf Second}, detection methods that use only the input, output (pre-softmax), or a specific DNN layer~\cite{roth2019odds, hendrycks2017baseline, feinman2017detecting} do not jointly exploit the properties exhibited by anomalous inputs across the layers. 
%
%
{\bf Third}, although methods that utilize information from multiple layers (listed earlier)
propose specific test statistics or features calculated from the layer representations (\eg local intrinsic dimensionality~\cite{ma2018characterizing_iclr}), there is a lack of a {\em general anomaly detection framework} where one can plug-in test statistics, aggregation, and scoring methods suitable for the detection task.  
{\bf Fourth}, existing unsupervised detection methods that are based on density (generative) modeling of the DNN layer representations~\cite{zheng2018robust, miller2019ada, feinman2017detecting} are not well-suited to handle the (often very) high dimensional layer representations.
{\bf Finally}, we observe that existing detection methods often do not utilize the predicted class of the DNN to focus on class-conditional properties of the layer representations~\cite{ma2018characterizing_iclr, li2017adversarial, xu2017feature1, yang2019ml}, which can lead to improved detection performance. 
While prior works such as \cite{roth2019odds, miller2019ada, zheng2018robust, sastry2019gram} are exceptions to this, there is still need for a unified approach in this regard.

Our contributions can be summarized as follows:
\begin{itemize}[leftmargin=*, topsep=1pt, noitemsep]
    \item We propose a general unsupervised framework \proposed for detecting anomalous inputs to a DNN using its layer representations. We first present a meta-algorithm and describe its components in general terms (\S~\ref{sec:proposed}). We then propose specific methods for realizing the components in a principled way (\S~\ref{sec:instance_meta_algo}). The proposed framework is modular, and a number of prior works for anomaly detection based on the layer representations can be cast into this meta-algorithm.
    %
    %
    %
    \item The importance of designing an adaptive, defense-aware adversary has been underscored in the literature~\cite{carlini2017bypassing, tramer2020adaptive}. We propose and evaluate against an adversarial attack that focuses on defenses (such as ours) that use the $k$-nearest neighbors of the layer representations of the DNN (\S~\ref{sec:dknn_attack}). 
    \item We report extensive experimental evaluations comparing \proposed with five baseline methods on three image classification datasets trained with suitably-complex DNN architectures. For adversarial detection, we evaluate on three well-known whitebox attacks and our proposed defense-aware attack (\S~\ref{sec_exp} and Appendix~\ref{app:additional})~\footnote{The code base associated with our work can be found at:\\ \url{https://github.com/jayaram-r/adversarial-detection}.}.
\end{itemize}

\section{Related Works}
\label{sec:related}
We provide a brief review of related works on adversarial and OOD detection, focusing on methods that use the internal layer representations of a DNN. A detailed discussion of closely-related prior works, and how they fit into the proposed anomaly detection framework is provided in Appendix~\ref{sec:app_related_works}.
Recent surveys on adversarial learning and anomaly detection for DNNs can be found in \cite{biggio2018wild, miller2020adversarial, bulusu2020survey}.

Prior works on adversarial and OOD detection can be broadly categorized into unsupervised and supervised methods. Supervised methods such as \cite{lee2018simple}, \cite{ma2018characterizing_iclr}, and \cite{yang2019ml} use a training set of adversarial or OOD samples (\ie known anomalies) to train a binary classifier that discriminates natural inputs from anomalies. They extract specific informative test statistics from the layer representations as features for the classifier. On the other hand, unsupervised methods such as \cite{roth2019odds, zheng2018robust, miller2019ada, sastry2019gram, li2017adversarial, xu2017feature1}, rely on interesting statistical properties and generative modeling of natural inputs at specific (\eg logit) or multiple layer representations of the DNN for detection. 
Works such as the trust score~\cite{jiang2018trust}, deep kNN~\cite{papernot2018deep}, and by \citet{jha2019abc} have explored the problem of developing a confidence metric that can independently validate the predictions of a classifier. 
Inputs with low confidence scores are likely to be misclassified and hence are detected as anomalies.

\section{Anomaly Detection Meta-algorithm}
\label{sec:proposed}
We first introduce the notation and problem setup, followed by a description of the proposed meta-algorithm.

\vspace{-3mm}
\subsection{Notations and Setup}
Consider the conventional classification problem where the goal is to accurately classify an input $\,\bfx \in \mathcal{X}\,$ into one of $m$ classes $\,[m] := \{1, \cdots, m\}$. We focus on DNN classifiers that learn a function of the form $\,\mathbf{F}(\bfx) \,=\, [F_1(\bfx), \cdots, F_m(\bfx)], ~\mathbf{F} : \mathcal{X} \mapsto \Delta_m\,$, where $\mathcal{X}$ is the space of inputs to the DNN and $\,\Delta_m = \{(p_1, \cdots, p_m) \in [0, 1]^m \,:\, \sum_i p_i = 1\}\,$ is the space of output class probabilities. 
The class prediction of the DNN based on its output class probabilities is defined as $\,\widehat{C}(\bfx) = \argmax_{c \in [m]} F_c(\bfx)\,$.
The multi-layer architecture of a DNN allows the input-output mapping to be expressed as a composition of multiple functions, \ie $\,\mathbf{F}(\bfx) = (\mathbf{g}^{}_L \circ \mathbf{g}^{}_{L-1} \cdots \circ \mathbf{g}^{}_1)(\bfx)\,$, where $L$ is the number of layers. The output from an intermediate layer $\,\ell \in \{1, \cdots, L\}\,$ of the DNN, $\mathbf{f}_\ell(\bfx) = (\mathbf{g}^{}_\ell \circ \cdots \circ \mathbf{g}^{}_1)(\bfx) \in \reals^{d_\ell}$, is referred to as its layer representation~\footnote{Layers with tensor-valued outputs (\eg convolution) are flattened into vectors. Boldface symbols are used for vectors and tensors.}. We also use the shorthand notation $\,\bfx^{(\ell)} = \mathbf{f}_\ell(\bfx)$, with $\,\bfx^{(0)} = \mathbf{f}_0(\bfx)\,$ denoting the vectorized input. 
The set of layers and distinct layer pairs are denoted by $\,\mathcal{L} = \{0, \cdots, L\}\,$ and $\,\mathcal{L}^2 = \{(\ell_1, \ell_2) \in \mathcal{L} \times \mathcal{L} \,:\, \ell_2 > \ell_1\}$.
Table~\ref{tab:notations_app} in the Appendix provides a quick reference for the notations.

We assume access to the trained DNN classifier to defend, and a labeled data set $\,\mathcal{D} = \lbrace(\bfx_n, c_n), ~n = 1, \cdots, N\rbrace\,$ that is different from the one used to train the DNN and does not contain any anomalous samples.
We define an augmented data set $\,\mathcal{D}_a = \lbrace(\bfx^{(0)}_n, \cdots, \bfx^{(L)}_n, c_n, \hat{c}_n), ~n = 1, \cdots, N\rbrace\,$ that is obtained by passing samples from $\mathcal{D}$ through the DNN and extracting their layer representations $\,\bfx^{(\ell)}_n = \mathbf{f}_\ell(\bfx_n), ~\ell \in \mathcal{L}\,$ and the class prediction $\,\hat{c}_n = \widehat{C}(\bfx_n)$. We also define subsets of $\mathcal{D}_a$ corresponding to each layer $\ell \in \mathcal{L}$, and each predicted class $\,\hat{c} \in [m]\,$ or true class $\,c \in [m]\,$ respectively as:
\vspace{-2mm}
\begin{align*}
\widehat{\mathcal{D}}_a(\ell, \hat{c}) ~&=~ \{(\bfx^{(\ell)}_n, c_n, \hat{c}_n), ~n = 1, \cdots, N \,:\, \hat{c}_n = \hat{c}\}, \\
\mathcal{D}_a(\ell, c) ~&=~ \{(\bfx^{(\ell)}_n, c_n, \hat{c}_n), ~n = 1, \cdots, N \,:\, c_n = c\}.
\end{align*}

\subsection{Components of the Meta-algorithm}
\label{sec:overview}
%
\begin{algorithm}[htb]
	\caption{Meta-algorithm for Anomaly Detection}
	\label{meta_algorithm}
	\algsetup{linenosize=\small}
    \small
	\begin{algorithmic}[1]
	    \STATE {\bfseries Inputs:} Trained DNN $\mathbf{F}(\cdot)$, Dataset $\mathcal{D}$, Test input $\bfx$, \\FPR $\alpha$ or detection threshold $\tau$.
		\STATE {\bfseries Output:} Detector decision -- normal $0$ or anomaly $1$.
		\medskip
		\STATE {\bf Preprocessing:}
		\STATE Calculate the detection threshold $\tau$ (if not specified).
		\STATE Calculate the class prediction and layer representations of $\bfx$.
		\STATE Create the data subsets corresponding to each layer, predicted class, and $m$ true classes.
		\smallskip
		\STATE {\bf I. Test statistics (TS)}:
		\begin{ALC@g}
		    \NoDo
		    \FOR{each layer $\ell$:}
		        \STATE Calculate the TS at layer $\ell$ conditioned on the predicted class and the $m$ candidate true classes.
	        \ENDFOR
            \smallskip
            \STATE Compile the $\,m + 1\,$ TS vectors from the layers. 
        \end{ALC@g}
        \smallskip
        \STATE {\bf II. Normalizing transformations}:
        \begin{ALC@g}
            \NoThen
            \IF{multivariate normalization:}
                \STATE Normalize each of the $\,m + 1\,$ TS vectors.
            \ELSE
                \NoDo
                \FOR{each layer $\ell$:}
	                \STATE Normalize the $\,m + 1\,$ TS from layer $\ell$.
	            \ENDFOR
                \smallskip
                \OptDo
                \FOR{each distinct layer pair ($\ell_1, \ell_2)$:}
	                \STATE Normalize the $\,m + 1\,$ TS pairs from layers $\ell_1, \ell_2$.
	            \ENDFOR
            \ENDIF
        \end{ALC@g}
        \smallskip
        \STATE {\bf III. Layerwise aggregation and scoring}:
        \begin{ALC@g}
            \NoThen
            \IF{multivariate normalization:}
                \STATE No need to aggregate the normalized TS.
            \ELSE
                \STATE Aggregate the normalized TS from the layers and layer pairs for the predicted class and each candidate true class.
            \ENDIF
            \smallskip
            \STATE Calculate the final score from the $\,m + 1\,$ aggregated normalized TS.
        \end{ALC@g}
        \smallskip
        \STATE {\bf IV. Detection decision}:
        \begin{ALC@g}
            \STATE Return anomaly ($1$) if the final score exceeds threshold; Else return normal ($0$).
        \end{ALC@g}
	\end{algorithmic}
\end{algorithm}
The proposed meta-algorithm for detecting anomalous inputs to a DNN classifier based on its layer representations is given in Algorithm~\ref{meta_algorithm}.
A more formal version of the same can be found in Algorithm~\ref{meta_algorithm_detailed} in the Appendix.
Details of the individual components of the meta-algorithm are discussed next.
For this discussion, consider a test sample $\bfx$ whose true class is unknown, predicted class is $\,\widehat{C}(\bfx) = \hat{c}$, and layer representations are $\,\bfx^{(\ell)}, ~\ell \in \mathcal{L}$.
\begin{figure}[thb]
\vspace{-2mm}
  \centering
  \includegraphics[scale=0.48]{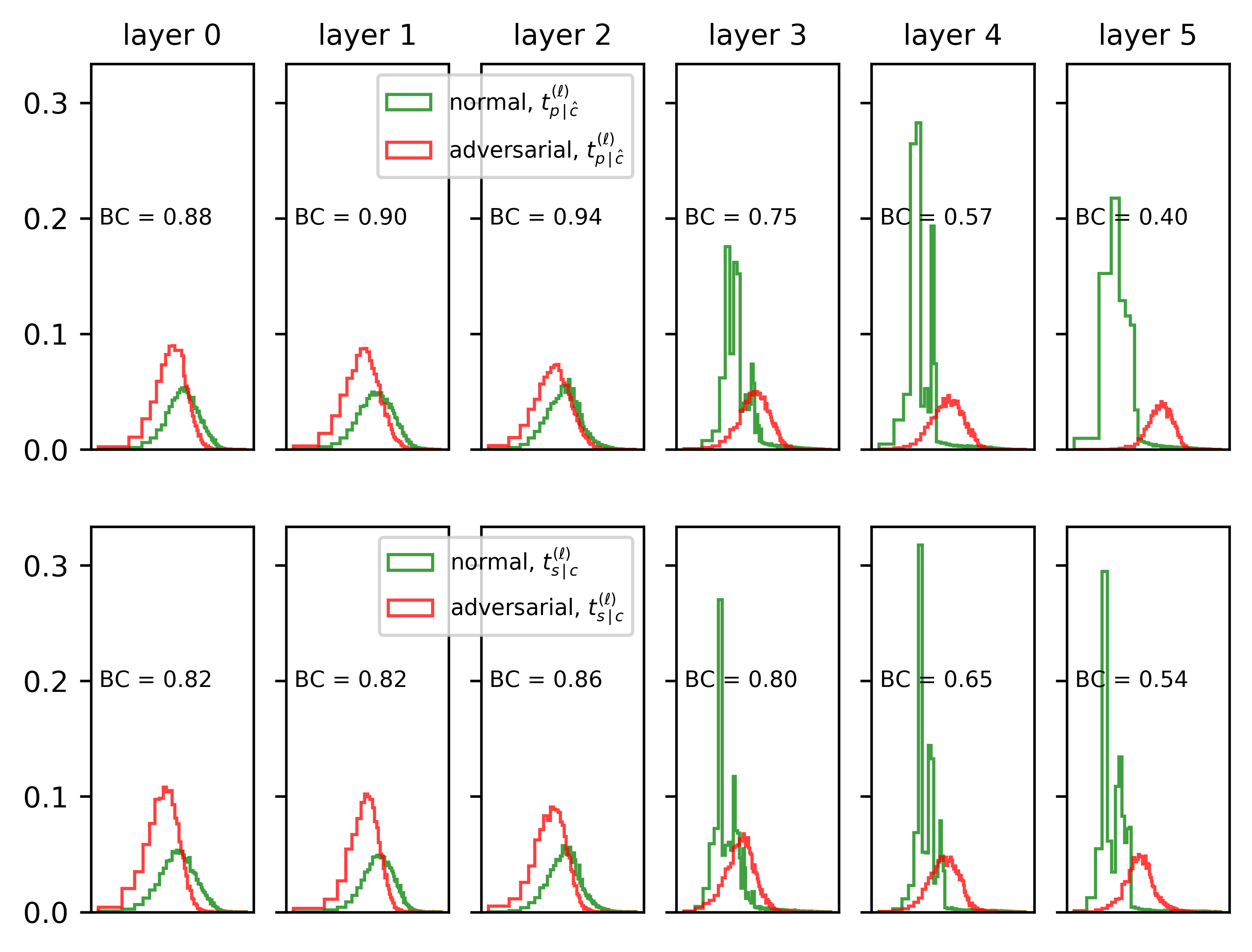}
  \vspace{-2mm}
  \caption{\small Distribution of test statistics corresponding to normal and adversarial samples (PGD, $\ell_\infty$ attack) from the layers of a DNN trained on the SVHN dataset. The top and bottom figures show the multinomial test statistic (\S~\ref{sec:multinom_test_stat}) conditioned on the predicted and true class respectively. BC is the Bhattacharya coefficient, which is a measure of distribution overlap.}
\vspace{-5mm}
\label{fig:test_stats_distr}
\end{figure}

\noindent{\bf I. Test Statistics:}
In line with prior works that use the layer representations of a DNN for detection, we define test statistics at each layer that capture a statistical property of the layer representation useful for detection (\eg Mahalanobis distances~\cite{lee2018simple}). 
The test statistics are defined to be conditioned on the predicted class and on each candidate true class (since the true class is unknown). 
The latter is particularly useful for adversarial samples, since they are known to have originated from one of the $m$ classes.
For a test input $\bfx$ predicted as class $\hat{c}$, the test statistic at any layer $\ell$ conditioned on the predicted class is defined as $\,T(\bfx^{(\ell)}, \hat{c}, \widehat{\mathcal{D}}_a(\ell, \hat{c}))\,$. It captures how anomalous the layer representation $\bfx^{(\ell)}$ is with respect to the distribution of natural inputs predicted into class $\hat{c}$ by the DNN. 
Similarly, a set of $m$ test statistics at layer $\ell$, conditioned on each candidate true class $c$, are defined as $\,T(\bfx^{(\ell)}, c, \mathcal{D}_a(\ell, c)), ~c \in [m]$. They capture how anomalous $\bfx^{(\ell)}$ is with respect to the distribution of natural inputs from a true class $c$. 
A mild requirement on the definition of the test statistic function is that larger values of $T(\cdot)$ correspond to larger deviations of the layer representation from the class-conditional distribution of natural inputs. 
When the input is clear from the context, we denote the test statistic random variables by $\,T^{(\ell)}_{p \cond \hat{c}} := T(\bfx^{(\ell)}, \hat{c}, \widehat{\mathcal{D}}_a(\ell, \hat{c}))\,$ and  $\,T^{(\ell)}_{s \cond c} := T(\bfx^{(\ell)}, c, \mathcal{D}_a(\ell, c))$. Specific values of the test statistic are denoted by $t^{(\ell)}_{p \cond \hat{c}}$ and $t^{(\ell)}_{s \cond c}$. 
The vector of test statistics across the layers is defined as $\,\bft_{p \cond \hat{c}} := [t^{(0)}_{p \cond \hat{c}}, \cdots, t^{(L)}_{p \cond \hat{c}}]\,$ given the predicted class $\hat{c}$, and as $\,\bft_{s \cond c} := [t^{(0)}_{s \cond c}, \cdots, t^{(L)}_{s \cond c}], ~\forall c \in [m]\,$ given each candidate true class.
In \S~\ref{sec:multinom_test_stat}, we propose a test statistic 
based on the multinomial likelihood ratio test (LRT) applied to class counts from the $k$-nearest neighbors (kNN) of a layer representation.
However, the above definitions are general and apply to test statistics proposed in prior works such as Gram matrix-based deviations~\cite{sastry2019gram}.

\noindent{\bf II. Distribution-Independent Normalization:}
In the absence of any prior assumptions, the class-conditional and marginal distributions of the test statistics are unknown and expected to change across the DNN layers (\eg see Fig.~\ref{fig:test_stats_distr}). Therefore, in order to effectively combine the test statistics from the layers for anomaly scoring, it is important to apply a normalizing transformation that (ideally) makes the transformed test statistics distribution independent. Some prior works partially address this using heuristic approaches such as z-score normalization~\cite{roth2019odds} and scaling by the expected value~\cite{sastry2019gram} in order to account for the distribution and range differences of the test statistics across the layers. 
We {\em propose two approaches for applying normalizing transformations} -- the first one focuses on test statistics from the individual layers and layer pairs, and the second one focuses on the vector of test statistic across the layers. Considering test statistic pairs and the vector of test statistics allows our method to capture the joint effect of anomalous inputs on the layer representations.

Consider the first approach.
In the meta-algorithm, such normalizing transformations are defined as $\,q(t^{(\ell)}_{p \cond \hat{c}})\,$ for the test statistic at layer $\ell$ conditioned on the predicted class $\hat{c}$, and $\,q(t^{(\ell)}_{s \cond c}), ~\forall c \in [m]\,$ for the test statistic at layer $\ell$ conditioned on each candidate true class. For each pair of layers $\,(\ell_1, \ell_2) \in \mathcal{L}^2$, $\,q(t^{(\ell_1)}_{p \cond \hat{c}}, t^{(\ell_2)}_{p \cond \hat{c}})\,$ and $\,q(t^{(\ell_1)}_{s \cond c}, t^{(\ell_2)}_{s \cond c}), ~\forall c \in [m]\,$ define the normalizing transformations for the corresponding test statistic pair conditioned on the predicted class $\hat{c}$ and on each candidate true class respectively.
Since it is not efficient to include all the layer pairs beyond few tens of layers, this is specified as optional in Algorithm~\ref{meta_algorithm}.

In the second approach, $q(\bft_{p \cond \hat{c}})\,$ and $\,q(\bft_{s \cond c}), ~\forall c \in [m]\,$ define the normalizing transformations for a vector of test statistics from the layers conditioned on the predicted class $\hat{c}$ and on each candidate true class respectively~\footnote{Although the function $q(\cdot)$ is different depending on the class, layer(s), and the number of test statistic inputs, we use the same overloaded notation for clarity.}. 
In \S~\ref{sec:normalizing_pvalues}, we propose specific realizations for each case of the above normalizing transformations based on {\em class-conditional p-values}. They have the advantage of being nonparametric, and transform the test statistics into probabilities that, for natural inputs, will be 
approximately uniform on $[0, 1]$.

\noindent{\bf III. Layerwise Aggregation and Scoring:}
The normalized test statistics can be interpreted as anomaly scores that are each based on information from one or more layer representations and a specific (predicted or true) class. 
The goal of a scoring function is to aggregate the multiple anomaly scores in a principled way such that the {\em combined score is low} for inputs following the same distribution as normal inputs to the DNN, and {\em high for anomalies}. 
Prior works have taken approaches such as average or maximum of the normalized test statistics~\cite{miller2019ada, sastry2019gram}, or a weighted sum of unnormalized test statistics, with the weights trained using a binary logistic classifier~\cite{lee2018simple, ma2018characterizing_iclr, yang2019ml}. 
In our meta-algorithm, we define an aggregation function $\,r(\cdot)\,$ that combines the set of all normalized test statistics from the individual layers and (optionally) layer pairs 
as follows: $\,q^{}_{\textrm{agg}}(\bft_{p \cond \hat{c}}) \,=\, r(Q_{p \cond \hat{c}})\,$ and $\,q^{}_{\textrm{agg}}(\bft_{s \cond c}) \,=\, r(Q_{s \cond c}), ~\forall c \in [m]$, where
\vspace{-1mm}
\begin{align}
\label{eq:pvalue_sets}
Q_{p \cond \hat{c}} ~&=~ \{q(t^{(\ell)}_{p \cond \hat{c}}), ~\forall \ell \in \mathcal{L}\} \nonumber \\ 
~&\cup~ \{q(t^{(\ell_1)}_{p \cond \hat{c}}, t^{(\ell_2)}_{p \cond \hat{c}}), ~\forall (\ell_1, \ell_2) \in \mathcal{L}^2\} ~~\mbox{ and} \\
Q_{s \cond c} ~&=~ \{q(t^{(\ell)}_{s \cond c}), ~\forall \ell \in \mathcal{L}\} \nonumber \\
~&\cup~ \{q(t^{(\ell_1)}_{p \cond c}, t^{(\ell_2)}_{p \cond c}), ~\forall (\ell_1, \ell_2) \in \mathcal{L}^2\}, ~~\forall c \in [m] \nonumber
\end{align}
define the sets of normalized test statistics.

Motivated by ideas from multiple testing, {\em we propose specific aggregation functions} $\,r(\cdot)$ in \S~\ref{sec:pvalue_aggregation} for combining multiple p-values from the layers and layer pairs. 
For the normalization approach that directly transforms the test statistic vector from the layers, there is no need for an aggregation function; we simply set $\,q^{}_{\textrm{agg}}(\bft_{p \cond \hat{c}}) = q(\bft_{p \cond \hat{c}})\,$ and $\,q^{}_{\textrm{agg}}(\bft_{s \cond c}) = q(\bft_{s \cond c}), ~\forall c \in [m]$.
The final anomaly score in the meta-algorithm is defined to be a simple function of the aggregate, normalized test statistics, \ie $\,S(q^{}_{\textrm{agg}}(\bft_{p \cond \hat{c}}), q^{}_{\textrm{agg}}(\bft_{s \cond 1}), \cdots, q^{}_{\textrm{agg}}(\bft_{s \cond m}), \hat{c})$.
We propose specific realizations of the score functions for adversarial and OOD detection in \S~\ref{sec:scoring}.

\noindent{\bf IV. Detection Decision:}
The detection decision for a test input $\bfx$ predicted into class $\hat{c}$ is obtained by thresholding the final anomaly score as follows:
\vspace{-2mm}
\begin{align}
\label{eq:detection_rule}
&\psi^{}_{\tau}(\bfx^{(0)}, \cdots, \bfx^{(L)}, \hat{c}) ~= \nonumber \\
&\indicator\!\left[S(q^{}_{\textrm{agg}}(\bft_{p \cond \hat{c}}), q^{}_{\textrm{agg}}(\bft_{s \cond 1}), \cdots, q^{}_{\textrm{agg}}(\bft_{s \cond m}), \hat{c}) \geq \tau\right]
\end{align}
where decisions $0$ and $1$ correspond to natural and anomalous inputs respectively, and $\,\indicator[\cdot]\,$ is the indicator function.
The threshold $\tau$ is usually set based on a false positive rate (FPR) that is suitable for the target application. 
In order to operate the detector at an FPR $\,\alpha \in (0, 1)\,$ (usually a small value \eg 0.01), the threshold can be set by estimating the FPR $\,\widehat{P}^{}_{\textrm{F}}(\tau)\,$ from the set of natural inputs $\mathcal{D}_a$ as follows:
\vspace{-2mm}
\begin{align}
\label{eq:threshold}
\tau^{}_{\alpha} ~&=~ \sup\{\tau \in \reals \,:\, \widehat{P}^{}_{\textrm{F}}(\tau) \leq \alpha\}, \mbox{ where} \nonumber \\
\widehat{P}^{}_{\textrm{F}}(\tau) ~&=~ \frac{1}{N} \mysum_{n=1}^N \,\psi^{}_{\tau}(\bfx^{(0)}_n, \cdots, \bfx^{(L)}_n, \hat{c}_n).
\end{align}
This threshold choice ensures that natural inputs are accepted by the detector with a probability close to $\,1 - \alpha$.

\section{A Realization of the Meta-algorithm}
\label{sec:instance_meta_algo}
In this section, we propose concrete methods for realizing the components of the anomaly detection meta-algorithm. 

\vspace{-2mm}
\subsection{Test Statistic Based on kNN Class Counts}
\label{sec:multinom_test_stat}
Consider a set of natural inputs to the DNN that are predicted into a class $\hat{c} \in [m]$. 
The class counts from the kNN of its representations from a layer $\ell \in \mathcal{L}$ are expected to follow a certain distribution, wherein class $\hat{c}$ has a higher probability than the other classes.
A similar observation can be made for natural inputs from a candidate true class $c \in [m]$.
Let $\,(k^{(\ell)}_1, \cdots, k^{(\ell)}_m)\,$ denote the tuple of class counts from the kNN $\,N^{(\ell)}_k(\bfx^{(\ell)})\,$ of a layer representation $\bfx^{(\ell)}$, such that $\,k^{(\ell)}_i \in \lbrace0, 1, \cdots, k\rbrace\,$ and $\,\sum_{i=1}^m k^{(\ell)}_i = k$. The null (natural) distribution of the kNN class counts at a layer $\ell$ conditioned on the predicted class $\hat{c}$ can be modeled using the following multinomial distribution
\vspace{-2mm}
\begin{equation}
\label{eq:multinom}
p(k^{(\ell)}_1, \cdots, k^{(\ell)}_m \cond \widehat{C}=\hat{c}) ~=~ k! \,\myprod_{i=1}^m \frac{[\pi^{(\ell)}_{i \cond \hat{c}}]^{k^{(\ell)}_i}}{k^{(\ell)}_i!},
\end{equation}
where $\,(\pi^{(\ell)}_{1 \cond \hat{c}}, \cdots, \pi^{(\ell)}_{m \cond \hat{c}})\,$ are the multinomial probability parameters specific to class $\hat{c}$ and layer $\ell$ (they are non-negative and sum to $1$). 
We estimate these parameters from the labeled subset $\widehat{\mathcal{D}}_a(\ell, \hat{c})\,$ using maximum-a-posteriori (MAP) estimation with the Dirichlet conjugate prior distribution~\cite{barber2012}~\footnote{We set the prior counts of the Dirichlet distribution to a small non-zero value to avoid $0$ estimates for the multinomial parameters.}.
For a test input $\bfx$ sampled from the natural data distribution that is predicted into class $\hat{c}$ by the DNN, we expect the multinomial distribution (\ref{eq:multinom}) to be a good fit for the class counts observed from its kNN at layer $\ell$.
In order to test whether the observed class counts $\,(k^{(\ell)}_1, \cdots, k^{(\ell)}_m)\,$ from layer $\ell$ given predicted class $\hat{c}$ are consistent with distribution (\ref{eq:multinom}), 
we apply the well-known multinomial LRT~\cite{read2012goodness}, whose log-likelihood ratio statistic is given by
\vspace{-2mm}
\begin{equation}
\label{eq:test_stat_lrt_pred}
T(\bfx^{(\ell)}, \hat{c}, \widehat{\mathcal{D}}_a(\ell, \hat{c})) ~=\, \mysum_{i=1}^m \,k^{(\ell)}_i \,\log\frac{k^{(\ell)}_i}{k \,\pi^{(\ell)}_{i \cond \hat{c}}}.
\end{equation}
This test statistic is a class count deviation measure which is always non-negative, with larger values corresponding to a larger deviation from the null distribution (\ref{eq:multinom}).
In a similar way, the test statistics conditioned on each candidate true class are defined as
\vspace{-2mm}
\begin{equation}
\label{eq:test_stat_lrt_true}
T(\bfx^{(\ell)}, c, \mathcal{D}_a(\ell, c)) ~=\, \mysum_{i=1}^m \,k^{(\ell)}_i \,\log\frac{k^{(\ell)}_i}{k \,\widetilde{\pi}^{(\ell)}_{i \cond c}}, ~\forall c \in [m].
\end{equation}
Here, $\,(\widetilde{\pi}^{(\ell)}_{1 \cond c}, \cdots, \widetilde{\pi}^{(\ell)}_{m \cond c})\,$ are the multinomial parameters specific to class $c$ and layer $\ell$, which are estimated from the corresponding data subset $\mathcal{D}_a(\ell, c)$.

\subsection{Normalizing Transformations Based on p-values}
\label{sec:normalizing_pvalues}
Recall that we are interested in designing a normalizing transformation that, for natural inputs, makes the transformed test statistics across the layers and classes follow the same distribution. 
One such approach is to use the p-value, that calculates the probability of a test statistic taking values (as or) more extreme than the observed value.
More generally, a p-value is defined as any transformation of the test statistic (possibly a vector), following the null hypothesis distribution, into a uniformly distributed probability~\cite{root2016learning}.
%
%
This provides a simple approach for normalizing the class-conditional test statistics in both the univariate and multivariate cases, as discussed next.

\noindent{\bf A. p-values at Individual Layers and Layer Pairs}

For an input predicted into class $\hat{c}$ with class-conditional test statistics at a layer $\ell$ given by $\,t^{(\ell)}_{p \cond \hat{c}}, t^{(\ell)}_{s \cond 1}, \cdots, t^{(\ell)}_{s \cond m}$, the normalizing p-value transformations are defined as~\footnote{We use one-sided, right-tailed p-values since larger values of the test statistic correspond to a larger deviation.}:
%
%
\vspace{-2mm}
\begin{align}
\label{eq:pval_scalar}
q(t^{(\ell)}_{p \cond \hat{c}}) ~&=~ \proba(\,T^{(\ell)}_{p \cond \hat{c}} \geq t^{(\ell)}_{p \cond \hat{c}} \cond \widehat{C} = \hat{c}\,) \nonumber \\
q(t^{(\ell)}_{s \cond c}) ~&=~ \proba(\,T^{(\ell)}_{s \cond c} \geq t^{(\ell)}_{s \cond c} \cond C = c\,), ~~\forall c \in [m].
\end{align}
Similarly the normalizing p-value transformations for test statistic pairs from layers $\,(\ell_1, \ell_2)\,$ are defined as:
\vspace{-1mm}
\begin{align}
\label{eq:pval_pair}
&q(t^{(\ell_1)}_{p \cond \hat{c}}, t^{(\ell_2)}_{p \cond \hat{c}}) ~=~ \proba(\,T^{(\ell_1)}_{p \cond \hat{c}} \geq t^{(\ell_1)}_{p \cond \hat{c}}, T^{(\ell_2)}_{p \cond \hat{c}} \geq t^{(\ell_2)}_{p \cond \hat{c}} \cond \widehat{C} = \hat{c}\,) \nonumber \\
&q(t^{(\ell_1)}_{s \cond c}, t^{(\ell_2)}_{s \cond c}) ~= \nonumber \\
&\proba(\,T^{(\ell_1)}_{s \cond c} \geq t^{(\ell_1)}_{s \cond c}, T^{(\ell_2)}_{s \cond c} \geq t^{(\ell_2)}_{s \cond c} \cond C = c\,), ~~\forall c \in [m].
\end{align}
%
Since the class-conditional distributions of the test statistics are unknown, we estimate the p-values using the empirical cumulative distribution function of the test statistics calculated from the corresponding data subsets of $\mathcal{D}_a$
\footnote{In our implementation, we averaged the p-value estimates from a hundred bootstrap samples in order to reduce the variance. 
}.

\noindent{\bf B. Multivariate p-value Based Normalization}

In this approach, we consider the class-conditional joint density of a test statistic vector from the layers, and propose a normalizing transformation $\,q : \reals^{L + 1} \mapsto [0, 1]\,$ based on the idea of multivariate p-values.
Consider an input predicted into a class $\hat{c}$, that has a vector of test statistics $\bft^{}_{p \cond \hat{c}} = \bft\,$ from the layers. 
Suppose $\,f^{}_0(\bft_{p \cond \hat{c}} \cond \hat{c})\,$ denotes the true null-hypothesis density of $\,\bft_{p \cond \hat{c}}\,$ conditioned on the predicted class $\hat{c}$, then the region outside the level set of constant density equal to $\,f^{}_0(\bft \cond \hat{c})\,$ is given by $\,\{\bft_{p \cond \hat{c}} \in \reals^{L+1} \,:\, f^{}_0(\bft_{p \cond \hat{c}} \cond \hat{c}) \,<\, f^{}_0(\bft \cond \hat{c})\}\,$. 
The multivariate p-value for $\bft$ is the probability of this region under the null hypothesis probability measure~\cite{root2016learning}.

We use the averaged localized p-value estimation method using kNN graphs (aK-LPE) proposed by \cite{qian2012new}. 
The main idea is to define a score function based on nearest neighbor graphs $\,G(\bft)\,$ that captures the local relative density around $\bft$. They show that a score function defined as the average distance from $\bft$ to its $\frac{k}{2}$-th through $\frac{3k}{2}$-th nearest neighbors provides the following asymptotically-consistent p-value estimate:
\vspace{-2mm}
\begin{equation}
\label{eq:pval_klpe}
q^{}_{\textrm{lpe}}(\bft) ~=~ \frac{1}{|\mathcal{D}_t|} \mysum_{\bft_n \in \mathcal{D}_t} \indicator[G(\bft) \leq G(\bft_n)],
\end{equation}
where $\mathcal{D}_t$ is a large sample of test statistic vectors corresponding to natural inputs.
In our problem, we apply the above p-value transformation (using the appropriate data subsets) to normalize the $m + 1$ test statistic vectors giving: $q^{}_{\textrm{lpe}}(\bft^{}_{p \cond \hat{c}}), q^{}_{\textrm{lpe}}(\bft^{}_{s \cond 1}), \cdots, q^{}_{\textrm{lpe}}(\bft^{}_{s \cond m})$.
%

\subsection{Aggregation of p-values}
\label{sec:pvalue_aggregation}
The p-value based normalized test statistics capture the extent of deviation of the test statistics of an input relative to their distribution on natural inputs; smaller p-values correspond to a larger deviation. 
For approach A in \S~\ref{sec:normalizing_pvalues}, we can consider each p-value to correspond to a hypothesis test involving a particular layer or layer pair. We are interested in combining the evidence from these multiple tests~\cite{dudoit2007multiple} into a single p-value for the overall problem of testing for natural versus anomalous inputs. 
We investigate two methods for combining p-values from multiple tests and define the corresponding aggregation functions.
Note that there is no need to aggregate p-values for approach B in \S~\ref{sec:normalizing_pvalues}, and we simply set $\,q^{}_{\textrm{agg}}(\bft^{}_{p \cond \hat{c}}) = q^{}_{\textrm{lpe}}(\bft^{}_{p \cond \hat{c}})\,$ and $\,q^{}_{\textrm{agg}}(\bft^{}_{s \cond c}) = q^{}_{\textrm{lpe}}(\bft^{}_{s \cond c}), ~\forall c \in [m]$.

Fisher's method~\cite{fisher1992statistical} provides a principled way of combining p-values from multiple independent tests based on the idea that, under the null hypothesis, the sum of the log of multiple p-values follows a ${\chi}^2$-distribution.
The aggregate p-value function based on this method is given by
\vspace{-1mm}
\begin{equation}
\log \,q^{}_{\textrm{fis}}(\bft) ~=~ \log \,r(Q) ~= \mysum_{q \in Q} \,\log q , 
\end{equation}
where $Q$ is one of the sets $Q_{p \cond \hat{c}}$ or $Q_{s \cond c}$ defined in Eq. (\ref{eq:pvalue_sets}), and $\bft$ is the corresponding test statistic vector~\footnote{$q^{}_{\textrm{lpe}}(\cdot)$, $q^{}_{\textrm{fis}}(\cdot)$, and $q^{}_{\textrm{hmp}}(\cdot)$ are specific instances of $\,q^{}_{\textrm{agg}}(\cdot)$.}. 
An apparent weakness of Fisher's method is its assumption of independent p-values.
We briefly provide the aggregate p-value function for an alternate harmonic mean p-value (HMP) method for combining p-values from multiple dependent tests~\cite{wilson2019harmonic}, and discuss its details in Appendix~\ref{sec:hmp_method}.
\vspace{-1mm}
\begin{equation}
\label{eq:pvalue_hmp}
q_{\textrm{hmp}}(\bft)^{-1} ~=~ r(Q)^{-1} ~=~ \mysum_{q \in Q} \,q^{-1}.
\end{equation}

\vspace{-3mm}
\subsection{Scoring for Adversarial and OOD Detection}
\label{sec:scoring}
We propose different score functions for detecting adversarial and general OOD inputs.
An adversarial input predicted into class $\hat{c}$ by the DNN is expected to be anomalous at one or more of its layer representations relative to the distribution of natural inputs predicted into the same class.
This implies that its aggregate p-value conditioned on the predicted class, $q^{}_{\textrm{agg}}(\bft_{p \cond \hat{c}})$, should have a small value.
Moreover, since the adversarial input was created from a source class different from $\hat{c}$, it is expected to be a typical sample relative to the distribution of natural inputs from the unknown source class $\,c \ne \hat{c}$. 
This implies that its aggregate p-value conditioned on a candidate true class (different from $\hat{c}$) should have a relatively large value. 
Combining these ideas, we define the score function for adversarial inputs as
\vspace{-1mm}
\begin{align}
\label{eq:score_func_adver}
&S(q^{}_{\textrm{agg}}(\bft_{p \cond \hat{c}}), q^{}_{\textrm{agg}}(\bft_{s \cond 1}), \cdots, q^{}_{\textrm{agg}}(\bft_{s \cond m}), \hat{c}) \nonumber \\ 
&=~ \log\left( \frac{\displaystyle\max_{c \in [m] \setminus \{\hat{c}\}} \,q^{}_{\textrm{agg}}(\bft_{s \cond c})}{q^{}_{\textrm{agg}}(\bft_{p \cond \hat{c}})} \right).
\end{align}
The table below provides additional insight on this score function by considering the numerator and denominator terms (inside the $\log$) for different categories of input.
\begin{table}[H]
\captionsetup{font=small,skip=10pt}
\small
\centering
\vspace{-2mm}
\resizebox{0.45\textwidth}{!}{%
\begin{tabular}{@{}llll@{}}
\toprule
Input type \& prediction & Numerator & Denominator & Score \\ 
\midrule
$\bfx$ natural, $\,\widehat{C}(\bfx) = c$ & Low & High & Low \\
$\bfx$ natural, $\,\widehat{C}(\bfx) \neq c$ & High & High & Medium \\
$\bfx$ adversarial, $\,\widehat{C}(\bfx) \neq c$ & High & Low & High \\ 
\bottomrule
\end{tabular}%
}
\label{tab:adversarial_score}
\vspace{-3mm}
\end{table}
Similar to adversarial inputs, OOD inputs are also expected to exhibit anomalous patterns at the layers of the DNN relative to the distribution of natural inputs predicted into the same class. 
Since OOD inputs are not created by intentionally perturbing natural inputs from a true class different from the predicted class, we simplify score function (\ref{eq:score_func_adver}) for OOD detection as follows
%
\vspace{-1mm}
\begin{align*}
\label{eq:score_func_ood}
S(q^{}_{\textrm{agg}}(\bft_{p \cond \hat{c}}), q^{}_{\textrm{agg}}(\bft_{s \cond 1}), \cdots, q^{}_{\textrm{agg}}(\bft_{s \cond m}), \hat{c}) 
\,=\, -\log \,q^{}_{\textrm{agg}}(\bft_{p \cond \hat{c}})
\end{align*}

%
OOD inputs are expected to have a low aggregate p-value $\,q^{}_{\textrm{agg}}(\bft_{p \cond \hat{c}})$, and hence a high value for the above score.

\subsection{Implementation and Computational Complexity}
\label{sec:comp_complexity}
We briefly discuss some practical aspects of implementing \proposed efficiently.
We apply the neighborhood preserving projection method~\cite{he2005neighborhood} to perform dimensionality reduction on the DNN layer representations since they can be very high dimensional (details in Appendix~\ref{app:code}).
In order to efficiently construct and query from kNN graphs at the layer representations, we use the fast approximate nearest neighbors method {\em NNDescent}~\cite{dong2011efficient} \footnote{We use the following implementation of NNDescent: \url{https://github.com/lmcinnes/pynndescent}}.
Together, these two techniques significantly reduce the memory utilization and running time of \proposed.

The computational complexity of the proposed instantiation of \proposed at prediction (test) time can be expressed as $\BigO{L\,(d_{\max} \,N^{\rho} \,+\, m^2 \,+\, B\,N)}$ when layer pairs are not used, and $\,\BigO{L^2\,(d_{\max} \,N^{\rho} \,+\, m^2 \,+\, B\,N)}\,$ when layer pairs are used. Here $d_{\max}$ is the maximum dimension of the projected layer representations, $m$ is the number of classes, $N$ is the number of samples, $B$ is the number of bootstrap replications used for estimating p-values, and $\rho \in (0, 1)$ is an unknown factor associated with the approximate nearest neighbor queries (that are sub-linear in $N$). The p-value calculation can be made faster and independent of $N$ by pre-computing the empirical class-conditional CDFs. 
A comparison of the running time of \proposed with other detection methods can be found in Appendix~\ref{sec:app_running_time}.

\section{Defense-Aware Adaptive Attack}
\label{sec:dknn_attack}
The importance of evaluating adversarial detection methods against an adaptive, defense-aware adversary has been highlighted in prior works~\cite{carlini2017bypassing, athalye2018obfuscated, tramer2020adaptive}. We consider a gray-box adversary that is assumed to have full knowledge of the DNN architecture and parameters, and partial knowledge of the detection method~\footnote{For example, the detection threshold and specific layers of the DNN used may be unknown.}. 

Consider a clean input sample $\bfx$ from class $c$ that is correctly classified by the DNN. Let $\eta^{}_\ell$ denote the distance between $\,\bfx^{(\ell)} = \mathbf{f}_\ell(\bfx)\,$ and its $k$-th nearest neighbor from layer $\ell$. The number of samples from any class $i$ among the kNNs of $\bfx^{(\ell)}$, relative to the dataset $\,\mathcal{D}_a$, can be expressed as
\vspace{-3mm}
\begin{equation*}
\label{eq:class_count}
%
k^{(\ell)}_i ~=~ \!\mysum_{n=1\,:\,c_n = i}^N u(\eta_\ell - d(\mathbf{f}_\ell(\bfx), \mathbf{f}_\ell(\bfx_n))), ~~i = 1, \cdots, m,
\end{equation*}
where $\,u(\cdot)\,$ is the unit step function. Consider the following probability mass function over the class labels: $\,p_i \,=\, k_i \,/\, \sum_{j=1}^m k_j, ~i \in [m]\,$, where $\,k_i = \sum_{\ell=0}^L k^{(\ell)}_i\,$ is the cumulative kNN count from class $i$ across the layers. In order to fool a defense method relying on the kNN class counts from the layer representations, our attack finds an adversarial input $\,\bfx^\prime \,=\, \bfx + \bfdelta\,$ with target class $\,c^\prime \neq c\,$ that minimizes the following log-ratio of probabilities:
\vspace{-3mm}
\begin{equation}
\label{eq:adver_obj1}
\log \frac{p_c}{p_{c^\prime}} ~=~ \log k_c ~-~ \log k_{c^\prime} ~=~ \log\sum_{\ell=0}^L k^{(\ell)}_c ~-~ \log\sum_{\ell=0}^L k^{(\ell)}_{c^\prime},
\end{equation}
subject to a penalty on the norm of the perturbation $\bfdelta$~\footnote{A similar type of attack on kNN-based models has been recently proposed in \cite{sitawarin2020minimum}.}. 
To address the non-smoothness arising from the step function in the class counts, we use the Gaussian (RBF) kernel $\,h_{\sigma}(\bfx, \bfy) \,=\, e^{-\frac{1}{\sigma^2} \,d(\bfx, \bfy)^2}\,$ to obtain a smooth approximation of the class counts $\,k^{(\ell)}_i$.
The attack objective function to minimize is a weighted sum of the $\ell_2$-perturbation norm and the kernel-smoothed log-ratio of probabilities, given by
\vspace{-2mm}
\begin{align}
\label{eq:knn_attack_objec}
J(\bfdelta) ~&=~ \|\bfdelta\|^2_2 ~+~ \lambda \,\log \mysum_{\ell=0}^L \mysum_{\substack{n=1\,:\\ c_n = c}}^N h_{\sigma^{}_\ell}(\mathbf{f}_\ell(\bfx + \bfdelta), \mathbf{f}_\ell(\bfx_n)) \nonumber \\
~&-~ \lambda \,\log \mysum_{\ell=0}^L \mysum_{\substack{n=1\,:\\ c_n = c^\prime}}^N h_{\sigma^{}_\ell}(\mathbf{f}_\ell(\bfx + \bfdelta), \mathbf{f}_\ell(\bfx_n)).
\end{align}
Here $\,\sigma^{}_\ell > 0\,$ is the kernel scale for layer $\ell$ and $\,\lambda > 0\,$ is a constant that sets the relative importance of the terms in the objective function. 
The method used for setting the kernel scale per layer and minor extensions of the proposed attack are described in Appendix~\ref{sec:app_custom_attack}. 
Details of the optimization method and the choice of $\lambda$ are given in Appendix~\ref{app:attacks}.
In our experiments, we chose the class with the second highest probability predicted by the DNN as the target attack class.

\section{Experimental Results}
\label{sec_exp}
We evaluated \proposed on the following well-known image classification datasets: CIFAR-10~\cite{krizhevsky2009learning}, SVHN~\cite{netzer2011reading}, and MNIST~\cite{lecun1998gradient}. We used the training partition provided by the datasets for training standard CNN architectures, including a Resnet for CIFAR-10. We performed class-stratified 5-folds cross-validation on the test partition provided by the datasets; the training folds are used for estimating the detector parameters, and the test folds are used solely for calculating performance metrics (which are then averaged across the test folds). 
We used the Foolbox library~\cite{rauber2017foolbox} for generating adversarial samples from the following attack methods: (i) Projected Gradient Descent (PGD) with $\ell_\infty$ norm~\cite{madry2018towards}, (ii) Carlini-Wagner (CW) attack with $\ell_2$ norm~\cite{carlini2017towards}, and (iii) Fast gradient sign method (FGSM) with $\ell_\infty$ norm~\cite{goodfellow2015explain}. We also implement and generate adversarial samples from the adaptive attack proposed in \S~\ref{sec:dknn_attack}. 
More details on the datasets, DNN architectures, and the attack parameters used are provided in Appendix~\ref{app:datasets}. 

\noindent{\bf Methods Compared.}
We evaluated the following two variants of \proposed using the multinomial test statistic: 1) p-value normalization at the layers and layer pairs using Fisher's method for aggregation, 2) multivariate p-value normalization based on the aK-LPE method. 
The score functions from \S~\ref{sec:scoring} for adversarial and OOD detection are used for the respective tasks.
The number of nearest neighbors is the {\bf only hyperparameter} of the proposed instantiation of \proposed. This is set to be a function of the number of in-distribution training samples $n$ using the heuristic $\,k = \ceil{n^{0.4}}$.
%
\begin{figure*}[htb]
\includegraphics[width=0.8\linewidth]{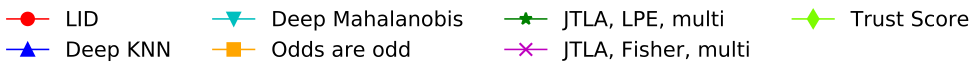}
\vspace{-3mm}
\centering
\subfloat[{{\small CW, confidence $= 0$}}]{\label{fig:avg_prec_cifar_cw}{\includegraphics[width=0.3\linewidth]{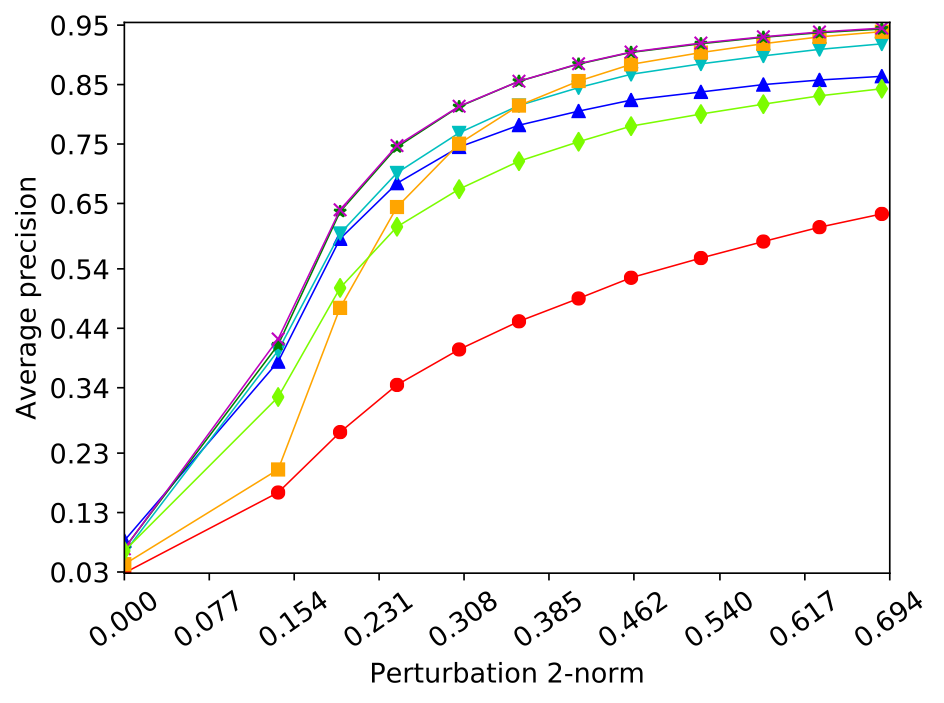}} }
\subfloat[{{\small Adaptive attack}}]{\label{fig:avg_prec_cifar_Custom}{\includegraphics[width=0.3\linewidth]{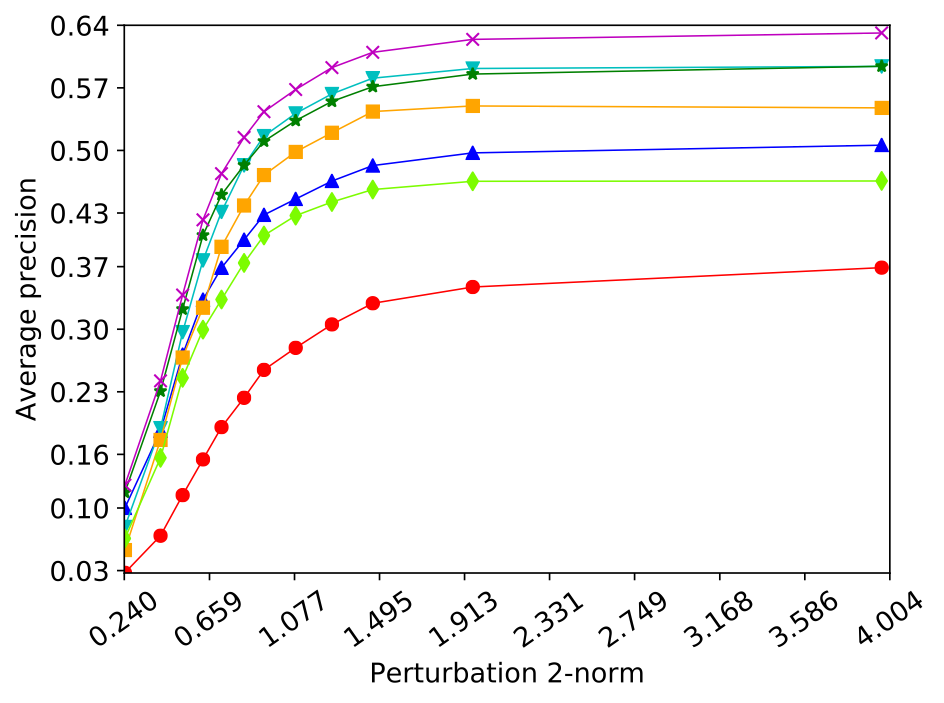}} }
\subfloat[{{\small PGD, $\epsilon = 1/255$}}]{\label{fig:avg_prec_cifar_pgd}{\includegraphics[width=0.3\linewidth]{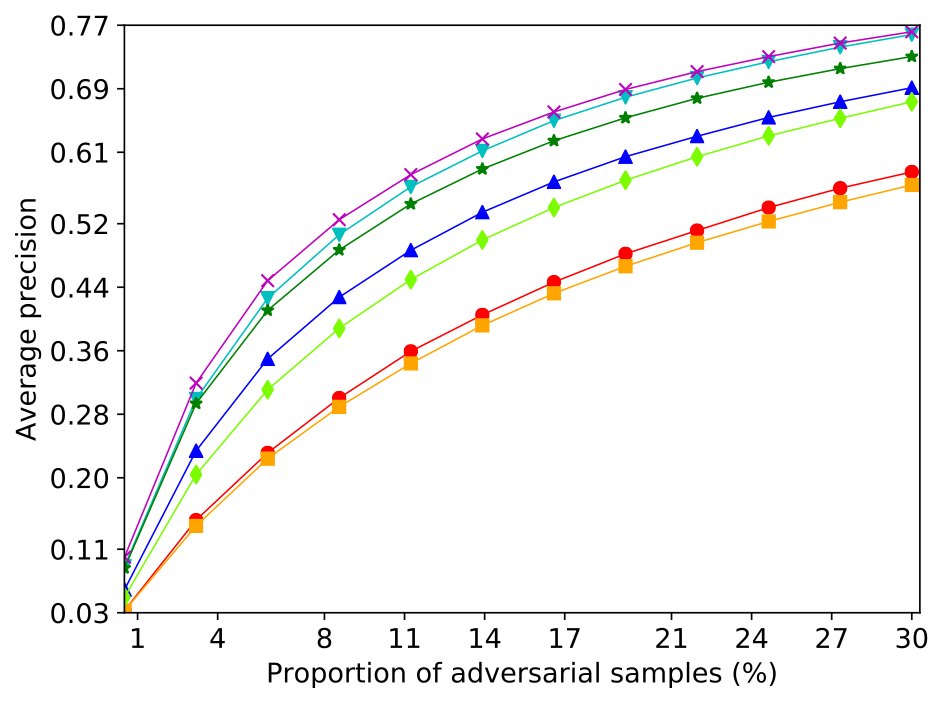}} }
\caption{\small Adversarial detection performance on CIFAR-10 under different attacks.}
\vspace{-2mm}
\label{fig:cifar10_prec}
\smallskip
\centering
\subfloat[{{\small CW, confidence $= 0$}}]{\label{fig:avg_prec_svhn_cw}{\includegraphics[width=0.3\linewidth]{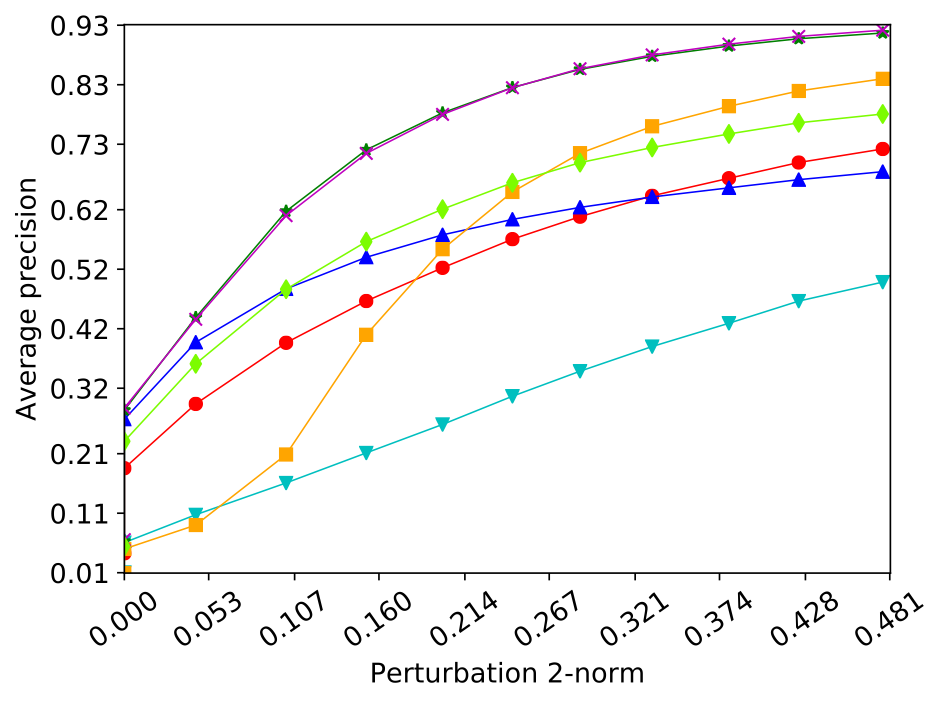}}  }
\subfloat[{{\small Adaptive attack}}]{\label{fig:avg_prec_svhn_Custom}{\includegraphics[width=0.3\linewidth]{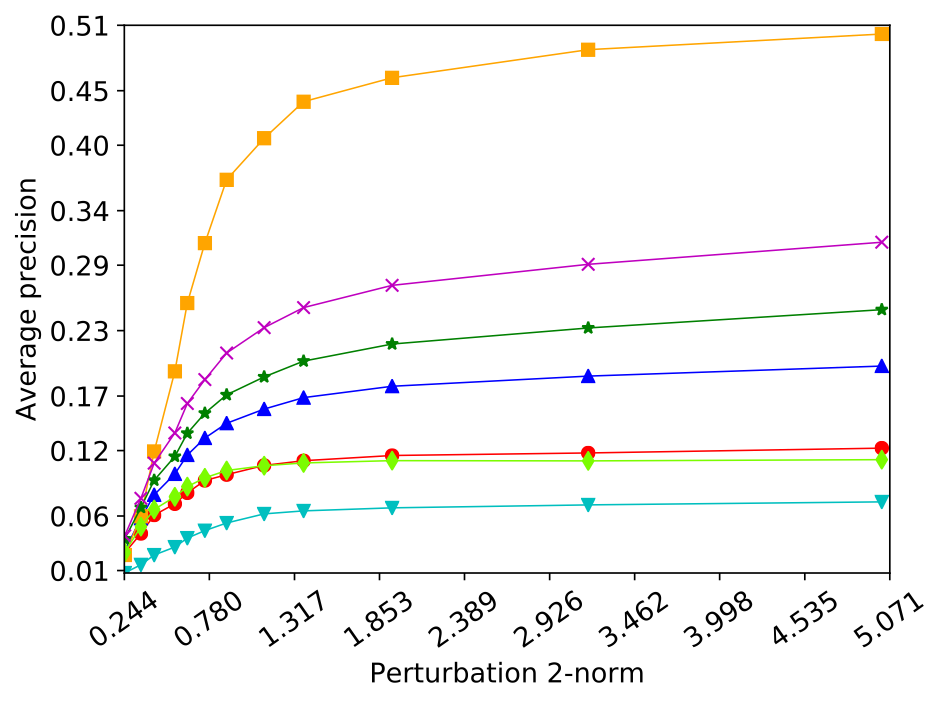}} }
\subfloat[{{\small PGD, $\epsilon = 1/255$}}]{\label{fig:avg_prec_svhn_pgd}{\includegraphics[width=0.3\linewidth]{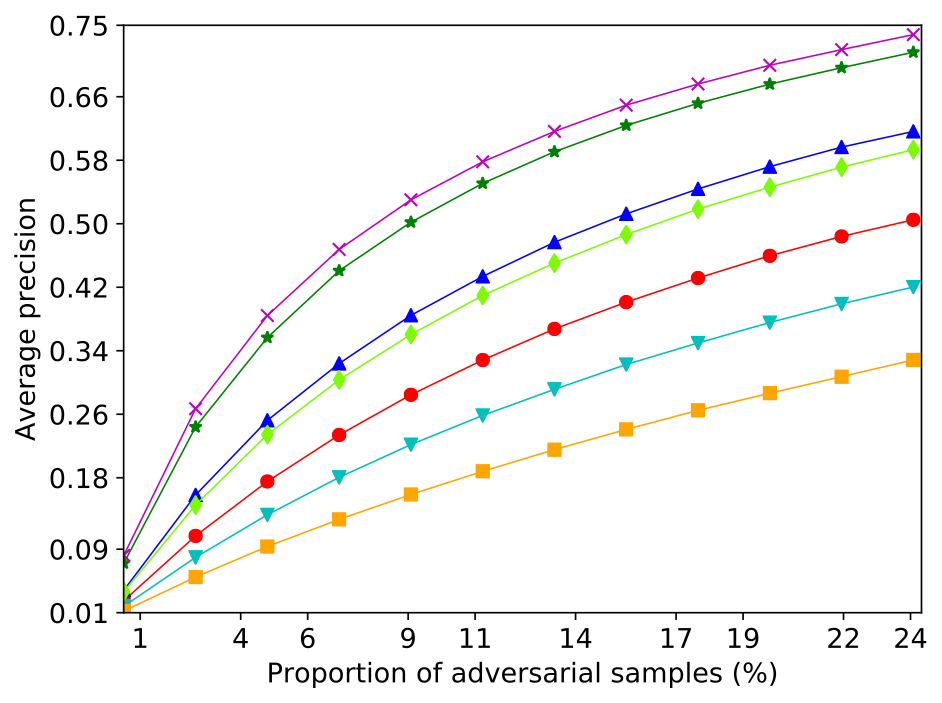}} }
\vspace{-2mm}
\caption{\small Adversarial detection performance on SVHN under different attacks.}
\vspace{-2mm}
\label{fig:svhn_prec}
\end{figure*}

We compared against the following recently-proposed methods: (i) Deep Mahalanobis detector (\dm)~\cite{lee2018simple}, (ii) Local Intrinsic Dimensionality detector (\lid)~\cite{ma2018characterizing_iclr}, (iii) The odds are odd detector (\odds)~\cite{roth2019odds}, (iv) Deep kNN (\dknn)~\cite{papernot2018deep}, and (v) Trust Score (\trust)~\cite{jiang2018trust}. \dm and \lid are supervised (they utilize adversarial or outlier data from the training folds), while the remaining methods are unsupervised. 
\lid and \odds are excluded from the OOD detection experiment because they specifically address adversarial samples.
Details on the implementation, hyperparameters, and layer representations used by the methods can be found in Appendix~\ref{app:code}.

\noindent{\bf Performance metrics.}
We evaluate detection performance using the precision-recall (PR) curve~\cite{davis2006relationship, flach2015precision} and the receiver operating characteristic (ROC) curve~\cite{fawcett2006introduction}. We use {\em average precision} as a threshold-independent metric to summarize the PR curve, and partial area under the ROC curve below FPR $\alpha$ ({\em pAUC-$\alpha$}) as the metric to summarize low-FPR region of the ROC curve.
Note that both the metrics do not require the selection of a threshold.
We do not report the area under the entire ROC curve because it is skew-insensitive and tends to have optimistic values when the fraction of anomalies is very small~\cite{ahmed2020semantic}.

\subsection{Detecting Adversarial Samples}
\label{detection_adv}
Figures~\ref{fig:cifar10_prec} and \ref{fig:svhn_prec} show the average precision of the detection methods as a function of the perturbation $\ell_2$ norm of the adversarial samples generated by the CW (confidence $= 0$) and adaptive attack methods. 
For the PGD attack ($\epsilon = 1 \,/\, 255$), the proportion of adversarial samples is shown on the x-axis instead of the perturbation norm because most of the samples from this attack have the same norm value. 
We observe that in almost all cases, \proposed outperforms the other baselines. Methods such as \dm, \odds, and \dknn perform well in some cases but fail on others, while \lid performs poorly in nearly all scenarios. 
We observe an outlying trend in Figure~\ref{fig:avg_prec_svhn_Custom}, where \odds outperforms \proposed on the adaptive attack applied to SVHN. However, a comparison of the pAUC-$0.2$ metric for this scenario (Figure~\ref{fig:svhn_pauc} in Appendix~\ref{app:pauc_results}) reveals that \proposed has higher pAUC-$0.2$ for low perturbation norm (where adversarial samples are likely to be more realistic and harder to detect). 
We provide additional results in Appendix~\ref{app:additional} that include: (i) attack transfer and attacks of varying strength, (ii) evaluation of the pAUC-$0.2\,$ metric, (iii) results on the MNIST dataset, and (iv) results on the FGSM attack.

\subsection{Detecting Out-Of-Distribution Samples}
\label{detection_ood}
\begin{figure*}[htb]
\vspace{-2mm}
\includegraphics[width=0.6\linewidth]{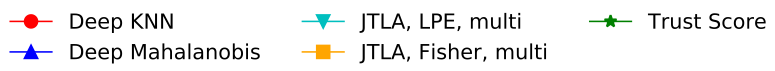}
\vspace{-3mm}
\centering
\subfloat[{{\small CIFAR-10 vs. SVHN}}]{\label{fig:cifar_vs_svhn}{\includegraphics[width=0.32\linewidth]{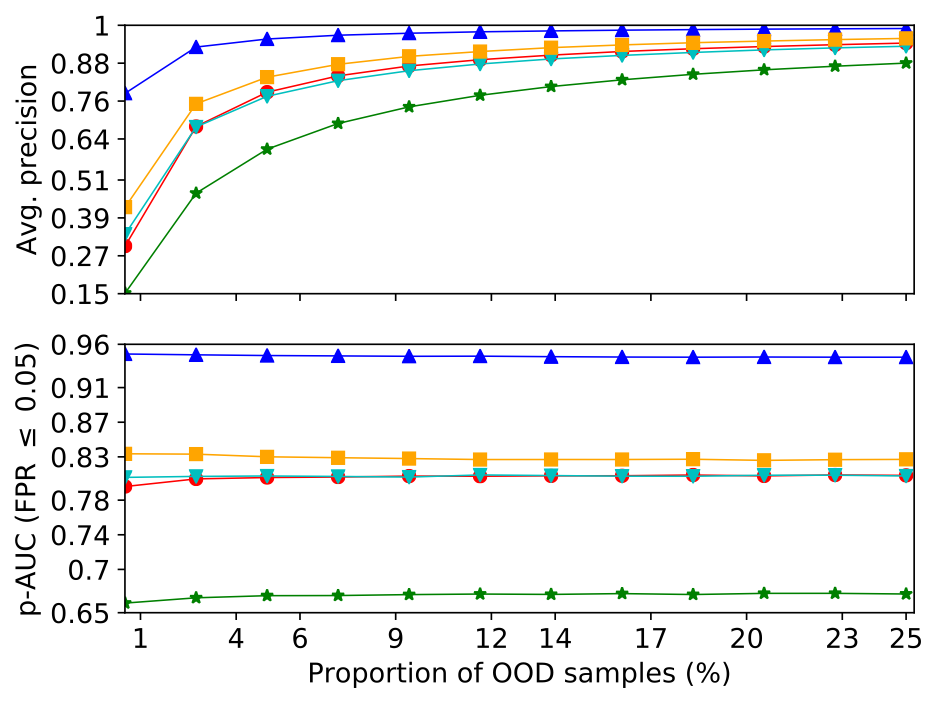}} }
\subfloat[{{\small MNIST vs. not-MNIST}}]{\label{fig:mnist_vs_not}{\includegraphics[width=0.32\linewidth]{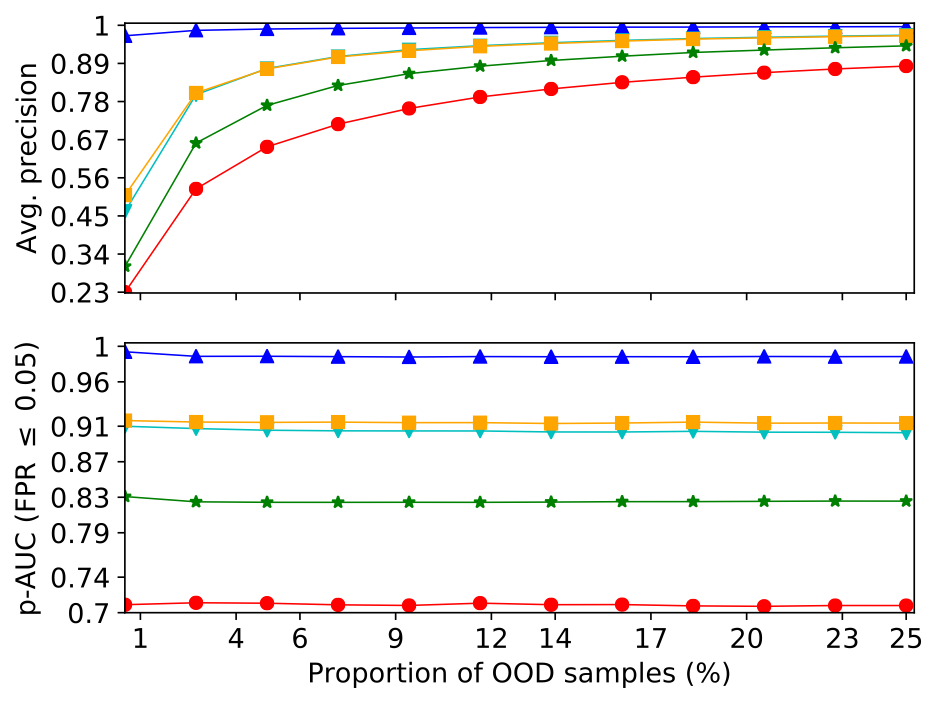}} }
\subfloat[{{\small CIFAR-10 vs. CIFAR-100}}]{\label{fig:cifar10_vs_cifar100}{\includegraphics[width=0.32\linewidth]{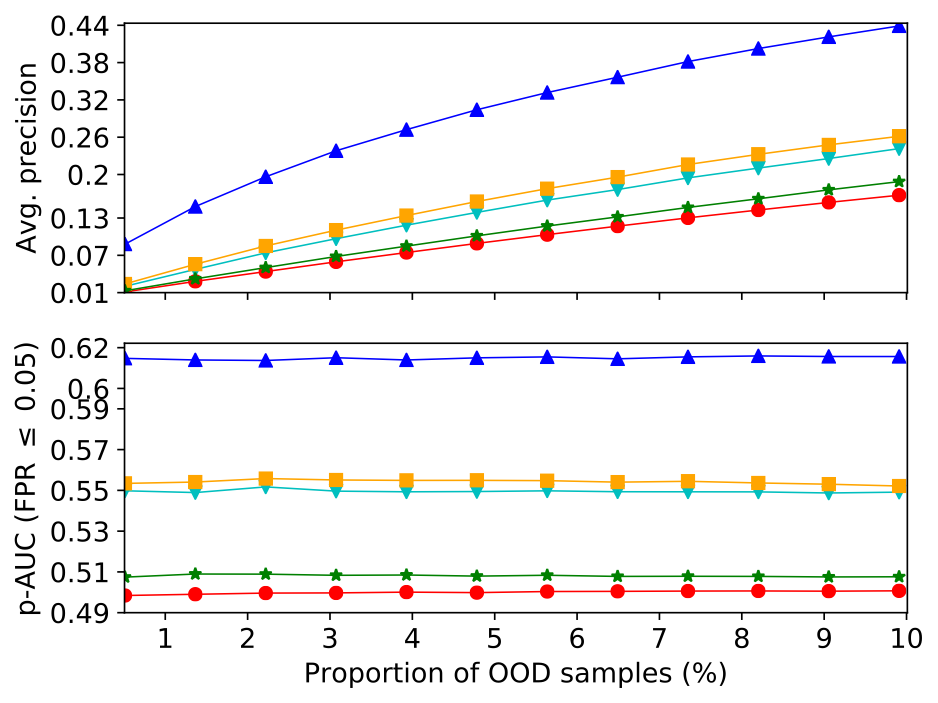}} }
\caption{\small Comparison of OOD detection performance.}
\vspace{-2mm}
\label{fig:ood_results}
\end{figure*}
We evaluated OOD detection using the following image dataset pairs, with the first dataset used as in-distribution (inliers) and the second dataset used as out-distribution (outliers): 1) MNIST vs. Not-MNIST~\cite{bulatov2011notmnist}, 2) CIFAR-10 vs. SVHN.
While a majority of papers on OOD detection evaluate using such dataset pairs, the importance of evaluating against outliers that are semantically meaningful has been emphasized by \citet{ahmed2020semantic}.
Therefore, we performed an experiment where object classes from the CIFAR-100 dataset (not in CIFAR-10) are treated as outliers relative to CIFAR-10.
This is a more realistic and challenging task since novel object categories can be considered semantically-meaningful anomalies.
Using the same 5-fold cross-validation setup, we compared the performance of \proposed with \dm, \dknn, and \trust (\odds and \lid are excluded because they focus on adversarial examples).
Since \dm is a supervised method, it uses both inlier and outlier data from the training folds, while the remaining methods (all unsupervised) use only the inlier data from the training folds. 
To promote fairness in the comparison, we excluded outlier data corresponding to one half of the classes from the training folds, and included outlier data from {\em only} the excluded classes in the test folds. 
Additionally, in the test folds we included image samples with random pixel values uniformly selected from the same range as valid images. 
The number of random samples is set equal to the average number of test-fold samples from a single class.
Figure~\ref{fig:ood_results} shows the average precision and pAUC-$0.05$ as a function of the proportion of OOD samples on the OOD detection tasks~\footnote{We report pAUC below $5\%$ FPR because the methods achieve high detection rates at very low FPR.}.
We observe that \proposed outperforms the unsupervised methods \dknn and \trust in all cases, but does not achieve the very good performance of \dm.
This should be considered in light of the fact that \dm uses outlier samples from the training folds to train a classifier and tune a noise parameter~\cite{lee2018simple}. 
However, in real-world settings, one is unlikely to have the prior knowledge and sufficient number (and variety) of outlier samples for training.

\vspace{-2mm}
\subsection{Ablation Studies}
\label{sec:exp_ablation}
We performed ablation studies to gain a better understanding of the different components of the proposed method. Specifically, we evaluated 1) the relative performance of the proposed p-value based normalization and aggregation methods, 2) the performance improvement from testing at layer pairs in addition to the individual layers, 3) the relative performance of using only the last few layers compared to using all the layer representations, and 4) the relative performance of the two scoring methods in \S~\ref{sec:scoring}. 
These results are discussed in Appendix~\ref{app:ablation}.

\vspace{-2mm}
\section{Conclusions}
\label{sec:conclusion}
We presented \proposed, a general framework for detecting anomalous inputs to a DNN classifier based on joint statistical testing of its layer representations. 
We presented a general meta-algorithm for this problem, and proposed specific methods for realizing the components of this algorithm in a principled way. 
The construction of \proposed is modular, allowing it to be used with a variety of test statistics proposed in prior works. 
Extensive experiments with strong adversarial attacks (including an adaptive defense-aware attack we proposed) and anomalous inputs to DNN image classifiers demonstrate the effectiveness of our method.

\vspace{-2mm}
\section*{Acknowledgements}
We thank the anonymous reviewers for their useful feedback that helped improve the paper.
VC, JR, and SB were supported in part through the following US NSF grants: CNS-1838733, CNS-1719336, CNS-1647152, CNS-1629833, CNS-1942014, CNS-2003129, and an award from the US Department of Commerce with award number 70NANB21H043.
SJ was partially supported by Air Force Grant FA9550-18-1-0166, the NSF Grants CCF-FMitF-1836978, SaTC-Frontiers-1804648 and CCF-1652140, and ARO grant number W911NF-17-1-0405.


\bibliography{references}
\bibliographystyle{icml2021}

\clearpage
\newpage
\appendix
\section*{Appendix}
\begin{table*}[htb]
\captionsetup{font=small,skip=10pt}
\centering
\caption{Notations and Definitions}
\resizebox{0.95\textwidth}{!}{%
\begin{tabular}{@{} l l @{}}
\toprule
\multicolumn{1}{ c }{{\bf Term}} & \multicolumn{1}{c }{{\bf Description}} \\ 
\midrule
\midrule
$[m] \,:=\, \{1, \cdots, m\}$ & Set of classes. \\
$\mathcal{L} = \{0, \cdots, L\}$, $\,\mathcal{L}^2 = \{(\ell_1, \ell_2) \in \mathcal{L} \times \mathcal{L} \,:\, \ell_2 > \ell_1\}$ & Set of layers and distinct layer pairs.\\
$\|\cdot\|_p$ & $\ell_p$ norm of a vector. \\
$\,d(\bfx, \bfy)\,$ & Distance between a pair of vectors; cosine distance unless specified otherwise. \\
$\indicator[\cdot]$ & Indicator function mapping an input condition to $1$ if true and $0$ if false. \\
$\,h_{\sigma}(\bfx, \bfy) \,=\, e^{-\frac{1}{\sigma^2} \,d(\bfx, \bfy)^2}\,$ & Gaussian or RBF kernel. \\
$\mathbf{F}(\bfx) \,=\, [F_1(\bfx), \cdots, F_m(\bfx)], ~\mathbf{F} : \mathcal{X} \mapsto \Delta_m$ & Input-output mapping learned by the DNN classifier. \\
$\Delta_m \,=\, \{(p_1, \cdots, p_m) \in [0, 1]^m \,:\, \sum_i p_i = 1\}$ & Space of output probabilities for the $m$ classes. \\
$\widehat{C}(\bfx) \,=\, \argmax_{c \in [m]} F_c(\bfx)$ & Class prediction of the DNN. \\
$\bfx^{(\ell)} \,:=\, \mathbf{f}_\ell(\bfx) \in \reals^{d_\ell}, ~\ell = 0, 1, \cdots, L$ & Flattened layer representations of the DNN ($\ell = 0$ refers to the input).\\
$C$ and $\widehat{C}$ & Random variables corresponding to the true class and DNN-predicted class. \\
$\,\mathcal{D}_a = \{(\bfx^{(0)}_n, \cdots, \bfx^{(L)}_n, c_n, \hat{c}_n), ~n \,=\, 1, \cdots, N\}\,$ & Labeled dataset with the layer representations, true class, and predicted class. \\
$\widehat{\mathcal{D}}_a(\ell, \hat{c})\,$ & Subset of $\mathcal{D}_a$ corresponding to layer $\ell$ and predicted class $\hat{c}$. \\
$\,\mathcal{D}_a(\ell, c)$ & Subset of $\mathcal{D}_a$ corresponding to layer $\ell$ and true class $c$. \\
\vspace{1mm}
$T(\bfx^{(\ell)}, \hat{c}, \widehat{\mathcal{D}}_a(\ell, \hat{c})) \,=\, T^{(\ell)}_{p \cond \hat{c}}$ & Test statistic from layer $\ell$ conditioned on the predicted class $\hat{c}$. \\
$T(\bfx^{(\ell)}, c, \mathcal{D}_a(\ell, c)) \,=\, T^{(\ell)}_{s \cond c}, ~c \in [m]$ & Test statistic from layer $\ell$ conditioned on a candidate true class $c$. \\
$\bft_{p \cond \hat{c}} \,=\, [t^{(0)}_{p \cond \hat{c}}, \cdots, t^{(L)}_{p \cond \hat{c}}]$ & Vector of test statistics from the layers conditioned on the predicted class $\hat{c}$. \\
$\bft_{s \cond c} \,=\, [t^{(0)}_{s \cond c}, \cdots, t^{(L)}_{s \cond c}], ~c \in [m]$ & Vector of test statistics from the layers conditioned on a candidate true class $c$. \\
\vspace{1mm}
$q^{}_{\textrm{agg}}(\bft_{p \cond \hat{c}}), q^{}_{\textrm{agg}}(\bft_{s \cond 1}), \cdots, q^{}_{\textrm{agg}}(\bft_{s \cond m})$ & Aggregate normalized test statistic from the predicted class and candidate true classes. \\
\vspace{1mm}
$(k^{(\ell)}_1, \cdots, k^{(\ell)}_m) \in \{0, 1, \cdots, k\}^m\,$ s.t. $\,\sum_i k^{(\ell)}_i = k$ & Class counts from the $k$-nearest neighbors of a representation vector from layer $\ell$. \\
$N^{(\ell)}_k(\bfx^{(\ell)}) \,\subset\, \{1, \cdots, N\}$ & Index set of the k-nearest neighbors of a layer representation $\bfx^{(\ell)}$ relative to $\mathcal{D}_a$. \\ 
\bottomrule
\end{tabular}%
}
\label{tab:notations_app}
\vspace{-2mm}
\end{table*}
%
\begin{algorithm}[htb]
	\caption{Meta-algorithm for Anomaly Detection}
	\label{meta_algorithm_detailed}
	\algsetup{linenosize=\small}
    \small
	\begin{algorithmic}
	    \STATE {\bfseries Inputs:} Trained DNN $\mathbf{F}(\cdot)$, Dataset $\mathcal{D}$, Test input $\bfx$, \\FPR $\alpha$ or detection threshold $\tau$.
		\STATE {\bfseries Output:} Detector decision -- normal $0$ or anomaly $1$.
		\medskip
		\STATE {\bf Preprocessing:}
		\STATE Calculate the detection threshold $\tau$ (if not specified).
		\STATE Class prediction of $\,\bfx$: $\hat{c} = \argmax_{c \in [m]} F_c(\bfx)$.
		\STATE Layer representations of $\,\bfx$: $\bfx^{(\ell)} = \mathbf{f}_\ell(\bfx), \,\forall \ell \in \mathcal{L}$.
		\STATE Data subsets: $\widehat{\mathcal{D}}_a(\ell, \hat{c})$ and $\mathcal{D}_a(\ell, c), \,\forall \,\ell \in \mathcal{L}, \,c \in [m]$.
		\smallskip
		\STATE {\bf I. Test statistics}:
		\begin{ALC@g}
		    \NoDo
		    \FOR{each layer $\,\ell \in \mathcal{L}$:}
		        \STATE Calculate $\,t^{(\ell)}_{p \cond \hat{c}} \,=\, T(\bfx^{(\ell)}, \hat{c}, \widehat{\mathcal{D}}_a(\ell, \hat{c}))$.
		        \STATE Calculate $\,t^{(\ell)}_{s \cond c} \,=\, T(\bfx^{(\ell)}, c, \mathcal{D}_a(\ell, c)), ~\forall c \in [m]$.
            \ENDFOR 
            \smallskip
            \STATE Compile the vectors: $\bft^{}_{p \cond \hat{c}}, \bft^{}_{s \cond 1}, \cdots, \bft^{}_{s \cond m}$.
        \end{ALC@g}
        \smallskip
        \STATE {\bf II. Normalizing transformations}:
        \begin{ALC@g}
            \NoThen
            \IF{multivariate normalization:}
                \STATE Calculate $\,q(\bft_{p \cond \hat{c}}), q(\bft_{s \cond 1}), \cdots, q(\bft_{s \cond m})$.
            \ELSE
                \NoDo
                \FOR{each layer $\,\ell \in \mathcal{L}$:}
                    \STATE Calculate $\,q(t^{(\ell)}_{p \cond \hat{c}}), q(t^{(\ell)}_{s \cond 1}), \cdots, q(t^{(\ell)}_{s \cond m})$.
                \ENDFOR
                \smallskip
                \OptDo
                \FOR{each layer pair $\,(\ell_1, \ell_2) \in \mathcal{L}^2$:}
                    \STATE Calculate $\,q(t^{(\ell_1)}_{p \cond \hat{c}}, t^{(\ell_2)}_{p \cond \hat{c}})$.
                    \STATE Calculate $\,q(t^{(\ell_1)}_{s \cond 1}, t^{(\ell_2)}_{s \cond 1}), \cdots, q(t^{(\ell_1)}_{s \cond m}, t^{(\ell_2)}_{s \cond m})$.
                \ENDFOR
            \ENDIF
        \end{ALC@g}
        \smallskip
        \STATE {\bf III. Layerwise aggregation and scoring}:
        \begin{ALC@g}
            \NoThen
            \IF{multivariate normalization:}
                \STATE Set $\,q^{}_{\textrm{agg}}(\bft_{p \cond \hat{c}}) \,=\, q(\bft_{p \cond \hat{c}})$.
                \STATE Set $\,q^{}_{\textrm{agg}}(\bft_{s \cond c}) \,=\, q(\bft_{s \cond c}), ~\forall c \in [m]$.
            \ELSE
                \STATE Create the sets $\,Q_{p \cond \hat{c}}\,$ and $\,Q_{s \cond c}, ~\forall c \in [m]$.
                \STATE Calculate $\,q^{}_{\textrm{agg}}(\bft_{p \cond \hat{c}}) \,=\, r(Q_{p \cond \hat{c}})\,$.
                \STATE Calculate $\,q^{}_{\textrm{agg}}(\bft_{s \cond c}) \,=\, r(Q_{s \cond c}), ~\forall c \in [m]$.
            \ENDIF
            \smallskip
            \STATE Calculate the final score:\\$S(q^{}_{\textrm{agg}}(\bft_{p \cond \hat{c}}), q^{}_{\textrm{agg}}(\bft_{s \cond 1}), \cdots, q^{}_{\textrm{agg}}(\bft_{s \cond m}), \hat{c})\,$.
        \end{ALC@g}
        \smallskip
        \STATE {\bf IV. Detection decision}:
        \begin{ALC@g}
            \NoThen
            \IF{$S(q^{}_{\textrm{agg}}(\bft_{p \cond \hat{c}}), q^{}_{\textrm{agg}}(\bft_{s \cond 1}), \cdots, q^{}_{\textrm{agg}}(\bft_{s \cond m}), \hat{c}) \,\geq\, \tau$:}
                \STATE Return anomaly ($1$).
            \ELSE
                \STATE Return normal ($0$).
            \ENDIF
        \end{ALC@g}
	\end{algorithmic}
\end{algorithm}
%
\begin{figure*}[thb]
\vspace{-2mm}
  \centering
  \includegraphics[height=0.36\textheight]{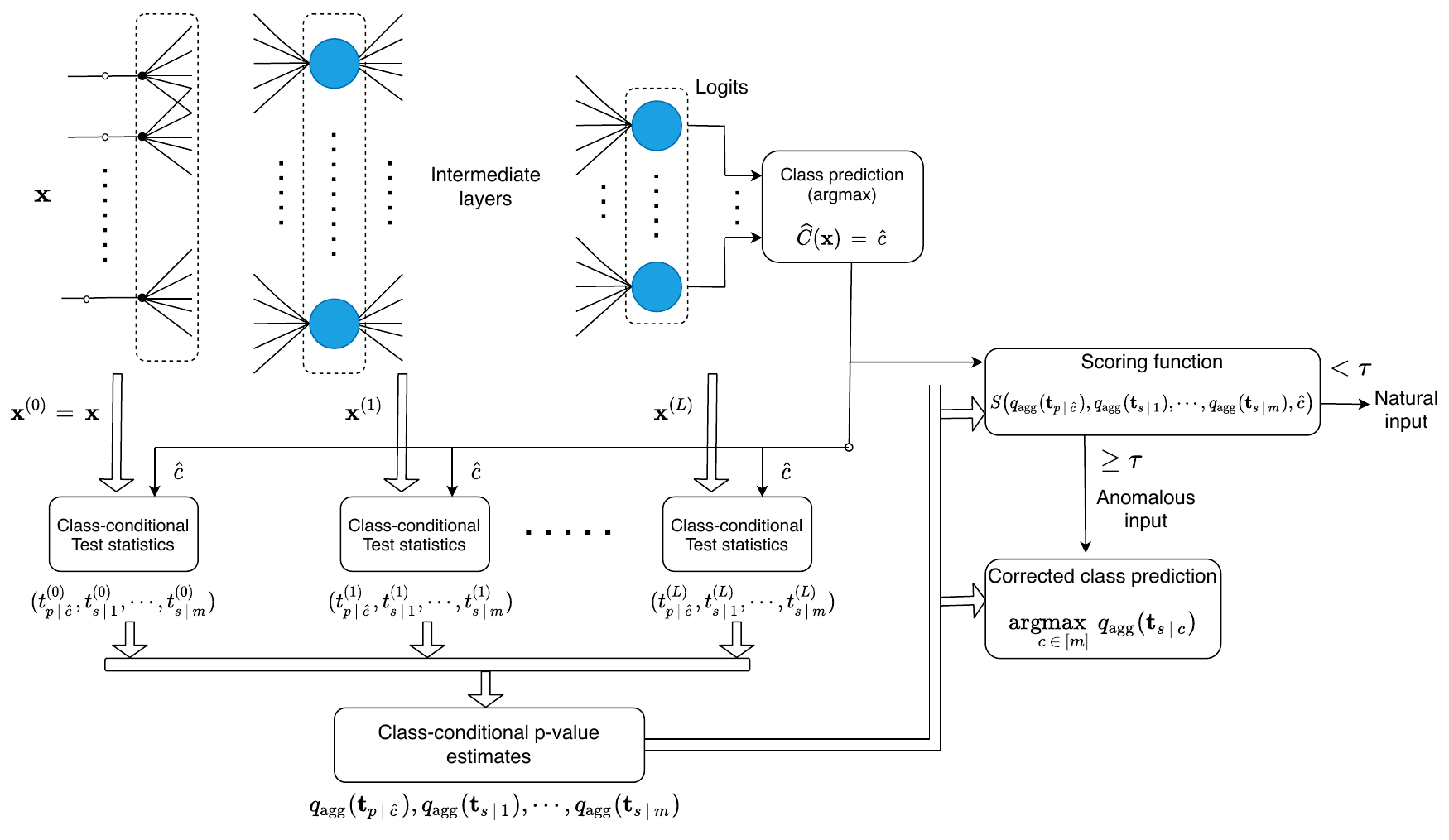}
  \caption{\small Overview of the proposed detection framework. The DNN classifies a test input $\mathbf{x}$ into one of the $m$ classes. \proposed uses the predicted class and the layer representations to calculate $m + 1$ class-conditional test statistics at each layer. The test statistics from the layers are aggregated into normalized p-value estimates, which are then used by the scoring function to determine whether the input is natural or anomalous. 
  }
\vspace{-2mm}
\label{fig:overview}
\end{figure*}
The appendices are organized as follows.
\begin{itemize}[leftmargin=*, topsep=1pt, noitemsep]
    \item Appendix \ref{sec:app_background} provides a background on adversarial attacks and defenses, and discusses prior works in more detail.
    \item Appendix \ref{sec:hmp_method} describes the harmonic mean p-value (HMP) method for aggregating p-values.
    \item Appendix \ref{sec:app_custom_attack} discusses details of the adaptive attack method that were left out of \S~\ref{sec:dknn_attack}.
    \item Appendix \ref{app:datasets} discusses the implementation details such as computing platform, datasets, DNN architectures, experimental setup, and the method implementation details, hyperparameters etc.
    \item Appendix \ref{app:additional} discusses additional experiments that we performed including ablation studies, performance on attack transfer and attacks of varying strength, and a comparison of running times.
\end{itemize}

\section{Background and Related Works}
\label{sec:app_background}
We provide a background on adversarial attacks and defense methods, followed by a detailed discussion of related works.

\vspace{-2mm}
\subsection{Adversarial Attacks}
\label{sec:adversarial_attacks}
Adversarial attacks may be broadly classified into training-time or data poisoning attacks, and test-time or evasion attacks~\cite{biggio2018wild}. Data poisoning attacks focus on tampering with the training set of a learning algorithm by introducing malicious input patterns that steer the learning algorithm to a sub-optimal solution, causing degradation in performance~\cite{biggio2012poisoning, gonzalez2017towards, steinhardt2017certified}. Evasion attacks focus on tampering with test inputs to an already-trained ML model such that the model predicts incorrectly on them. In both cases, an adversary aims to modify the inputs in such a way that the changes are not easily perceived by a human or flagged by a detector. For example, a test image of the digit $1$ may be modified by introducing minimal perturbations to the pixels such that a classifier is fooled into classifying it as the digit $7$, while the image still looks like the digit $1$ to a human. In this work, we focus on detecting test-time evasion attacks.

Given a test input $\,\bfx \in \mathcal{X}\,$ from class $c$, adversarial attack methods aim to create a minimally-perturbed input $\,\bfx^\prime = \bfx + \bfdelta\,$ that is mis-classified by the classifier either into a specific class (targeted attack) or into any class other than $c$ (untargeted attack). This is formulated as an optimization problem, which in its most general form looks like
\vspace{-1mm}
\begin{align*}
&\displaystyle\min_{\bfdelta} ~\|\bfdelta\|_p ~~~\text{s.t.} \\
&\bfx + \bfdelta \,\in\, \mathcal{X} \\
&\widehat{C}(\bfx + \bfdelta) ~=~ c^\prime ~~~\text{(targeted)} \\
\text{or }~&\widehat{C}(\bfx + \bfdelta) ~\neq~ c ~~~\text{(untargeted)} 
\end{align*}
A number of adversarial attack methods have been proposed based on this general formulation~\cite{szegedy2013intriguing, goodfellow2015explain, madry2018towards, carlini2017towards, Moosavi2016deepfool, papernot2016limitations, kurakin2017adversarial, moosavi2017universal}. Some of the well-known methods for generating adversarial data include the fast gradient sign method (FGSM)~\cite{goodfellow2015explain}, projected gradient descent (PGD) attack~\cite{madry2018towards}, Carlini-Wagner attack~\cite{carlini2017towards}, and DeepFool attack~\cite{Moosavi2016deepfool}. These attack methods can be categorized into {\it white-box} or {\it black-box} depending on the extent of their knowledge about the classifier's structure, parameters, loss function, and learning algorithm. Most of the commonly used test-time attacks on DNN classifiers are strongly white-box in that they assume a complete knowledge of the system. For a detailed discussion and taxonomy of adversarial attack methods, we refer readers to the recent surveys~\cite{akhtar2018survey, yuan2019adversarial}.

\subsection{Adversarial Defenses}
\label{sec:adversarial_defenses}
On the defense side of adversarial learning, the focus can be broadly categorized into (1) {\em adversarial training} - where the objective is to train robust classifiers that generalize well in the face of adversarial attacks~\cite{madry2018towards, fawzi2016robustness, fawzi2018analysis, goodfellow2015explain}, (2) {\em adversarial detection} - where the objective is to detect inputs that are adversarially manipulated by identifying signatures or patterns that make them different from natural inputs seen by the classifier at training time~\cite{feinman2017detecting, xu2017feature1, lee2018simple, ma2018characterizing_iclr}. One approach to adversarial training involves augmenting the training set of the classifier with adversarial samples created from one or multiple attack methods along with their true labels, and retraining the classifier on the augmented training set (possibly initialized with parameters from a prior non-adversarial training)~\cite{goodfellow2015explain}. A limitation of this approach is that the resulting classifier may still fail on attacks that were not seen by the classifier during training. This has lead to research in the direction of robust optimization, where the learning objective of the classifier (usually an empirical risk function) is modified into a min-max optimization problem, with the inner maximization performed on a suitably chosen perturbation set (\eg an $\ell_\infty$ norm ball)~\cite{madry2018towards}. Adversarial detection, on the other hand, does not usually involve special training or modification of the underlying classifier to predict accurately on adversarial inputs. Instead, the focus is on methods that can detect adversarial inputs while operating at low false positive (natural inputs detected as adversarial) rates using ideas from the anomaly detection literature. The detector flags inputs that are suspicious and likely to be misclassified by the classifier so that they may be analyzed by an expert (possibly another ML system) for decision making down the line. There have been a plethora of works on the defense side adversarial learning. Recent surveys on this topic can be found in \cite{biggio2018wild, miller2020adversarial}.

\subsection{Adversarial Samples as Anomalies}
\label{sec:adversarial_anomalies}
Adversarial detection is closely related to the problem of anomaly or outlier detection~\cite{chandola2009anomaly, zhao2009anomaly} with some important distinctions. The objective of anomaly detection is to determine if an input follows a pattern that is significantly different from that observed on a given data set. Stated differently, an input is said to be anomalous if it has a very low probability under the reference marginal distribution $\,p_0(\bfx)\,$ underlying a given data set. On the other hand, adversarial inputs from a test-time attack are not necessarily anomalous with respect to the marginal data distribution because of the way they are created by minimally perturbing a valid input $\,\bfx \sim p_0(\cdot \cond c)\,$ from a given class $c$. It is useful to consider the following notion of adversarial inputs. Suppose a clean input $\bfx$ from class $c$ (\ie $\,\bfx \sim p_0(\cdot \cond c)$) is perturbed to create an adversarial input $\bfx^\prime$ that appears to be from the same class $c$ according to the true (Bayes) class posterior distribution, \ie $\,c \,=\, \argmax_k \,p_0(k \cond \bfx^\prime)$. However, it is predicted into a different class $c^\prime$ by the classifier, \ie $\,\widehat{C}(\bfx^\prime) \,=\, \argmax_k F_k(\bfx^\prime) \,=\, c^\prime$. From this standpoint, we hypothesize that an adversarial input $\bfx^\prime$ is likely to be a typical sample from the conditional distribution $\,p_0(\cdot \cond c)\,$ of the true class, while it is also likely to be an anomalous sample from the conditional distribution $\,p_0(\cdot \cond c^\prime)\,$ of the predicted class. 
We note that \cite{miller2019ada} use a similar hypothesis to motivate their detection method.

\subsection{Related Works}
\label{sec:app_related_works}
We categorize and review some closely-related prior works on adversarial and OOD detection. For works that are based on multiple layer representations of a DNN, we discuss how the methods fit into the proposed meta-algorithm for anomaly detection.

\smallskip
\noindent{\bf Supervised Methods}

%
In \cite{lee2018simple}, the layer representations of a DNN are modeled class-conditionally using multivariate Gaussian densities, which leads to the Mahalanobis distance confidence score being used as a test statistic (feature) at the layers. The feature vector of Mahalanobis distances from the layers is used to train a binary logistic classifier for discriminating adversarial (or OOD samples) from natural samples. While this method uses the class-conditional densities of the layer representations, it uses a rather simple parametric model based on multivariate Gaussians, which may not be suitable for the intermediate layer representations (issues include high dimensionality, non-negative activations, and multimodality). Also, this method does not use the predicted class of the DNN; instead it finds the ``closest'' class corresponding to each layer representation. In the context of the meta-algorithm, this method does not explicitly apply a normalizing transformation to the test statistics. However, the Mahalanobis distance can be interpreted as the negative-log-density, which follows the Chi-squared distribution (at each layer) under the Gaussian density assumption. Finally, we note that using a binary logistic classifier is equivalent to using a weighted linear combination of the test statistics for scoring as shown in Appendix \ref{app:logistic_scoring}.

\cite{ma2018characterizing_iclr} propose using the local intrinsic dimensionality (LID) of the layer representations as a test statistic for characterizing adversarial samples. Similar to the approach of \cite{lee2018simple}, they calculate a test statistic vector of LID estimates from the layer representations, which is used for training a logistic classifier for discriminating adversarial samples from natural samples. In the context of the meta-algorithm, this method does not calculate class-conditional test statistics. The LID is estimated based on the marginal distribution of the layer representation manifold. The LID test statistics are not normalized in any way, and use of the logistic classifier implies that this method also uses a weighted linear combination of the test statistics for scoring.

In \cite{yang2019ml}, feature attribution methods are used to characterize the input and intermediate layer representations of a DNN. They find that adversarial inputs drastically alter the feature attribution map compared to natural inputs. Statistical dispersion measures such as the inter-quartile range (IQR) and median absolute deviation (MAD) are used to quantify the dispersion in the distribution of attribution values, which are then used as test statistics (features) to train a logistic classifier for discriminating adversarial samples from natural samples. In the context of the meta-algorithm, this method does not calculate class-conditional test statistics. The test statistics based on IQR or MAD are not normalized in any way. Similar to \cite{lee2018simple} and \cite{ma2018characterizing_iclr}, this method uses a weighted linear combination of the test statistics for scoring.

\smallskip
\noindent{\bf Unsupervised Methods}

\cite{roth2019odds} show that the log-odds ratio of inputs to a classifier (not necessarily a DNN) can reveal some interesting properties of adversarial inputs. They propose test statistics based on the expected log-odds of noise-perturbed inputs from different source and predicted class pairs. 
These test statistics are z-score normalized and thresholded to detect adversarial inputs. 
Their method does not use multiple layer representations for detection.
They also propose a reclassification method for correcting the classifier's prediction on adversarial inputs. This is based on training logistic classifiers (one for each predicted class) that use the expected noise-perturbed log-odds ratio as features.

In \cite{zheng2018robust}, the class-conditional distributions of fully-connected intermediate layer representations are modeled using Gaussian mixture models~\footnote{Convolutional layers are not modeled in their approach.}. Inputs with a likelihood lower than a (class-specific) threshold under each class-conditional mixture model are rejected as adversarial. A key limitation of this method is that it is challenging to accurately model the high-dimensional layer representations of a DNN using density models such as Gaussian mixtures. 

In \cite{miller2019ada}, an anomaly detection method focusing on adversarial attacks is proposed, which in its basic form can be described as follows.
The class-conditional density of the layer representations of a DNN are modeled using Gaussian mixture models, and they are used to estimate a Bayes class posterior at each layer. The Kullback-Leibler divergence between the class posterior of the DNN (based on the softmax function) and the estimated Bayes class posterior from each layer are used as test statistics for detecting adversarial samples. 
A number of methods are proposed for combining these class-conditional test statistics from the layers (\eg maximum and weighted sum across the layers).
Their method does not apply any explicit normalization of the test statistics from the layers.

\cite{sastry2019gram} propose a method for detecting OOD samples based on analyzing the Gram matrices of the layer representations of a DNN. 
The Gram matrices of different orders (order $1$ corresponds to the standard definition) capture the pairwise correlations between the elements of a layer representation. 
At training time, their detector records the element-wise minimum and maximum values of the Gram matrices (of different orders) from a training set of natural inputs, conditioned on each predicted class. 
The extent of deviation from the minimum and maximum values observed at training time is used to calculate a class-specific deviation test statistic at each layer. The final score for OOD detection is obtained by adding up the normalized deviations from the layers, where the normalization factor for a layer is the expected deviation observed on a held-out validation dataset.
We note that the deviation statistic based on Gram matrices from the layer activations proposed in \cite{sastry2019gram} can be used as a test statistic for \proposed as well.

\noindent{\bf Confidence Metrics for Classifiers}

Works such as the trust score~\cite{jiang2018trust} and deep KNN~\cite{papernot2018deep} have explored the problem of developing a confidence metric that can be used to independently validate the predictions of a classifier. Inputs with a low confidence score are expected to be misclassified and can be flagged as potentially adversarial or OOD. 
Deep KNN~\cite{papernot2018deep} uses the class labels of the k-nearest neighbors of DNN layer representations to calculate a non-conformity score corresponding to each class. Large values of non-conformity corresponding to the predicted class indicate that an input may not have a reliable prediction. The method calculates empirical p-values of the non-conformity scores to provide a confidence score, credibility score, and an alternate (corrected) class prediction for test inputs.
We note that the deep KNN test statistic based on the kNN class counts from the predicted class $k^{(\ell)}_{\hat{c}}$ and its complement $\,k - k^{(\ell)}_{\hat{c}}\,$ can be considered as a binomial specialization of the multinomial test statistic (\S~\ref{sec:multinom_test_stat}).
The trust score~\cite{jiang2018trust} estimates the $\alpha$-high density (level) set for each class, and calculates the distances from a test point to the $\alpha$-high density sets from the classes to define a confidence metric. These methods can also be categorized as unsupervised.

\subsection{Scoring with a Logistic classifier}
\label{app:logistic_scoring}
Consider a binary logistic classifier that takes a vector of test statistics $\bft$ as input and produces an output probability for class $1$ given by
\begin{equation*}
P(Y = 1 \cond \bft) ~=~ \frac{1}{1 ~+~ \exp(-\bfw^T \,\bft \,-\, b)},
\end{equation*}
where $\bfw$ and $b$ are weight vector and bias parameters.
A detection decision of $1$ (anomaly) is made when the output probability exceeds a threshold $\tau$. It is easy to see that this results in the following decision rule:
\vspace{-1mm}
\begin{equation*}
\psi(\bfx^{(0)}, \cdots, \bfx^{(L)}, \hat{c}) ~=
\begin{cases}
1 & \mbox{ if } \,\bfw^T \,\bft \,\geq\, \log \frac{\tau}{1 - \tau} \,-\, b \\
0 & \mbox{ otherwise }
\end{cases}
\end{equation*}
In other words, the score function is a weighted linear combination of the test statistics from the layers.

\section{Harmonic Mean p-value Method}
\label{sec:hmp_method}
The harmonic mean p-value (HMP)~\cite{wilson2019harmonic} is a recently proposed method for combining p-values from a large number of dependent tests. It is rooted in ideas from Bayesian model averaging and has some desirable properties such as robustness to positive dependency between the p-values, and an ability to detect small statistically-significant groups from a large number of p-values (tests).
The main result of \cite{wilson2019harmonic} can be summarized as follows. Given a set of p-values $\,p_1, \cdots, p_m\,$ from $m$ hypothesis tests, the weighted harmonic p-value of any subset $\,\mathcal{R} \subset \{1, \cdots, m\}\,$ of the p-values is given by
\vspace{-1mm}
\begin{equation*}
p^{-1}_{\textrm{agg}} ~=~ \frac{ \mysum_{i \in \mathcal{R}} \,w_i p^{-1}_i }{ \mysum_{i \in \mathcal{R}} \,w_i },
\end{equation*}
%
where the weights $w_i$ are non-negative and satisfy $\sum_{i=1}^m w_i \,=\, 1$. 
In our problem, we apply the HMP method with the weights all set to the same value, resulting in the p-value aggregation function $r(\cdot)$ defined in Eq. (\ref{eq:pvalue_hmp}).

In our experiments, we found the HMP method to have comparable or slightly worse performance than Fisher's method of combining p-values. Results from these ablation experiments can be found in Appendix \ref{app:ablation}.

\section{Details and Extensions of the Adaptive Attack}
\label{sec:app_custom_attack}
We first discuss the method we used for setting the scale of the Gaussian kernel per layer in the adaptive attack method of \S~\ref{sec:dknn_attack}. We then discuss an untargeted variant of the proposed adaptive attack, followed by an alternate formulation for the attack objective function. 

\subsection{Setting the Kernel Scale}
\label{sec:kernel_scale}
For a given clean input $\bfx$, the scale of the Gaussian kernel for each layer $\,\sigma^{}_\ell\,$ determines the effective number of samples that contribute to the soft count that approximates the kNN counts in Eq. (\ref{eq:knn_attack_objec}).
Intuitively, we would like the kernel to have a value close to $1$ for points within a distance of $\,\eta^{}_\ell$ (the kNN radius centered on $\,\mathbf{f}_\ell(\bfx)$), and decay rapidly to $0$ for points further away. Let $\,\mathcal{N}_k(\mathbf{f}_\ell(\bfx))\,$ denote the index set of the $k$-nearest neighbors of $\,\mathbf{f}_\ell(\bfx)\,$ from the $\ell$-th layer representations of the dataset $\mathcal{D}_a$.
The probability of selecting the $k$-nearest neighbors from the set of $N$ samples in $\mathcal{D}_a$ can be expressed as
\vspace{-2mm}
\begin{equation*}
s_1(\sigma_\ell) ~=~ \frac{\mysum_{n \,\in\, \mathcal{N}_k(\mathbf{f}_\ell(\bfx))} \!\!\!h_{\sigma_\ell}(\mathbf{f}_\ell(\bfx), \mathbf{f}_\ell(\bfx_n))}{\mysum_{n=1}^N \,h_{\sigma_\ell}(\mathbf{f}_\ell(\bfx), \mathbf{f}_\ell(\bfx_n))}.
\end{equation*}
We could choose $\,\sigma_\ell\,$ to maximize this probability and push it close to $1$. However, this is likely to result in a very small value for $\,\sigma_\ell$, which would concentrate all the probability mass on the nearest neighbor. 
To ensure that the probability mass is distributed sufficiently uniformly over the $k$-nearest neighbors we add the following normalized entropy~\footnote{The entropy is divided by the maximum possible value of $\log k$ to scale it to the range $[0, 1]$.} term as a regularizer
\vspace{-2mm}
\begin{align*}
&s_2(\sigma_\ell) ~=~ -\frac{1}{\log k} \mysum_{i \,\in\, \mathcal{N}_k(\mathbf{f}_\ell(\bfx))} p_i(\sigma_\ell) \,\log p_i(\sigma_\ell), \\
&\mbox{where} \\
&p_i(\sigma_\ell) ~=~ \frac{ h_{\sigma_\ell}(\mathbf{f}_\ell(\bfx), \mathbf{f}_\ell(\bfx_i)) }{\mysum_{j \,\in\, \mathcal{N}_k(\mathbf{f}_\ell(\bfx))} \!\!h_{\sigma_\ell}(\mathbf{f}_\ell(\bfx), \mathbf{f}_\ell(\bfx_j))}.
\end{align*}
Both terms $\,s_1(\sigma_\ell)\,$ and $\,s_2(\sigma_\ell)\,$ are bounded to the interval $[0, 1]$. We find a suitable $\,\sigma_\ell\,$ by maximizing a convex combination of the two terms, \ie
\vspace{-1mm}
\begin{equation}
\label{eq:opt_kernel_scale}
\max_{\sigma_\ell \in \reals_+} \,(1 - \alpha) \,s_1(\sigma_\ell) ~+~ \alpha \,s_2(\sigma_\ell).
\end{equation}
We set $\alpha$ to $0.5$ in our experiments, and used a simple line search to find the approximate maximizer of Eq. (\ref{eq:opt_kernel_scale}).

\subsection{Untargeted Attack Formulation}
\label{sec:untargeted_custom}
The formulation in \S\,\ref{sec:dknn_attack}, where a specific class $\,c^\prime \neq c\,$ is chosen, is used to create a targeted attack. Alternatively, a simple modification to Eq. (\ref{eq:adver_obj1}) that considers the log-odds of class $c$, $\,\log\frac{p_c}{1 - p_c}\,$, can be used to create an untargeted attack, resulting in the following objective function to be minimized:
\vspace{-2mm}
\begin{align}
\label{eq:knn_attack_untar_objec}
J(\bfdelta) ~&=~ \|\bfdelta\|^2_2 ~+~ \lambda \,\log \mysum_{\ell=0}^L \mysum_{\substack{n=1\,:\\ c_n = c}}^N h_{\sigma^{}_\ell}(\mathbf{f}_\ell(\bfx + \bfdelta), \mathbf{f}_\ell(\bfx_n)) \nonumber \\ 
&-~ \lambda \,\log \mysum_{\ell=0}^L \mysum_{\substack{n=1\,:\\ c_n \neq c}}^N h_{\sigma^{}_\ell}(\mathbf{f}_\ell(\bfx + \bfdelta), \mathbf{f}_\ell(\bfx_n)).
\end{align}

\subsection{Alternate Attack Loss Function}
\label{sec:alt_loss_function_custom}
Starting from equation Eq. (\ref{eq:adver_obj1}), but now assuming that the probability estimate for each class based on the kNN model factors into a product of probabilities across the layers of the DNN (\ie independence assumption), we get
\vspace{-1mm}
\begin{align}
\log \frac{p_c}{p_{c^\prime}} ~&=~ \log \frac{k_c}{k_{c^\prime}} ~=~ \log\frac{\prod_\ell k^{(\ell)}_c \,/\, k}{\prod_\ell k^{(\ell)}_{c^\prime} \,/\, k} \nonumber \\
&=~ \mysum_{\ell=0}^L \log k^{(\ell)}_c ~-~ \mysum_{\ell=0}^L \log k^{(\ell)}_{c^\prime}. \nonumber
\end{align}
Using the soft count approximation based on the Gaussian kernel (as before) leads to the following alternative loss function for the targeted adaptive attack
\vspace{-1mm}
\begin{align}
\label{eq:knn_attack_objec_alt}
J(\bfdelta) ~&=~ \|\bfdelta\|^2_2 \,+\, \lambda\, \mysum_{\ell=0}^L \log \mysum_{\substack{n=1\,:\\ c_n = c}}^N h_{\sigma^{}_\ell}(\mathbf{f}_\ell(\bfx + \bfdelta), \mathbf{f}_\ell(\bfx_n)) \nonumber \\
&-~ \lambda\, \mysum_{\ell=0}^L \log \mysum_{\substack{n=1\,:\\ c_n = c^\prime}}^N h_{\sigma^{}_\ell}(\mathbf{f}_\ell(\bfx + \bfdelta), \mathbf{f}_\ell(\bfx_n)).
\end{align}
In contrast to Eq. (\ref{eq:knn_attack_objec}), this loss function considers the class probability estimates from each layer, instead of a single class probability estimate based on the cumulative kNN counts across the layers. 
%
A special case of the loss function Eq. (\ref{eq:knn_attack_objec_alt}) that includes only the final (logit) layer of the DNN $\,\mathbf{f}_L(\bfx)$ can be directly compared to the Carlini-Wagner attack formulation~\cite{carlini2017towards}. With this formulation, the logit layer representations of an adversarial input will be guided closer to the neighboring representations from class $c^\prime$, and away from the neighboring representations from class $c$.

\section{Additional Implementation Details}
\label{app:datasets}
\subsection{Computing Platform}
Our experiments were performed on a single server running Ubuntu 18.04 with 128\,GB memory, 4 NVIDIA GeForce RTX 2080 GPUs, and 32 CPU cores. 
\subsection{Datasets \& DNN Architectures}
A summary of the image classification datasets we used, the architecture and test set performance of the corresponding DNNs are given in Table~\ref{tab:dataset}. Each dataset has $10$ image classes. We followed recommended best practices for training DNNs on image classification problems (using techniques like Dropout). We did not train a DNN on the Not-MNIST dataset because this dataset is used only for evaluation in the OOD detection experiments.
\begin{table*}[htb]
\centering
\caption{\small Datasets \& DNN Architectures.}
\vspace{-2mm}
\resizebox{0.88\textwidth}{!}{%
\begin{tabular}{@{}llllll@{}}
\toprule
\multicolumn{1}{l}{\textbf{Dataset}}                                             & \multicolumn{1}{c}{\textbf{\begin{tabular}[l]{@{}l@{}}Input\\ dimension\end{tabular}}} & \multicolumn{1}{c}{\textbf{\begin{tabular}[l]{@{}l@{}}Train\\ size\end{tabular}}} & \multicolumn{1}{c}{\textbf{\begin{tabular}[l]{@{}l@{}}Test\\ size\end{tabular}}} & \multicolumn{1}{c}{\textbf{\begin{tabular}[l]{@{}l@{}}Test\\ accuracy (\%)\end{tabular}}} & \multicolumn{1}{c}{\textbf{\begin{tabular}[l]{@{}l@{}}DNN \\ architecture\end{tabular}}} \\ 
\midrule
MNIST~\cite{lecun1998gradient}                                                   & $28\times28\times 1$                                                                   & 50,000                                                                            & 10,000                                                                           & 99.12                                                                                & 2 Conv. + 2 FC layers~\cite{lecun1998gradient}                                                                                 \\
SVHN~\cite{netzer2011reading}                                                    & $32\times32\times3$                                                                    & 73,257                                                                            & 26,032                                                                           & 89.42                                                                                & 2 Conv. + 3 FC layers                                                                    \\
\begin{tabular}[c]{@{}l@{}}CIFAR-10\\ \cite{krizhevsky2009learning}\end{tabular} & $32\times32\times3$                                                                    & 50,000                                                                            & 10,000                                                                           & 95.45                                                                                & ResNet-34~\cite{he2016deep}                                                              \\
Not-MNIST~\cite{bulatov2011notmnist}                                             & $28\times28\times 1$                                                                   & 500,000                                                                           & 18,724                                                                           & N/A                                                                                  & N/A                                                                                      \\ 
\bottomrule
\end{tabular}%
}
\vspace{-2mm}
\label{tab:dataset}
\end{table*}

\subsection{Method Implementations}
\label{app:code}
The code associated with our work can be accessed \href{https://github.com/jayaram-r/adversarial-detection}{here} \footnote{\url{https://github.com/jayaram-r/adversarial-detection}}. We utilized the authors original implementation for the following methods: (i) deep mahalanobis detector~\footnote{\url{https://github.com/pokaxpoka/deep_Mahalanobis_detector}}, (ii) the odds are odd detector~\footnote{\url{https://github.com/yk/icml19_public}}. We implemented the remaining detection methods, viz. \lid, \dknn, and \trust, and this is available as part of our released code. All implementations are in \texttt{Python3} and are based on standard scientific computing libraries such as \texttt{Numpy}, \texttt{Scipy}, and \texttt{Scikit-learn}~\cite{pedregosa2011scikit}. We used \texttt{PyTorch} as the deep learning and automatic differentiation backend~\cite{paszke2017automatic}. We perform approximate nearest neighbor search using the \texttt{NNDescent} method~\cite{dong2011efficient} to efficiently construct and query from kNN graphs at the DNN layer representations. We used the implementation of NNDescent provided by the library \texttt{PyNNDescent} \footnote{\url{https://github.com/lmcinnes/pynndescent}}. Our implementation of \proposed is highly modular, allowing for new test statistics to be easily plugged in to the existing implementation. We provide implementations of the following test statistics: (i) multinomial class counts, (ii) binomial class counts, (iii) trust score, (iv) local intrinsic dimensionality, and (v) average kNN distance. 

In our experiments with \proposed, where Fisher's method or HMP method are used for combining the p-values, we included p-values estimated from the individual layers (listed in Tables~\ref{tab:layers_mnist}, \ref{tab:layers_svhn}, and \ref{tab:layers_cifar}) and from all distinct layer pairs. 
The number of nearest neighbors $k$ is the {\bf only hyperparameter} of the proposed method. This is set to be a function of the number of in-distribution training samples $n$ using the heuristic $\,k = \ceil{n^{0.4}}$. 
For methods like \dknn, \lid, and \trust that also depend on $k$, we found that setting $k$ in this way produces comparable results to that obtained over a range of $k$ values. Therefore, to maintain consistency, we set $k$ for all the (applicable) methods using the rule $\,k = \ceil{n^{0.4}}\,$. 

For the method \trust, we applied the trust score to the input, logit (pre-softmax) layer, and the fully-connected layer prior to the logit (pre-logit) layer. Since it was reported in \cite{jiang2018trust} that the trust score works well mainly in low-dimensional settings, we applied the same dimensionality reduction that was applied to \proposed (see Tables~\ref{tab:layers_mnist}, \ref{tab:layers_svhn}, and \ref{tab:layers_cifar}) on the input and pre-logit layer representations. We found the pre-logit layer to produce the best detection performance in our experiments. Hence, we report results for \trust with the pre-logit layer in all our experiments. The constant $\alpha$, which determines the fraction of samples with lowest empirical density to be excluded from the density level sets, is set to $0$ in our experiments. We explored a few other values of $\alpha$, but did not find a significant difference in performance. This is consistent with the observation in Section 5.3 of \cite{jiang2018trust}.

The methods \dm and \lid train a logistic classifier to discriminate adversarial samples from natural samples. The regularization constant of the logistic classifier is found by searching (over a set of $10$ logarithmically spaced values between $0.0001$ to $10000$) for the value that leads to the largest average test-fold area under the ROC curve, in a 5-fold stratified cross-validation setup.

For the method \odds, the implementation of the authors returns a binary (0 / 1) detection decision instead of a continuous score that can be used to rank adversarial samples. The binary decision is based on applying z-score normalization to the original score, and comparing it to the $99.9$-th percentile of the standard Gaussian density. Instead of using the thresholded decision, we used the z-score-normalized score of \odds in order to get a continuous score that is required by the metrics average precision and pAUC-$\alpha$.

\smallskip
\noindent{\bf Details on the DNN Layer Representations}

The DNN layers used in our experiments and their raw dimensionality are listed for the three datasets in Tables~\ref{tab:layers_mnist}, \ref{tab:layers_svhn}, and \ref{tab:layers_cifar}. An exception to this is the \lid method, for which we follow the implementation of \cite{ma2018characterizing_iclr} and calculate the LID features from all the intermediate layers. For \dknn and \lid, we did not apply any dimensionality reduction or pre-processing of the layer representations to be consistent with the respective papers. For \dm, the implementation of the authors performs global average pooling at each convolutional layer to transform a $\,C \times W \times H\,$ tensor (with $C$ channels) to a $C$-dimensional vector. 
\begin{table}[htb]
\captionsetup{font=small,skip=10pt}
\centering
\caption{\small Layer representations of the DNN trained on MNIST. The output of the block listed in the first column is used as the layer representation.}
\vspace{-1mm}
\resizebox{0.48\textwidth}{!}{%
\begin{tabular}{@{} l l l l l @{}}
\toprule
{\bf Layer block} & \begin{tabular}[l]{@{}l@{}}{\bf Layer} \\ {\bf index} \end{tabular} & \begin{tabular}[l]{@{}l@{}}{\bf Original} \\ {\bf dimension}\end{tabular} & \begin{tabular}[l]{@{}l@{}}{\bf Intrinsic} \\ {\bf dimension}\end{tabular} & \begin{tabular}[l]{@{}l@{}}{\bf Projected} \\ {\bf dimension}\end{tabular} \\ 
\midrule
Input & 0 & 784 & 13 & 31 \\
Conv-1 + ReLu & 1 & 21632 & 22 & 53 \\
Conv-2 + Maxpool + Dropout & 2 & 9216 & 18 & 77 \\
FC-1 + ReLu + Dropout & 3 & 128 & 9 & 90 \\
FC-2 (Logit) & 4 & 10 & 6 & 10 \\
\bottomrule
\end{tabular}%
}
\vspace{-2mm}
\label{tab:layers_mnist}
\end{table}

For \proposed, we use the neighborhood preserving projection (NPP) method~\cite{he2005neighborhood} to perform dimensionality reduction on the layer representations. We chose NPP because it performs an efficient linear projection that attempts to preserve the neighborhood structure in the original space as closely as possible in the projected (lower dimensional) space. Working with the lower dimensional layer representations mitigates problems associated with the curse of dimensionality, and significantly reduces the memory utilization and the running time of \proposed. The original dimension, intrinsic dimension estimate, and the projected dimension of the layer representations for the three datasets are listed in Tables~\ref{tab:layers_mnist}, \ref{tab:layers_svhn}, and \ref{tab:layers_cifar}. We did not apply dimension reduction to the logit layer because it has only $10$ dimensions.
\begin{table}[htb]
\captionsetup{font=small,skip=10pt}
\centering
\caption{\small Layer representations of the DNN trained on SVHN. The output of the block listed in the first column is used as the layer representation.}
\vspace{-2mm}
\resizebox{0.48\textwidth}{!}{%
\begin{tabular}{@{} l l l l l @{}}
\toprule
{\bf Layer block} & \begin{tabular}[l]{@{}l@{}}{\bf Layer} \\ {\bf index} \end{tabular} & \begin{tabular}[l]{@{}l@{}}{\bf Original} \\ {\bf dimension}\end{tabular} & \begin{tabular}[l]{@{}l@{}}{\bf Intrinsic} \\ {\bf dimension} \end{tabular} & \begin{tabular}[l]{@{}l@{}}{\bf Projected} \\ {\bf dimension}\end{tabular} \\ 
\midrule
Input & 0 & 3072 & 18 & 43 \\ 
Conv-1 + ReLu & 1 & 57600 & 38 & 380 \\ 
Conv-2 + ReLu + Maxpool + Dropout & 2 & 12544 & 42 & 400 \\
FC-1 + ReLu + Dropout & 3 & 512 & 12 & 120 \\
FC-2 + ReLu + Dropout & 4 & 128 & 7 & 10 \\ 
FC-3 (Logit) & 5 & 10 & 4 & 10 \\ 
\bottomrule
\end{tabular}%
}
\vspace{-2mm}
\label{tab:layers_svhn}
\end{table}

The procedure we used for determining the projected dimension is summarized as follows. We used the training partition of each dataset (that was used to train the DNN) for this task in order to avoid introducing any bias on the test partition (which is used for the detection and corrected classification experiments). At each layer, we first estimate the intrinsic dimension (ID) as the median of the LID estimates of the samples, found using the method of \cite{amsaleg2015estimating}. The ID estimate $\,d_{\textrm{int}}\,$ is used as a lower bound for the projected dimension. Using a 5-fold stratified cross-validation setup, we search over $20$ linearly spaced values in the interval $\,[d_{\textrm{int}}, 10\,d_{\textrm{int}}]\,$ for the projected dimension (found using NPP) that results in the smallest average test-fold error rate for a standard k-nearest neighbors classifier. The resulting projected dimensions for each dataset (DNN architecture) are given in Tables~\ref{tab:layers_mnist}, \ref{tab:layers_svhn}, and \ref{tab:layers_cifar}.
\begin{table}[htb]
\captionsetup{font=small,skip=10pt}
\centering
\caption{\small Layer representations of the ResNet-34 trained on CIFAR-10. The output of the block listed in the first column is used as the layer representation. Legend: RB - Residual block, BN - BatchNorm}
\vspace{-2mm}
\resizebox{0.48\textwidth}{!}{%
\begin{tabular}{@{} l l l l l @{}}
\toprule
{\bf Layer block} & \begin{tabular}[l]{@{}l@{}}{\bf Layer} \\ {\bf index} \end{tabular} & \begin{tabular}[l]{@{}l@{}}{\bf Original} \\ {\bf dimension}\end{tabular} & \begin{tabular}[l]{@{}l@{}}{\bf Intrinsic} \\ {\bf dimension} \end{tabular} & \begin{tabular}[l]{@{}l@{}}{\bf Projected} \\ {\bf dimension}\end{tabular} \\ 
\midrule
Input & 0 & 3072 & 25 & 48 \\ 
Conv-1 + BN + ReLU & 1 & 65536 & 33 & 330 \\
RB-1 & 2 & 65536 & 58 & 580 \\ 
RB-2 & 3 & 32768 & 59 & 590 \\ 
RB-3 & 4 & 16384 & 28 & 28 \\ 
RB-4 & 5 & 8192 & 15 & 15 \\ 
2D Avg. Pooling & 6 & 512 & 9 & 9 \\ 
FC (Logit) & 7 & 10 & 8 & 10 \\ 
\bottomrule
\end{tabular}%
}
\vspace{-2mm}
\label{tab:layers_cifar}
\end{table}
%

\noindent{\bf Note on Performance Calculation}

The following procedure is used to calculate performance metrics as a function of the perturbation norm of adversarial samples. Suppose there are $N_a$ adversarial samples and $N_n$ natural samples in a test set, with $\,N = N_a + N_n$. Define the maximum proportion of adversarial samples $\,p_a \,=\, N_a \,/\, N$. The adversarial samples are first sorted in increasing order of their perturbation norm. The proportion of adversarial samples is varied over $12$ equally-spaced values between $0.005$ and $\,\min(0.3, p_a)$. For a given proportion $p_i$, the top $\,N_i \,=\, \ceil{p_i \,N}\,$ adversarial samples (sorted by norm) are selected. The perturbation norm of all these adversarial samples will be below a certain value; this value is shown on the x-axis of the performance plots. The performance metrics (average precision, pAUC-$\alpha$ etc.) are then calculated from the $N_i$ adversarial samples and the $N_n$ natural samples.

In order to calculate the performance metrics as a function of the proportion of adversarial or OOD (anomalous) samples, the only difference is that we do not sort the anomalous samples in a deterministic way. For a given proportion $p_i$, we select $\,N_i \,=\, \ceil{p_i \,N}\,$ anomalous samples at random uniformly, and calculate the performance metrics based on the $N_i$ anomalous and $N_n$ natural samples. To account for variability, this is repeated over $100$ random subsets of $N_i$ anomalous samples each, and the median value of the performance metrics is reported. Note that the detection methods need to score the samples only once, and the above calculations can be done based on the saved scores.

\subsection{Details on the Adversarial Attacks}
\label{app:attacks}
We list below the parameters of the adversarial attack methods we implemented using Foolbox~\cite{rauber2017foolbox}. We utilize the same variable names used by Foolbox in order to enable easy replication.
\begin{itemize}[leftmargin=*]
\item PGD attack~\cite{madry2018towards}: $\ell_\infty$ norm with the norm-ball radius $\epsilon$ linearly spaced in the interval $\,[\frac{1}{255}, \frac{21}{255}]$. Using the notation of the Foolbox library:\\
\texttt{epsilon = [$\frac{1}{255},\frac{3}{255},\cdots,\frac{21}{255}$], stepsize = 0.05, binary\_search = False, iterations = 40, p\_norm = inf}.
\item Carlini-Wagner (CW) attack~\cite{carlini2017towards}: $\ell_2$ norm with the confidence parameter of the attack varied over the set $\{0, 6, 14, 22\}$. Using the notation of the Foolbox library:\\
\texttt{confidence = [0,6,14,22], max\_iterations = 750, p\_norm = 2}.
\item FGSM attack~\cite{goodfellow2015explain}: $\ell_\infty$ norm with maximum norm-ball radius $\,\epsilon_{\max} = 1$. Using the notation of the Foolbox library: \\
\texttt{max\_epsilon = 1, p\_norm = inf}.
\end{itemize}
For the adversarial detection experiments in \S~\ref{detection_adv}, we reported results for the attack parameter setting that would produce adversarial samples with the lowest perturbation norm (least perceptible), in order to make the detection task challenging. This corresponds to $\,\epsilon = \frac{1}{255}\,$ for the PGD attack, $\textrm{confidence} = 0\,$ for the CW attack, and $\,\epsilon_{\max} = 1\,$ for the FGSM attack. 
These same parameters are also used in the experiments in Appendix F.2, F.3, F.4, F.6, and F.7. In Appendix F.1, the strength of the attack is varied and the attack parameters used are described there.
The adversarial samples are generated {\it once} from the clean samples corresponding to each train fold and test fold (from cross-validation), and saved to files for reuse in all the experiments. This way, we ensure that 
there is no randomness in the experiments arising due to the adversarial sample generation.

\smallskip
\noindent{\bf Adaptive Attack:}

Here we provide additional details on the adaptive (defense-aware) attack method proposed in \S~\ref{sec:dknn_attack}. The constant $\lambda$ in the objective function controls the perturbation norm of the adversarial sample, and smaller values of $\,\lambda\,$ lead to solutions with a smaller perturbation norm. We follow the approach of \cite{carlini2017towards} to set $\lambda$ to the smallest possible value that leads to a successful adversarial perturbation. This is found using a bisection line search over ten steps. An adversarial input is considered successful if it modifies the initially-correct class prediction of the defense method. 
We also follow the approach of \cite{carlini2017towards} to implicitly constraint the adversarial inputs to lie in the same range as the original inputs using the hyperbolic tangent and its inverse function. Suppose the inputs lie in the range $[a, b]^d$, the following sequence of transformations is applied to each component of the vectors
\vspace{-2mm}
\begin{align*}
\vspace{-0.25in}
z_i ~&=~ \tanh^{-1}(2 \,\frac{x_i - a}{b - a} - 1) \\
z^\prime_i ~&=~ z_i ~+~ w_i \\
x^\prime_i ~&=~ (b - a)\,\frac{1}{2}\,(1 + \tanh(z_i + w_i)) ~+~ a
\end{align*}
effectively allowing the perturbation $\bfw$ to be optimized over $\reals^d$. The resulting unconstrained optimization is solved using the RMSProp variant of stochastic gradient descent with adaptive learning rate~\cite{ruder2016overview}. The maximum number of gradient descent steps is set to $1000$. For all experiments based on the adaptive attack method, the attack targets the variant of \proposed based on p-value normalization at the individual layers and layer pairs, Fisher's method for p-value aggregation, and the multinomial test statistic.

\section{Additional Experiments}
\label{app:additional}
In this section, we supplement the results in \S~\ref{sec_exp} with more extensive experiments.

\subsection{Attack Transfer and Attacks of Varying Strength}
\label{app:attack_transfer}
We evaluate the performance of the detection methods on a task with different adversarial attacks used in the train and test sets. The strength of the attacks are also varied in both the train and test sets. 
Recall that we use a 5-fold cross-validation framework for evaluation. Hence, for a train/test split corresponding to fold $\,i \in \{1, 2, 3, 4, 5\}\,$, adversarial samples from Attack A are generated in the train split, while adversarial samples from Attack B are generated in the test split. Supervised methods, \dm and \lid, learn using both the adversarial and clean samples from the train split, while the unsupervised methods, \proposed, \odds, \dknn, and \trust, learn using only the clean samples from the train split.
\begin{figure*}[htb]
\centering
\includegraphics[width=0.8\linewidth]{Figures/legend_adversarial.png}
\vspace{-2mm}

\subfloat[{{\small CIFAR-10}}]{\label{fig:avg_prec_cifar_CW_to_PGD}{\includegraphics[width=0.3\linewidth]{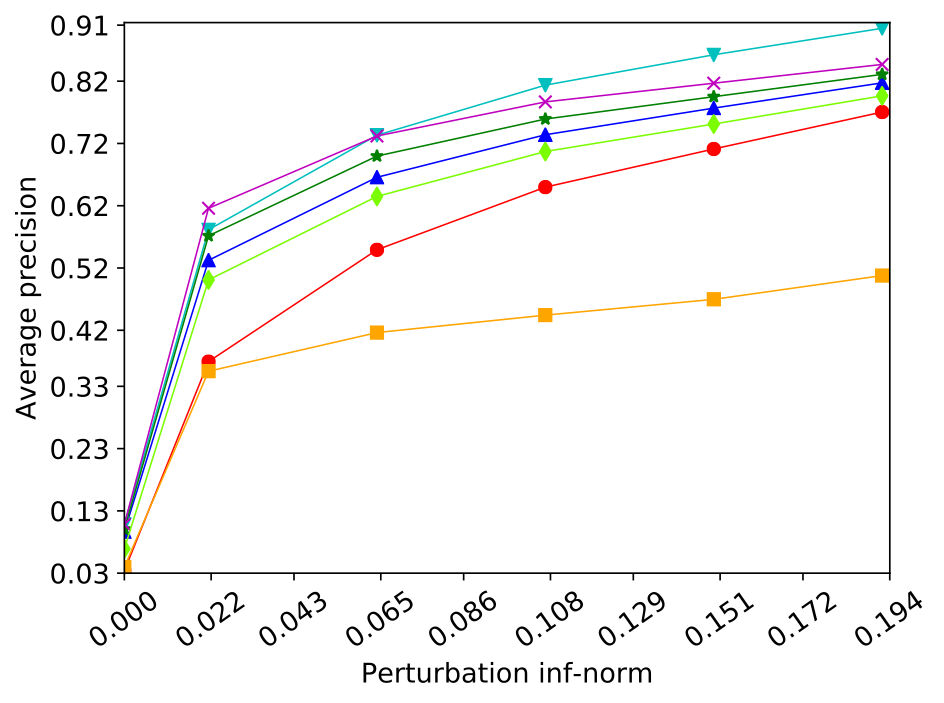}} }
\subfloat[{{\small SVHN}}]{\label{fig:avg_prec_svhn_CW_to_PGD}{\includegraphics[width=0.3\linewidth]{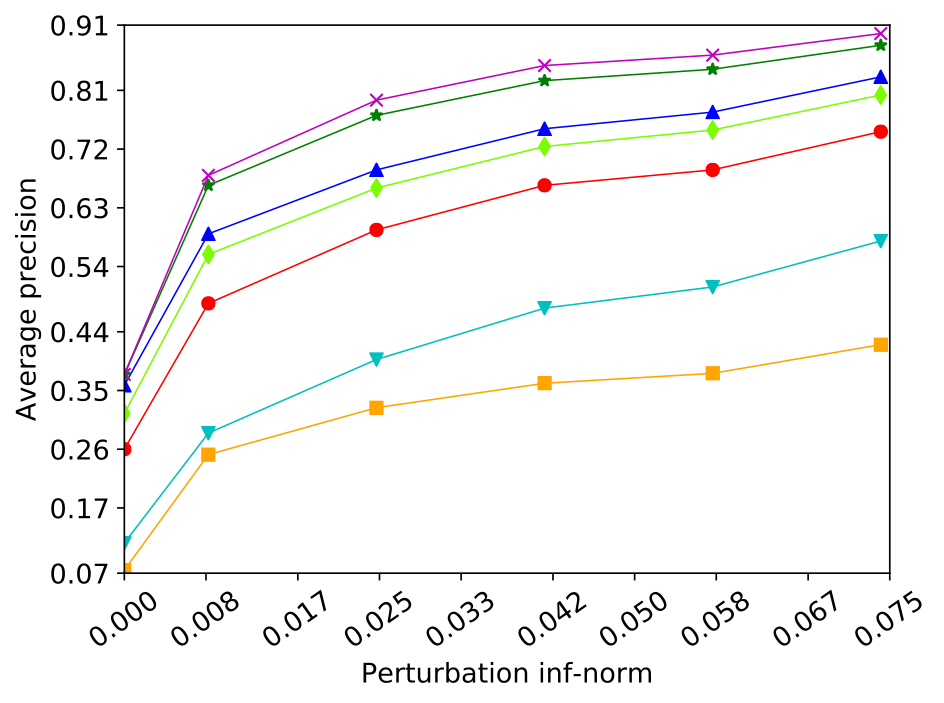}} } 
\subfloat[{{\small MNIST}}]{\label{fig:avg_prec_mnist_CW_to_PGD}{\includegraphics[width=0.3\linewidth]{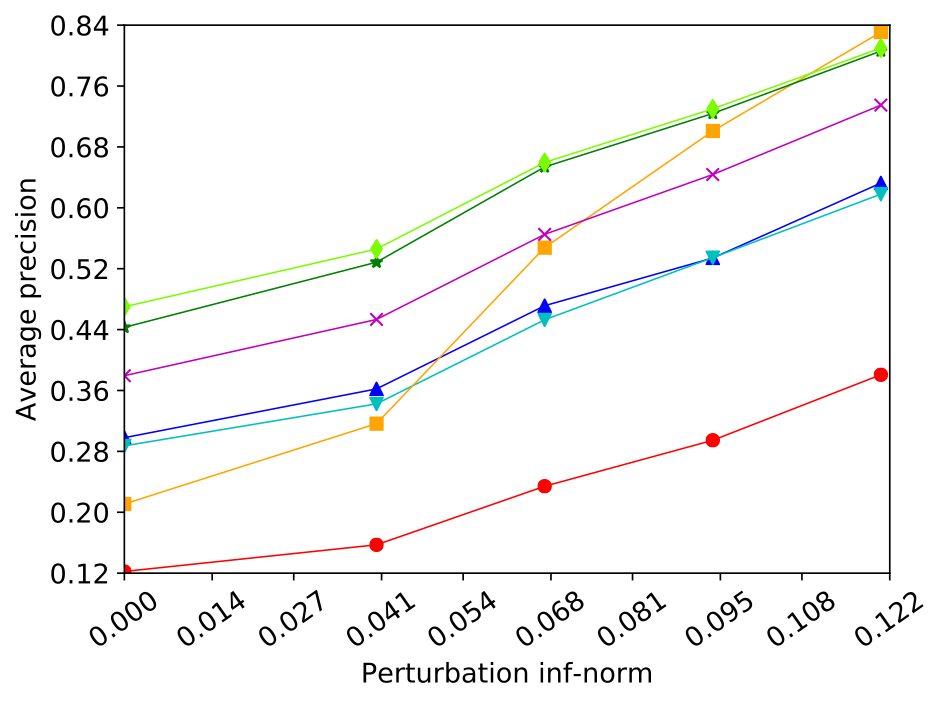}} }
\caption{Average precision on the attack transfer experiment: CW to PGD attack.}
\label{fig:transfer_CW_to_PGD}
\end{figure*}
\begin{figure*}[htb]
\centering
\includegraphics[width=0.8\linewidth]{Figures/legend_adversarial.png}
\vspace{-2mm}

\subfloat[{{\small CIFAR-10}}]{\label{fig:avg_prec_cifar_PGD_to_CW}{\includegraphics[width=0.3\linewidth]{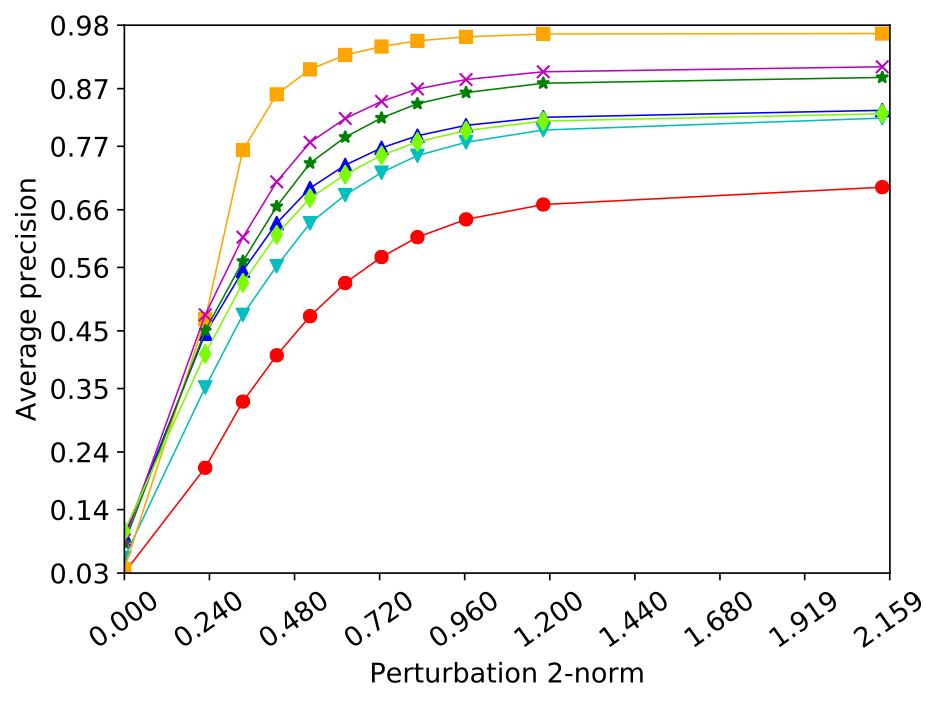}} }
\subfloat[{{\small SVHN}}]{\label{fig:avg_prec_svhn_PGD_to_CW}{\includegraphics[width=0.3\linewidth]{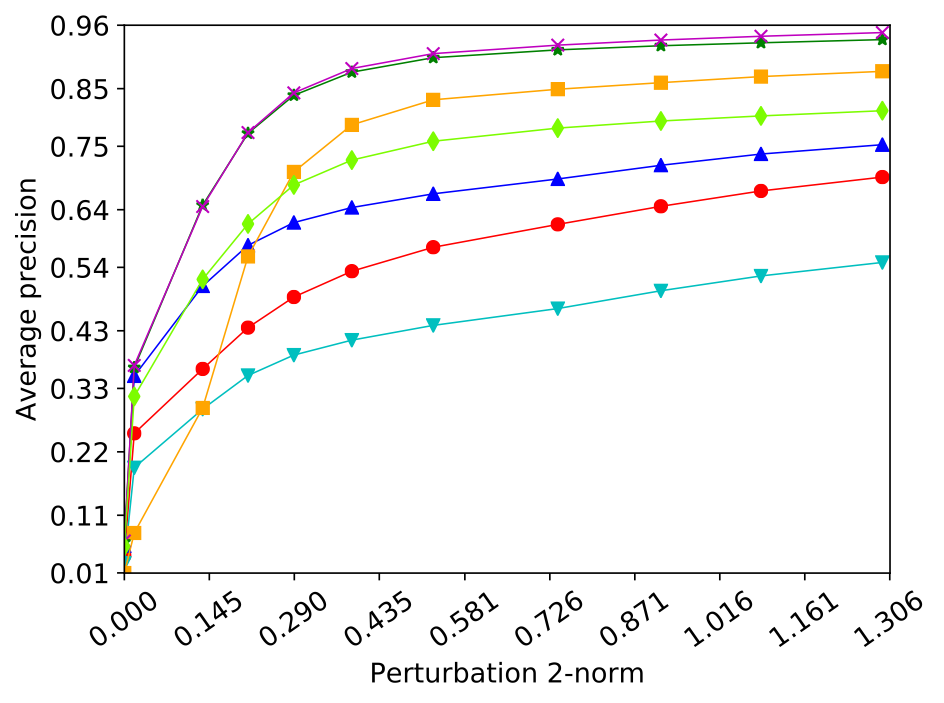}} } 
\subfloat[{{\small MNIST}}]{\label{fig:avg_prec_mnist_PGD_to_CW}{\includegraphics[width=0.3\linewidth]{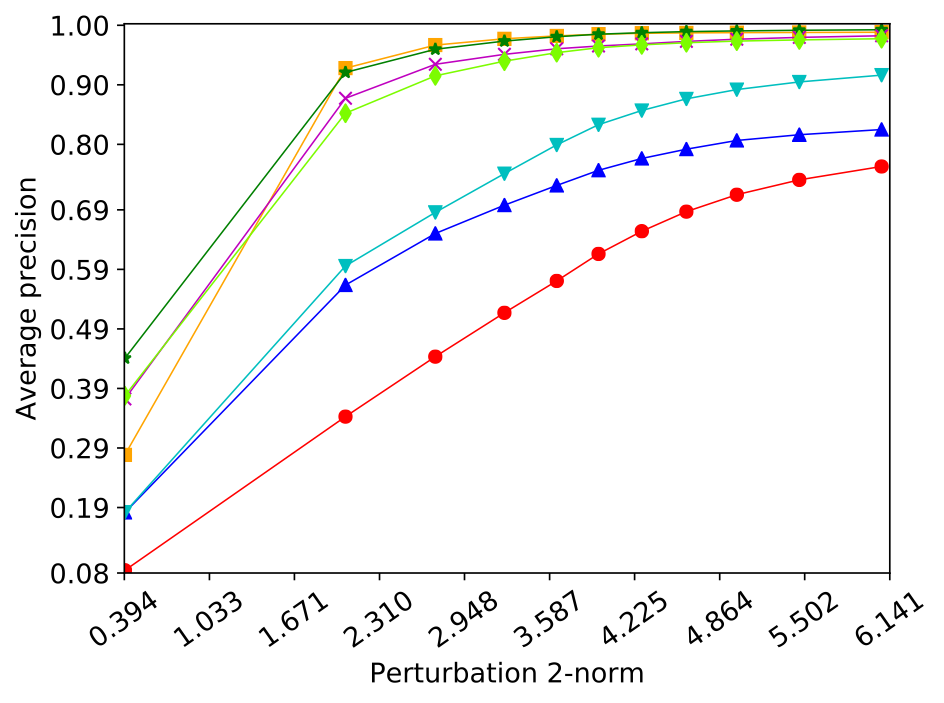}} }
\caption{Average precision on the attack transfer experiment: PGD to CW attack.}
\label{fig:transfer_PGD_to_CW}
\end{figure*}

\smallskip
\noindent{\bf $\ell_2$-CW attack to $\ell_\infty$-PGD attack}

The $\ell_\infty$-PGD~\cite{madry2018towards} attack is applied on the test split with perturbation strength parameter $\epsilon$ varied over the values $\,\{1/255, 3/255, 5/255, 7/255, 9/255\}$. For each clean sample, one of these $\epsilon$ values is randomly selected and used to create an attack sample. The $\ell_2$-CW attack~\cite{carlini2017towards} is applied to the train split with the confidence parameter value randomly selected from the set $\,\{0, 6, 14, 22\}$. The result of this experiment is given in Figure~\ref{fig:transfer_CW_to_PGD}, with the average precision of the compared methods plotted as a function of the perturbation norm on the x-axis. The $\,\ell_\infty\,$ norm is used on the x-axis to match the norm type used by the PGD attack (that is applied to the test split). We observe that \dm and both variants of \proposed have the best performance on CIFAR-10, while both variants of \proposed outperform the other methods on SVHN. On MNIST, \proposed based on the aK-LPE method (\S~\ref{sec:normalizing_pvalues}.B) and \trust have the best performance. The methods \dm, \trust, and \odds do not consistently have good performance, while the \lid method has poor performance on all datasets.

\smallskip
\noindent{\bf $\ell_\infty$-PGD attack to $\ell_2$-CW attack}

In this experiment, the $\ell_2$-CW attack is applied to the test split and the $\ell_\infty$-PGD attack is applied to the train split. The attack parameters (strength) are varied as described earlier. The result of this experiment is given in Figure~\ref{fig:transfer_PGD_to_CW}, with the average precision of the methods plotted as a function of the $\,\ell_2\,$ norm of the attack (since the $\ell_2$-CW attack is applied to the test split). From the figure, we make the following observations. On CIFAR-10, \odds outperforms the other methods, followed by the two variants of \proposed. On SVHN, the two variants of \proposed have the best performance, followed by \odds. On MNIST, the methods \proposed, \odds, and \trust have very similar average precision that is higher than the other methods. The methods \dm, \dknn, and \lid do not have good performance in this experiment. We hypothesize that the \odds method~\cite{roth2019odds} works well in detecting CW attacks because it uses a test statistic based on the noise-perturbed log-odds ratio of all pairs of classes, which is well-matched to the CW attack that is based on skewing the log-odds ratio of the class pair involved in the attack.

\subsection{Ablation Experiments}
\label{app:ablation}
We discuss the ablation experiments that we performed to get a better understanding of the components of the proposed method. Specifically, we are interested in understanding 
\begin{enumerate}
    \item The relative performance of the proposed p-value based normalization methods (\S~\ref{sec:normalizing_pvalues}) and the p-value aggregation methods (\S~\ref{sec:pvalue_aggregation}).
    \item The value of including p-values from test statistics at all layer pairs (in addition to the individual layers).
    \item The relative performance of using only the last few layer representations for detection.
    \item The relative performance of the two scoring methods in \S~\ref{sec:scoring} on the task of adversarial detection.
\end{enumerate}

Tables \ref{tab:ablation_res1}, \ref{tab:ablation_res2}, and \ref{tab:ablation_res3} summarize the results of these experiments on the task of detecting adversarial samples from the $\ell_\infty$-PGD attack with $\epsilon = 1 \,/\, 255$, and the $\ell_2$-CW attack with confidence set to $0$. DNNs trained on the SVHN and CIFAR-10 datasets (described earlier) are used, and average precision and pAUC-$0.2$ (partial AUROC below FPR $= 0.2$) are reported as detection metrics. Unlike the results in Fig. \ref{fig:cifar10_prec} and Fig. \ref{fig:svhn_prec}, we do not vary the proportion of adversarial samples, and report performance with all the adversarial samples included.

%
\begin{table*}[htb]
\centering
\caption{\small Ablation experiment - performance of different p-value normalization and aggregation methods, and the effect of including/excluding layer pairs. The multinomial test statistic is used in all cases.}
\vspace{-1mm}
\label{tab:ablation_res1}
\resizebox{0.95\textwidth}{!}{%
\begin{tabular}{ll|cc|cc|cc|cc}
\toprule
\multirow{2}{*}{\begin{tabular}[c]{@{}l@{}}Normalization \\ method\end{tabular}} &
  \multirow{2}{*}{\begin{tabular}[c]{@{}l@{}}Aggregation \\ method\end{tabular}} &
  \multicolumn{2}{c|}{SVHN, PGD ($\epsilon = 1/255$)} &
  \multicolumn{2}{c|}{SVHN, CW (confidence = $0$)} &
  \multicolumn{2}{c|}{CIFAR-10, PGD ($\epsilon = 1/255$)} &
  \multicolumn{2}{c}{CIFAR-10, CW (confidence = $0$)} \\ 
  \cline{3-10} 
  &
  &
  \begin{tabular}[c]{@{}c@{}}average \\ precision\end{tabular} &
  pAUC-0.2 &
  \begin{tabular}[c]{@{}c@{}}average \\ precision\end{tabular} &
  pAUC-0.2 &
  \begin{tabular}[c]{@{}c@{}}average \\ precision\end{tabular} &
  pAUC-0.2 &
  \begin{tabular}[c]{@{}c@{}}average \\ precision\end{tabular} &
  pAUC-0.2 \\ 
  \midrule
\multirow{2}{*}{\begin{tabular}[c]{@{}l@{}}p-values from layers\\ \& layer pairs (\S \ref{sec:normalizing_pvalues}.A)\end{tabular}} &
  Fisher &
  0.7382 &
  \textbf{0.8025} &
  0.9631 &
  0.9213 &
  0.7710 &
  \textbf{0.7790} &
  0.9664 &
  0.9377 \\
  &
  HMP &
  0.7296 &
  0.7966 &
  0.9611 &
  0.9171 &
  0.7614 &
  0.7705 &
  0.9653 &
  0.9344 \\ 
  \hline
\multirow{2}{*}{p-values from layers} &
  Fisher &
  \textbf{0.7393} &
  0.8005 &
  \textbf{0.9634} &
  \textbf{0.9214} &
  \textbf{0.7734} &
  0.7781 &
  \textbf{0.9667} &
  \textbf{0.9380} \\
  &
  HMP &
  0.7247 &
  0.7925 &
  0.9591 &
  0.9140 &
  0.7538 &
  0.7617 &
  0.9616 &
  0.9315 \\ 
  \hline
\begin{tabular}[c]{@{}l@{}}Multivariate p-value \\ (aK-LPE, \S \ref{sec:normalizing_pvalues}.B)\end{tabular} &
  None &
  0.7161 &
  0.7840 &
  0.9559 &
  0.9042 &
  0.7437 &
  0.7518 &
  0.9650 &
  0.9296 \\ 
  \bottomrule
\end{tabular}%
}
\end{table*}
Table \ref{tab:ablation_res1} focuses on points 1 and 2, and compares the proposed p-value normalization and aggregation methods. P-values from the layer pairs are included in the first case and not included in the second case.
The best performing configuration (across the rows) is highlighted in bold. We observe that p-value normalization at the layers with aggregation using Fisher's method has the best performance in most cases. Including the layer pairs did not result in a significant improvement in these experiments.
It is surprising that Fisher's method has better (in some cases comparable) performance compared to the HMP and the aK-LPE methods despite its simplistic assumption of independent tests (p-values).
We believe this can be attributed to the fact that 
the multivariate p-values estimated by the aK-LPE method (Eq. (\ref{eq:pval_klpe})) require a large sample size to converge to their true values. Since we apply this estimator class conditionally, the moderate number of samples per class (ranging from 500 to 5000 in our experiments) may result in estimation errors.
Also, we conjecture that Fisher's method achieves a higher detection rate (TPR) at the expense of a higher FPR, while the HMP method has a more conservative TPR with a lower FPR.

\begin{table*}[htb]
\centering
\caption{\small Ablation experiment: effect of including only the last few layers for detection.}
\vspace{-1mm}
\label{tab:ablation_res2}
\resizebox{0.95\textwidth}{!}{%
\begin{tabular}{lll|cc|cc|cc|cc}
\toprule
\multirow{2}{*}{\begin{tabular}[c]{@{}l@{}}Normalization \\ method\end{tabular}} &
  \multirow{2}{*}{\begin{tabular}[c]{@{}l@{}}Aggregation \\ method\end{tabular}} &
  \multirow{2}{*}{\begin{tabular}[c]{@{}l@{}}Layers \\ included\end{tabular}} &
  \multicolumn{2}{c|}{SVHN, PGD ($\epsilon = 1 \,/\, 255$)} &
  \multicolumn{2}{c|}{SVHN, CW (confidence = $0$)} &
  \multicolumn{2}{c|}{CIFAR-10, PGD ($\epsilon = 1 \,/\, 255$)} &
  \multicolumn{2}{c}{CIFAR-10, CW (confidence = $0$)} \\ 
  \cline{4-11} 
   &
   &
   &
  \begin{tabular}[c]{@{}c@{}}average \\ precision\end{tabular} &
  pAUC-0.2 &
  \begin{tabular}[c]{@{}c@{}}average \\ precision\end{tabular} &
  pAUC-0.2 &
  \begin{tabular}[c]{@{}c@{}}average \\ precision\end{tabular} &
  pAUC-0.2 &
  \begin{tabular}[c]{@{}c@{}}average \\ precision\end{tabular} &
  pAUC-0.2 \\ 
  \midrule
\multirow{4}{*}{\begin{tabular}[c]{@{}l@{}}p-values from layers\\ \& layer pairs (\S \ref{sec:normalizing_pvalues}.A)\end{tabular}} &
  \multirow{4}{*}{Fisher} &
  All &
  \textbf{0.7382} &
  \textbf{0.8025} &
  \textbf{0.9631} &
  \textbf{0.9213} &
  \textbf{0.7710} &
  \textbf{0.7790} &
  0.9664 &
  0.9377 \\
   &
   &
  Final (logits) &
  0.6347 &
  0.7396 &
  0.9141 &
  0.8538 &
  0.7399 &
  0.7636 &
  0.9664 &
  0.9371 \\
   &
   &
  Last 2 &
  0.6440 &
  0.7431 &
  0.9213 &
  0.8599 &
  0.7410 &
  0.7650 &
  \textbf{0.9688} &
  \textbf{0.9401} \\
   &
   &
  Last 3 &
  0.6847 &
  0.7674 &
  0.9388 &
  0.8857 &
  0.7473 &
  0.7657 &
  0.9660 &
  0.9358 \\ 
  \bottomrule
\end{tabular}%
}
\end{table*}
Table \ref{tab:ablation_res2} focuses on point 3 and compares the performance of using all the layer representations (listed in tables \ref{tab:layers_svhn} and \ref{tab:layers_cifar}) with the performance from using only the final one, two, or three layer representations~\footnote{We focus on the deeper layers since their representations are most useful for classification.}. We observe that including more layers generally increases the detection performance, confirming the intuition behind using multiple layers. For the CW attack on CIFAR-10, using only the final (logit) layer has comparable performance to using all the layers. This is consistent with the design of the CW attack based on only the logit layer representation.

\begin{table*}[htb]
\centering
\caption{\small Ablation experiment: comparison of the adversarial and OOD score functions (\S \ref{sec:scoring}) on different adversarial attacks. The adversarial score function outperforms the OOD score function in all cases.}
\vspace{-1mm}
\label{tab:ablation_res3}
\resizebox{0.95\textwidth}{!}{%
\begin{tabular}{lll|cc|cc|cc|cc}
\toprule
\multirow{2}{*}{\begin{tabular}[c]{@{}l@{}}Normalization \\ method\end{tabular}} &
  \multirow{2}{*}{\begin{tabular}[c]{@{}l@{}}Aggregation \\ method\end{tabular}} &
  \multirow{2}{*}{\begin{tabular}[c]{@{}l@{}}Scoring \\ method\end{tabular}} &
  \multicolumn{2}{c|}{SVHN, PGD ($\epsilon = 1 \,/\, 255$)} &
  \multicolumn{2}{c|}{SVHN, CW (confidence = $0$)} &
  \multicolumn{2}{c|}{CIFAR-10, PGD ($\epsilon = 1 \,/\, 255$)} &
  \multicolumn{2}{c}{CIFAR-10, CW (confidence = $0$)} \\ \cline{4-11} 
   &
   &
   &
  \begin{tabular}[c]{@{}c@{}}average \\ precision\end{tabular} &
  pAUC-0.2 &
  \begin{tabular}[c]{@{}c@{}}average \\ precision\end{tabular} &
  pAUC-0.2 &
  \begin{tabular}[c]{@{}c@{}}average \\ precision\end{tabular} &
  pAUC-0.2 &
  \begin{tabular}[c]{@{}c@{}}average \\ precision\end{tabular} &
  pAUC-0.2 \\ 
  \midrule
\multirow{4}{*}{\begin{tabular}[c]{@{}l@{}}p-values from layers\\ \& layer pairs (\S \ref{sec:normalizing_pvalues}.A)\end{tabular}} &
  \multirow{2}{*}{Fisher} &
  Adversarial &
  \textbf{0.7382} &
  \textbf{0.8025} &
  \textbf{0.9631} &
  \textbf{0.9213} &
  \textbf{0.7710} &
  \textbf{0.7790} &
  \textbf{0.9664} &
  \textbf{0.9377} \\
   &
   &
  OOD &
  0.7078 &
  0.7736 &
  0.9524 &
  0.8973 &
  0.7492 &
  0.7612 &
  0.9620 &
  0.9253 \\ \cline{2-11} 
  &
  \multirow{2}{*}{HMP} &
  Adversarial &
  \textbf{0.7296} &
  \textbf{0.7966} &
  \textbf{0.9611} &
  \textbf{0.9171} &
  \textbf{0.7614} &
  \textbf{0.7705} &
  \textbf{0.9653} &
  \textbf{0.9344} \\
   &
   &
  OOD &
  0.6946 &
  0.7617 &
  0.9482 &
  0.8882 &
  0.7396 &
  0.7501 &
  0.9599 &
  0.9205 \\ \hline
\multirow{2}{*}{\begin{tabular}[c]{@{}l@{}}Multivariate p-value \\ (aK-LPE, \S \ref{sec:normalizing_pvalues}.B)\end{tabular}} &
  \multirow{2}{*}{None} &
  Adversarial &
  \textbf{0.7161} &
  \textbf{0.7840} &
  \textbf{0.9559} &
  \textbf{0.9042} &
  \textbf{0.7437} &
  \textbf{0.7518} &
  \textbf{0.9650} &
  \textbf{0.9296} \\
   &
   &
  OOD &
  0.6986 &
  0.7664 &
  0.9511 &
  0.8941 &
  0.7065 &
  0.7247 &
  0.9575 &
  0.9149 \\ 
  \bottomrule
\end{tabular}%
}
\end{table*}
Table \ref{tab:ablation_res3} focuses on point 4 to understand if the score function (\ref{eq:score_func_adver}) is better suited for adversarial samples since it considers the aggregate p-values from the candidate true classes. 
From the table, we observe that the adversarial score function clearly outperforms the OOD score function for each combination of normalization and aggregation method.

\subsection{Results on the FGSM Attack}
\label{app:fgsm}
\begin{figure*}[htb]
\centering
\includegraphics[width=0.8\linewidth]{Figures/legend_adversarial.png}
\vspace{-2mm}

\subfloat[{{\small CIFAR-10}}]{\label{fig:avg_prec_cifar_FGSM}{\includegraphics[width=0.3\linewidth]{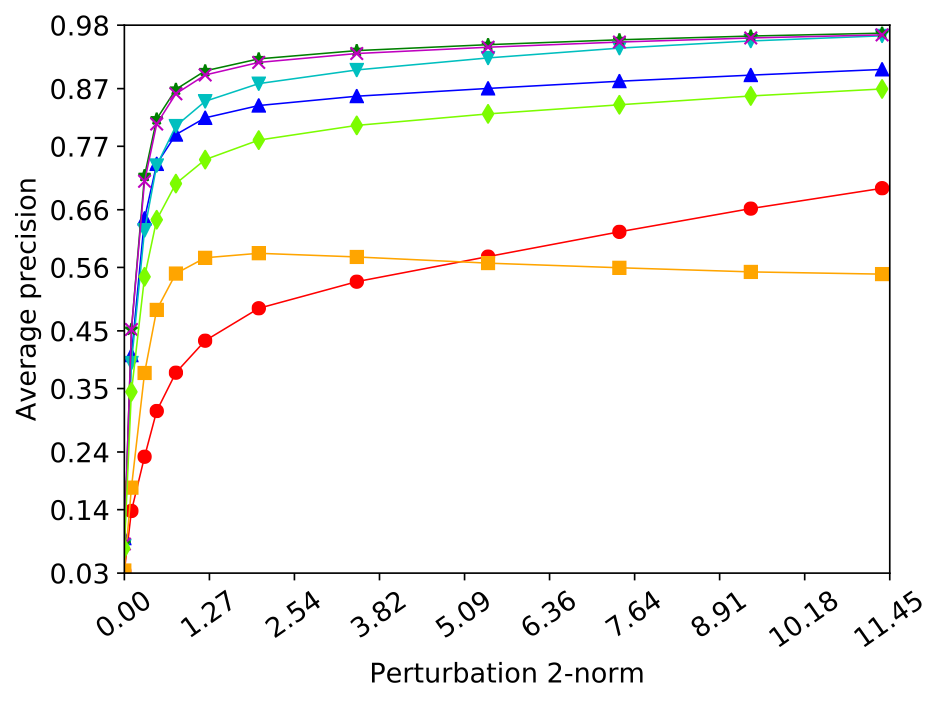}} }
\subfloat[{{\small SVHN}}]{\label{fig:avg_prec_svhn_FGSM}{\includegraphics[width=0.3\linewidth]{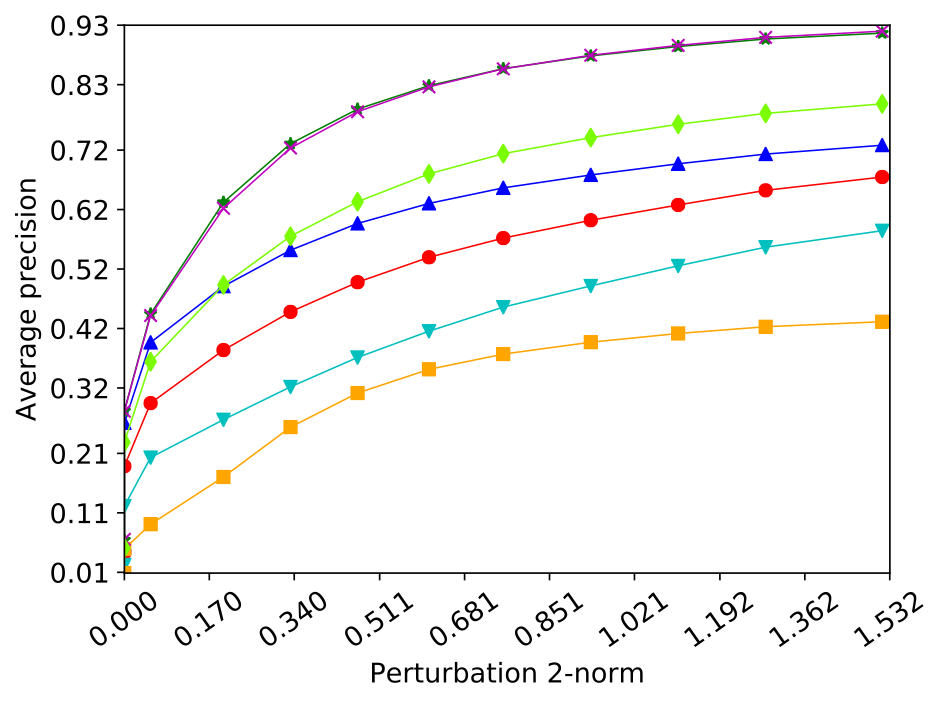}} } 
\subfloat[{{\small MNIST}}]{\label{fig:avg_prec_mnist_FGSM}{\includegraphics[width=0.3\linewidth]{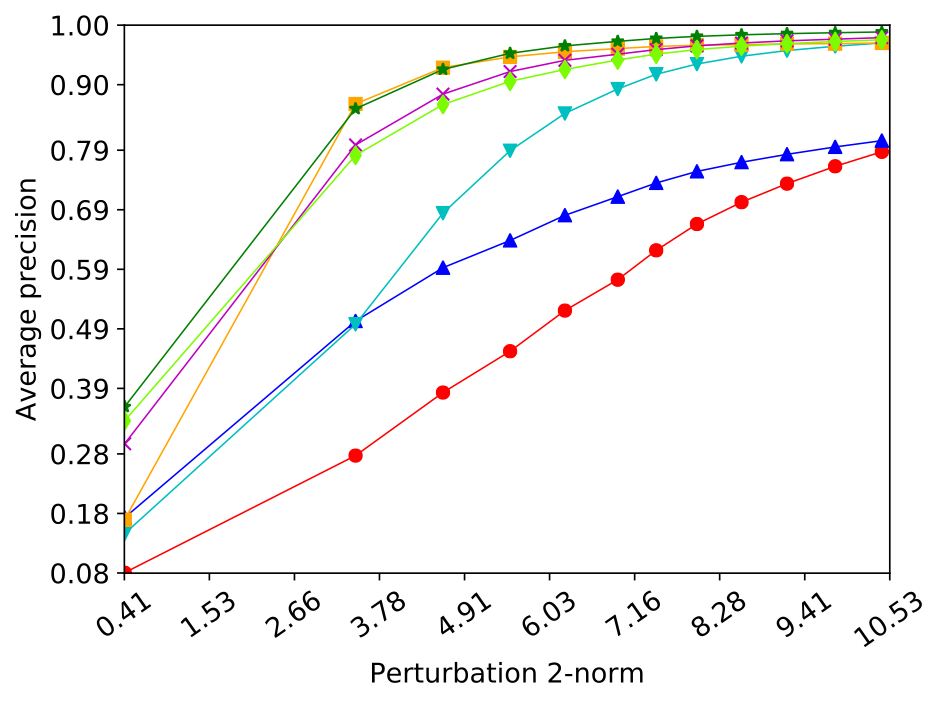}} }
\vspace{-1mm}
\caption{Average precision on the FGSM attack ($\epsilon^{}_{\max} = 1$) for all datasets.}
\vspace{-2mm}
\label{fig:fgsm}
\end{figure*}

Figure~\ref{fig:fgsm} presents the average precision of the different detection methods as a function of the $\ell_2$ norm of perturbation for the FGSM attack method. It is clear that both variants of \proposed outperform the other methods, consistent with the trend observed on other attacks in \S~\ref{sec_exp}.

\subsection{Evaluation of Partial Area Under the ROC Curve}
\label{app:pauc_results}
Here we compare the performance of methods using pAUC-$0.2$, a metric calculating the partial area under the ROC curve for FPR below $0.2$. Comparing the area under the entire ROC curve can lead to misleading interpretations because it includes FPR values that one would rarely choose to operate in. Therefore, to reflect realistic operating conditions, we chose a maximum FPR of $0.2$. Recall that for the PGD attack we vary the proportion of adversarial samples along the x-axis because most of the samples from this attack have the same perturbation norm.

\begin{figure*}
  \centering
  \includegraphics[width=0.7\linewidth]{Figures/legend_adversarial.png}
  
  \subfloat[{CW, confidence $= 0$}]{ \includegraphics[width=.29\textwidth]{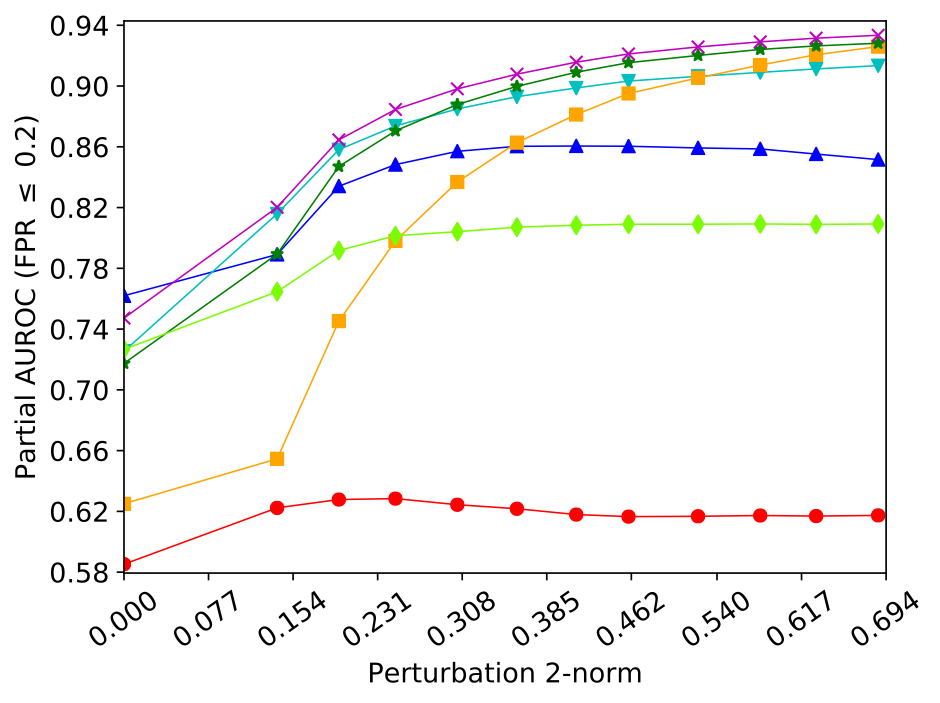} }
  \hspace{10mm}
  \subfloat[{Adaptive attack}]{ \includegraphics[width=.29\textwidth]{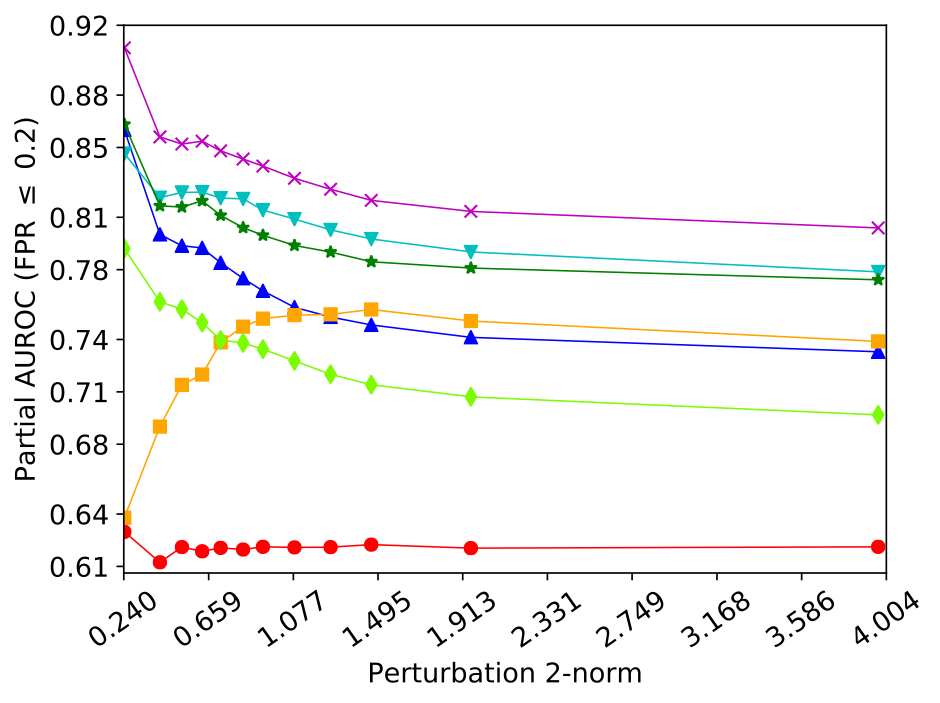} }
  
  \smallskip
  \subfloat[{FGSM, $\epsilon^{}_{\max} = 1$}]{ \includegraphics[width=.29\textwidth]{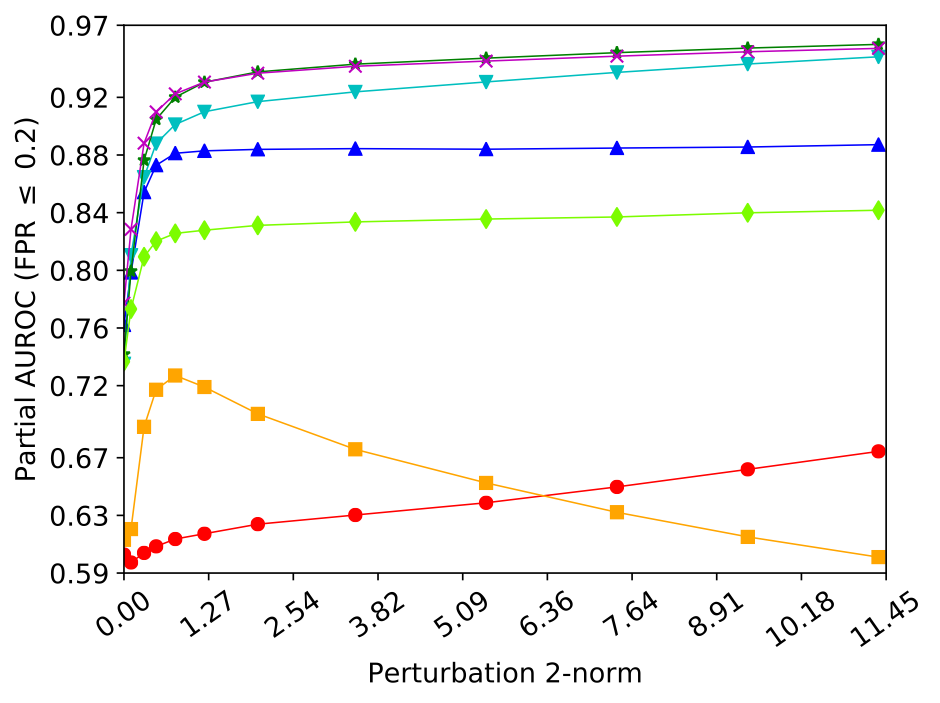} }
  \hspace{10mm}
  \subfloat[{PGD, $\epsilon = 1/255$}]{ \includegraphics[width=.29\textwidth]{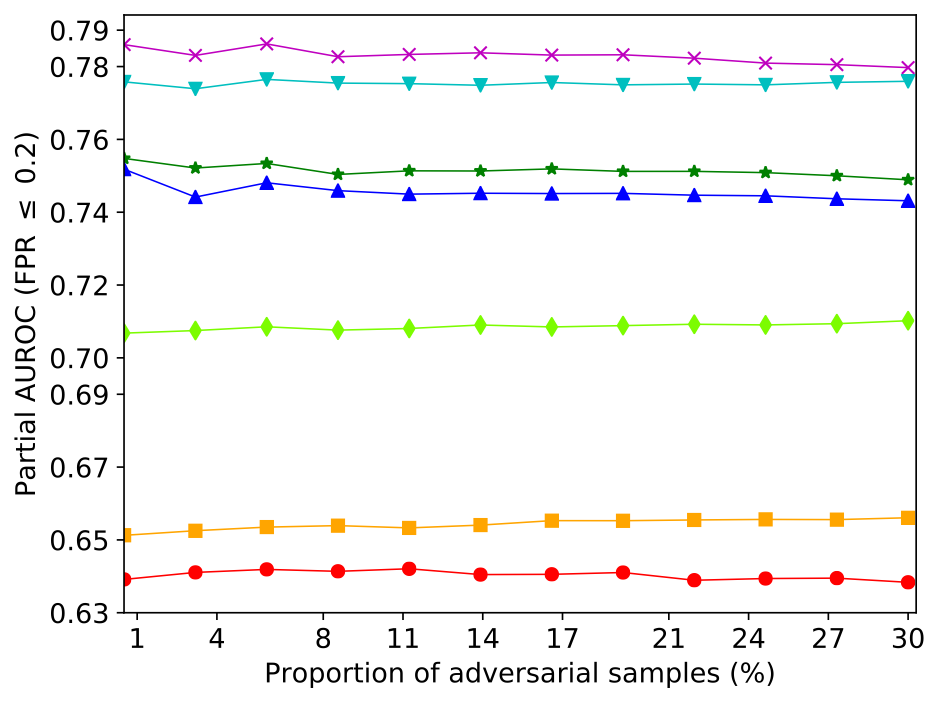} }
\caption{CIFAR-10 experiments: pAUC-$0.2$ for different attacks.}
\label{fig:cifar10_pauc}
\end{figure*}

\begin{figure*}
  \centering
  \includegraphics[width=0.7\linewidth]{Figures/legend_adversarial.png}
  
  \subfloat[{CW, confidence $= 0$}]{ \includegraphics[width=0.29\textwidth]{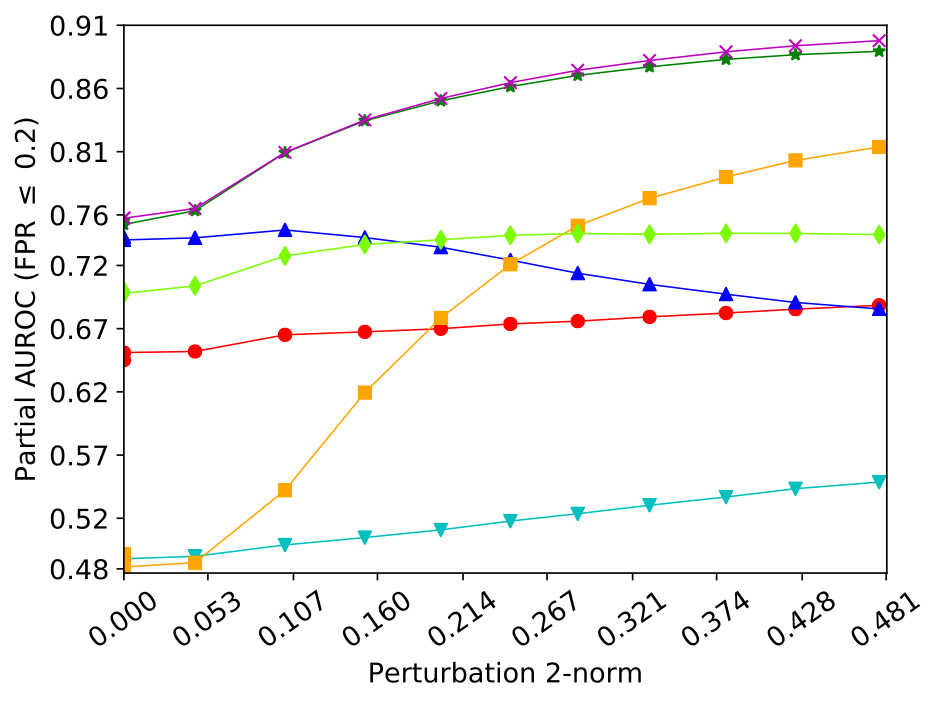} }
  \hspace{10mm}
  \subfloat[{Adaptive attack}]{ \label{fig:svhn_pauc_custom}{\includegraphics[width=.29\textwidth]{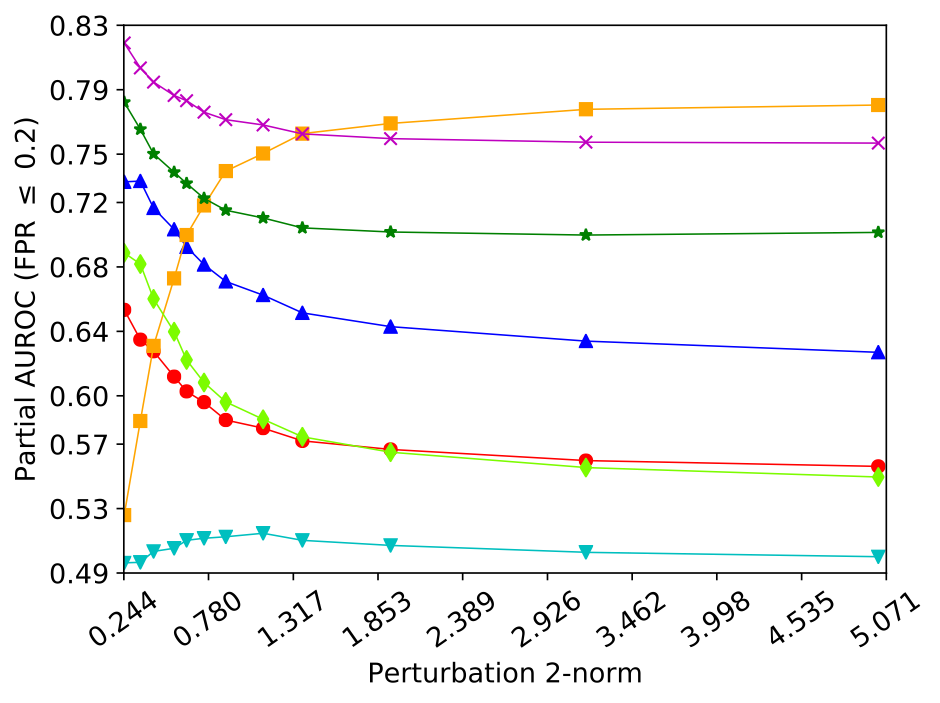}} }
  
  \smallskip
  \subfloat[{FGSM, $\epsilon^{}_{\max} = 1$}]{ \includegraphics[width=.29\textwidth]{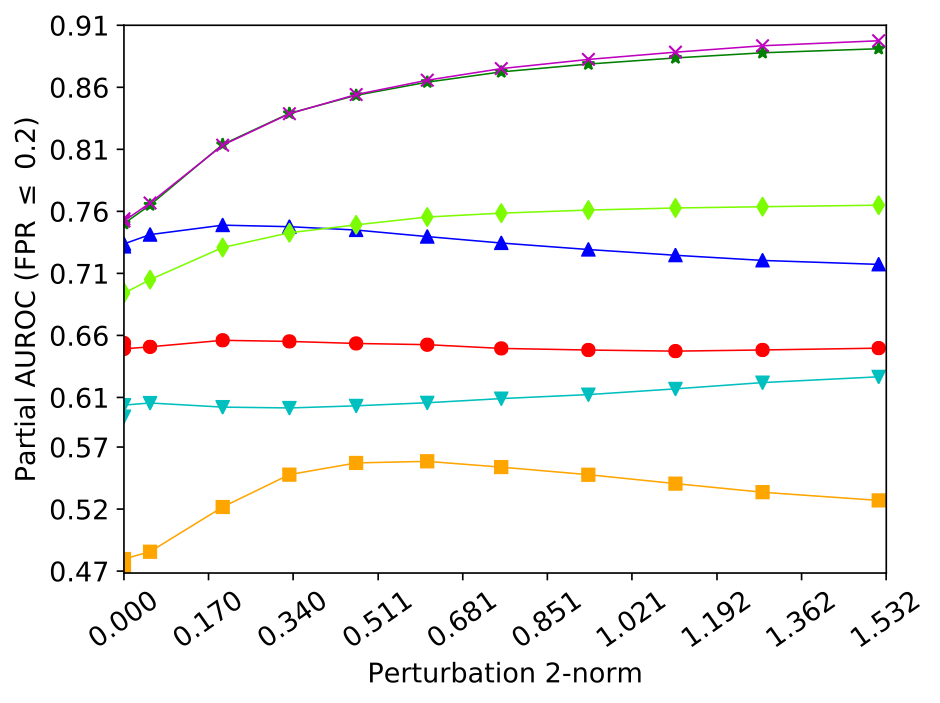} }
  \hspace{10mm}
  \subfloat[{PGD, $\epsilon = 1/255$}]{ \includegraphics[width=.29\textwidth]{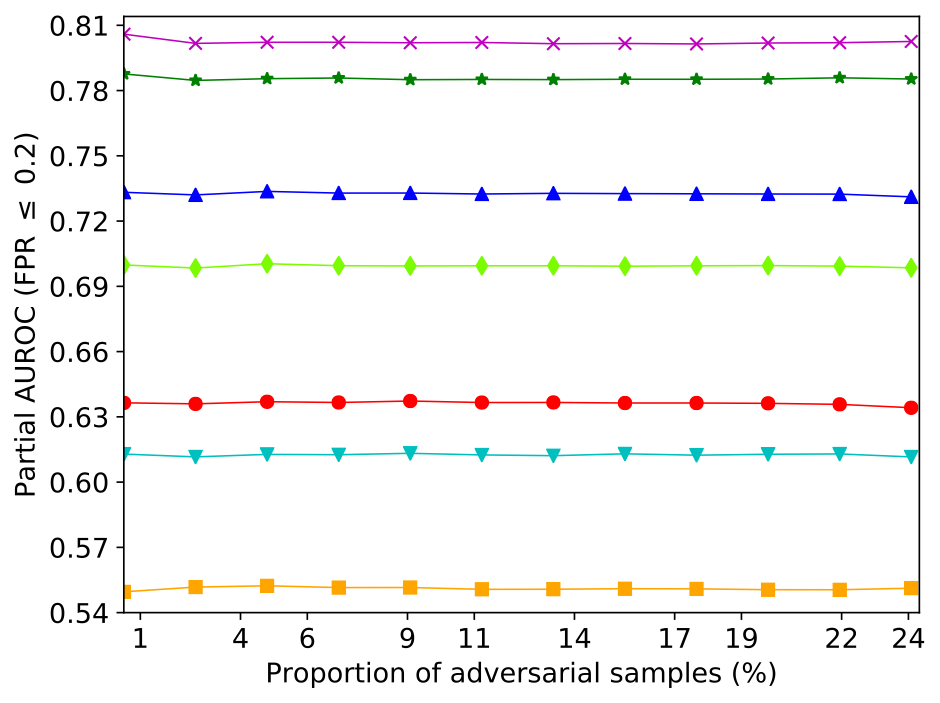} }
\caption{SVHN experiments: pAUC-$0.2$ for different attacks.}
\label{fig:svhn_pauc}
\end{figure*}

On the CIFAR-10 dataset (Figure~\ref{fig:cifar10_pauc}), we observe that \proposed has better performance than the other methods in most cases. On the adaptive attack, \dm has slightly better performance than \proposed with the multivariate p-value estimation method (aK-LPE). We make similar observations on the SVHN dataset (Figure~\ref{fig:svhn_pauc}), with a minor exception on the adaptive attack where the \odds method performs better than \proposed as the perturbation norm increases.

\begin{figure*}
  \centering
  \includegraphics[width=0.7\linewidth]{Figures/legend_adversarial.png}
  
  \subfloat[{CW, confidence $= 0$}]{ \includegraphics[width=.29\textwidth]{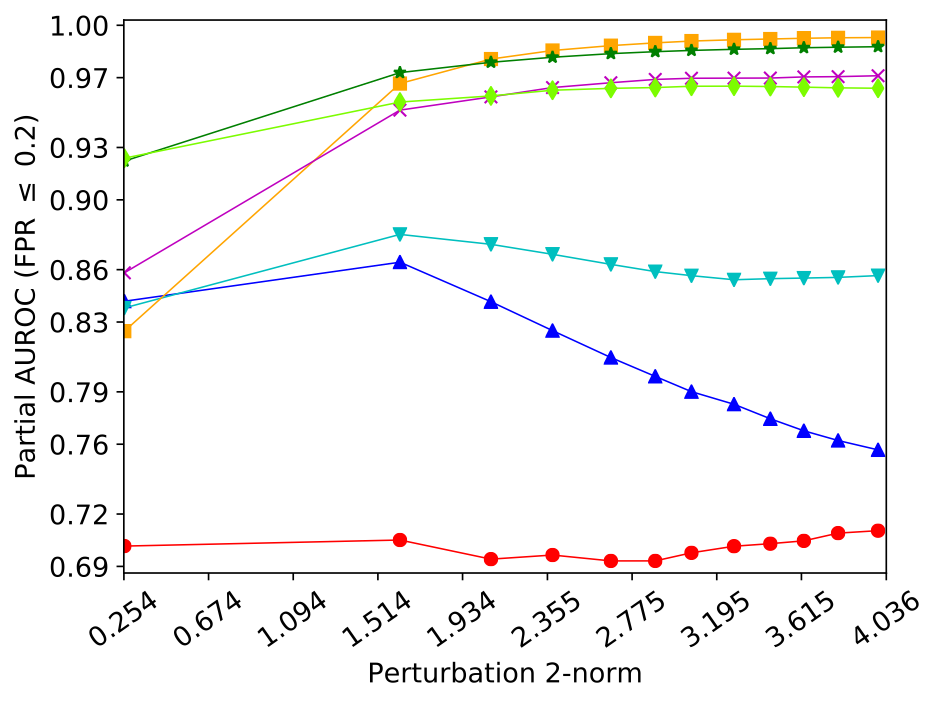} }
  \hspace{10mm}
  \subfloat[{Adaptive attack}]{ \includegraphics[width=.29\textwidth]{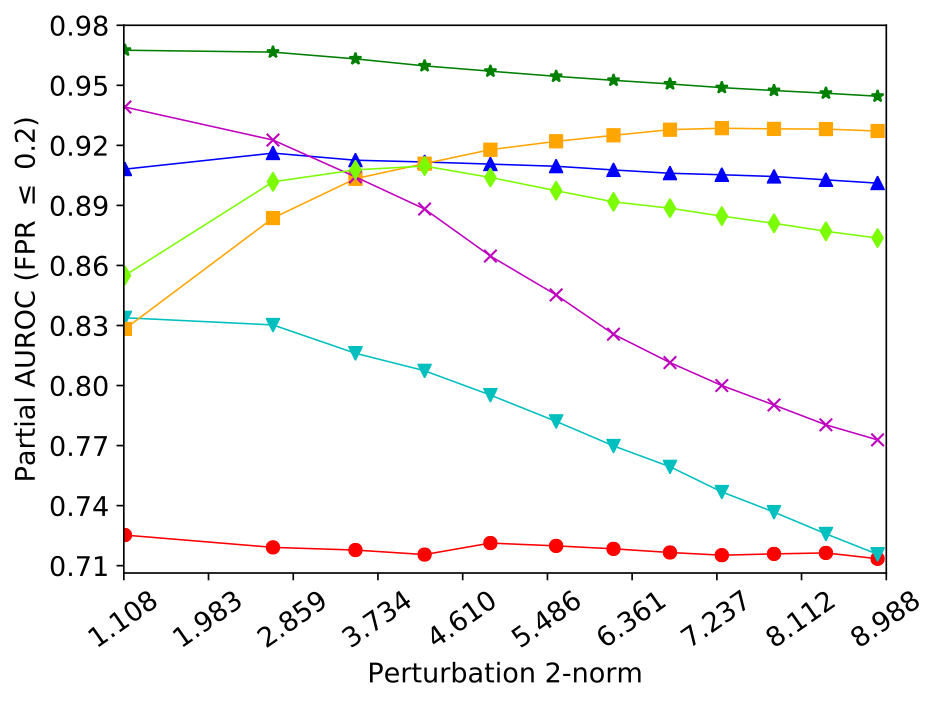} }
  
  \smallskip
  \subfloat[{FGSM, $\epsilon^{}_{\max} = 1$}]{ \includegraphics[width=.29\textwidth]{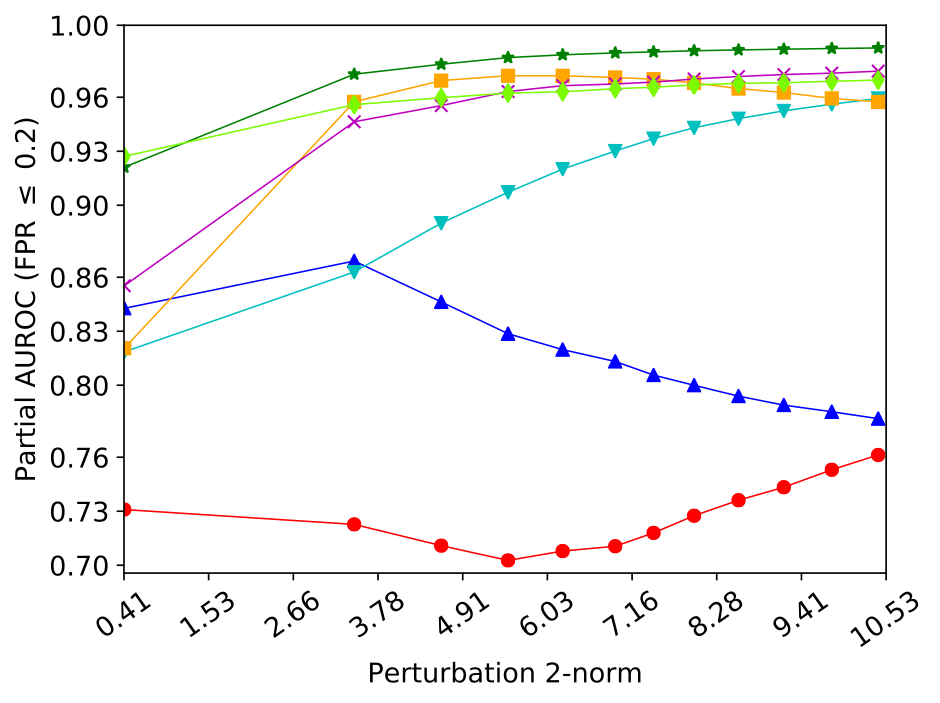} }
  \hspace{10mm}
  \subfloat[{PGD, $\epsilon = 1/255$}]{ \includegraphics[width=.29\textwidth]{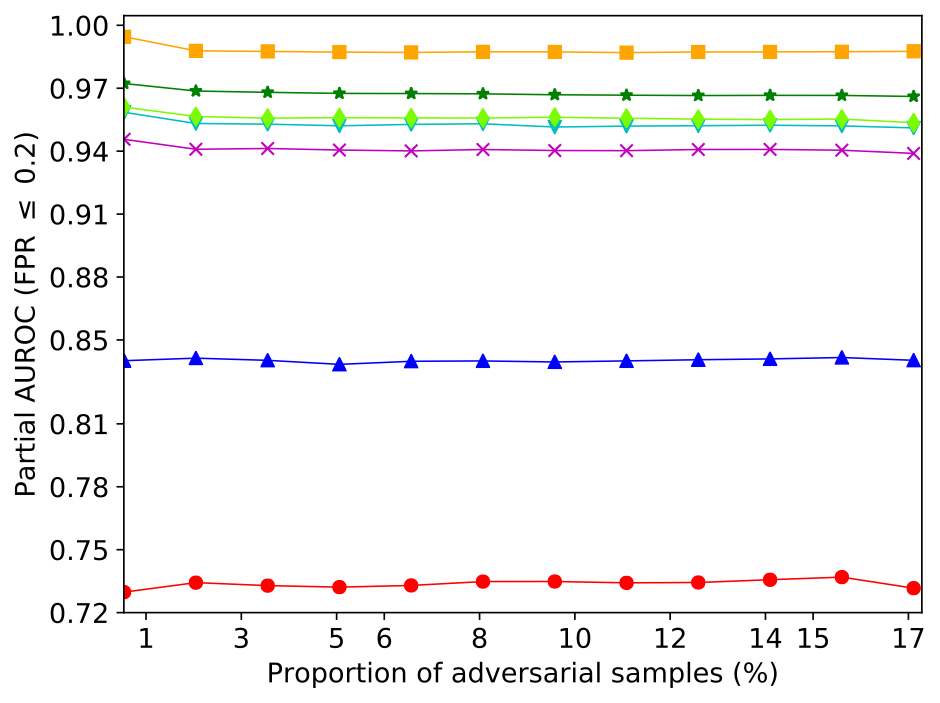} }
\caption{MNIST experiments: pAUC-$0.2$ for different attacks.}
\label{fig:mnist_pauc}
\end{figure*}

On the MNIST dataset (Figure~\ref{fig:mnist_pauc}), we observe some different trends in the performance compared to the other datasets. On the CW and FGSM attacks, the methods \odds and \trust perform comparably or slightly better than \proposed (particularly the variant based on Fisher's method). On the adaptive attack, the performance of \proposed based on Fisher's method decreases significantly as the perturbation norm increases. On the other hand, the variant of \proposed based on the aK-LPE method outperforms the other methods on this attack. We think that this contrast in performance is due to the fact that the adaptive attack samples were optimized to fool the variant of \proposed based on Fisher's method. Also, attack samples with higher perturbation norm are more likely to be successful. On the PGD attack, \odds outperforms the other methods, but the pAUC-$0.2$ of all methods, except \lid and \dknn, are higher than $0.95$ in this case. We conjecture that the good performance of most methods on MNIST could be due to the simplicity of the input space and the classification problem.

\subsection{Results on the MNIST Dataset}
\label{app:mnist_prec}
\begin{figure*}[htb]
\centering
\includegraphics[width=0.8\linewidth]{Figures/legend_adversarial.png}
\vspace{-2mm}
\centering
\subfloat[{{\small CW, confidence $= 0$}}]{\label{fig:avg_prec_mnist_cw}{\includegraphics[width=0.30\linewidth]{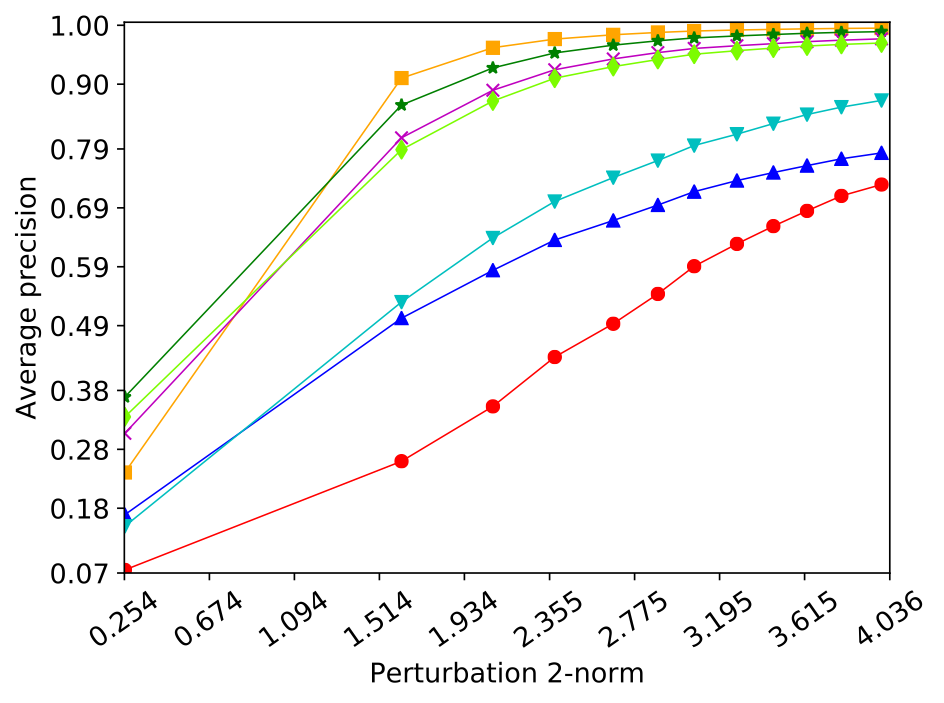}} }
\subfloat[{{\small Adaptive attack}}]{\label{fig:avg_prec_mnist_Custom}{\includegraphics[width=0.30\linewidth]{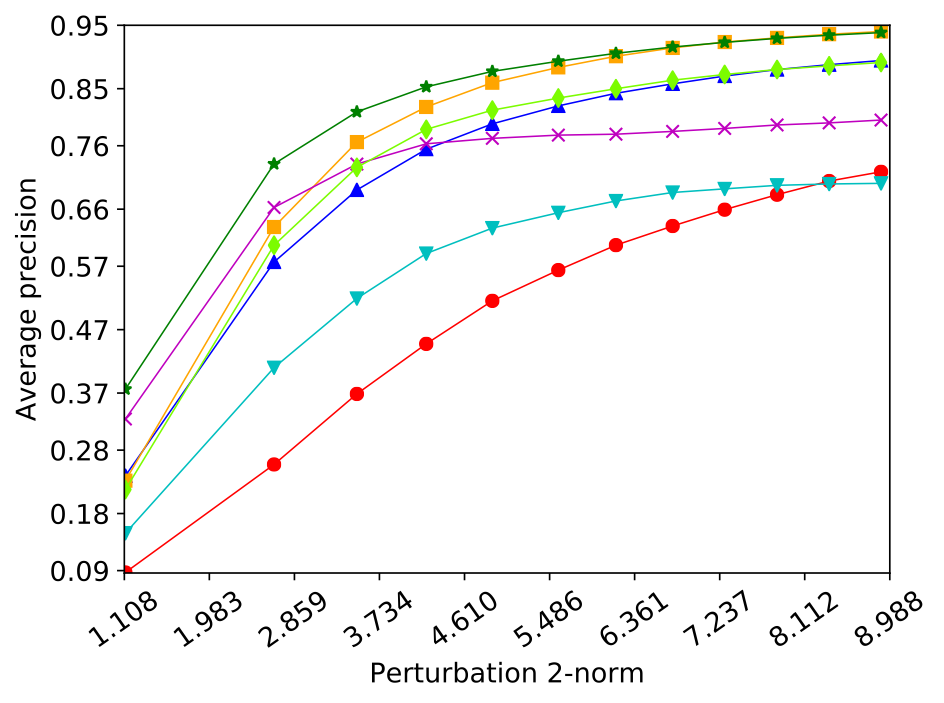}} }
\subfloat[{{\small PGD, $\epsilon = 1/255$}}]{\label{fig:avg_prec_mnist_pgd}{\includegraphics[width=0.30\linewidth]{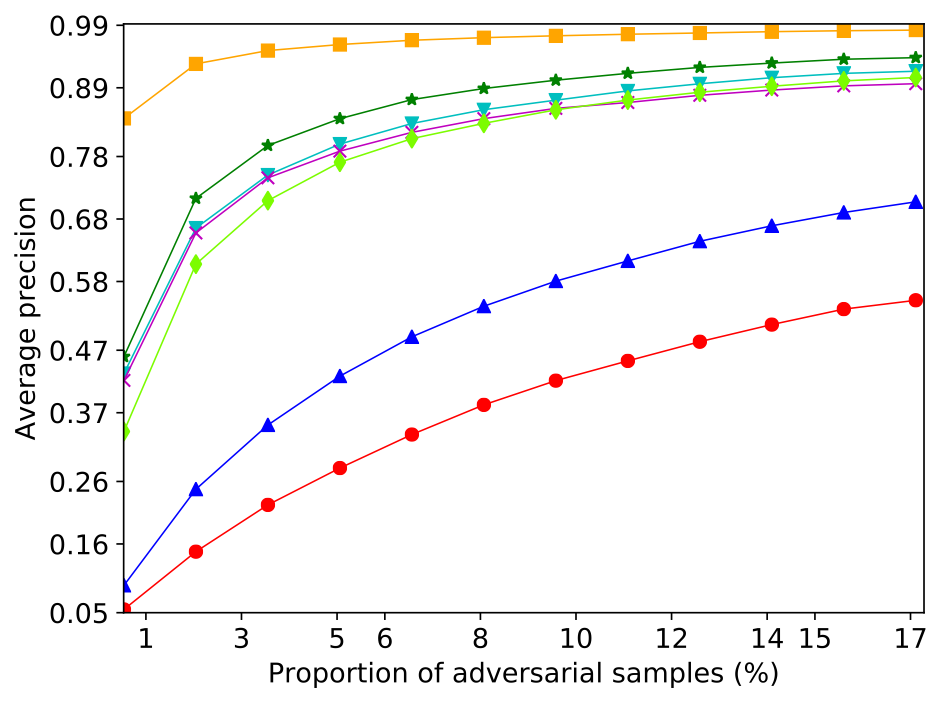}} }
\caption{MNIST experiments: average precision for different attacks.} 
\label{fig:mnist_prec}
\end{figure*}

In Figure~\ref{fig:mnist_prec}, we compare the average precision of different methods on the MNIST dataset for the CW, PGD, and adaptive attacks (results for the FGSM attack were presented in Appendix~\ref{app:fgsm}). We observe that \odds has good performance on this dataset, outperforming \proposed in some cases. The method \trust (which uses the pre-logit, fully connected layer) also performs well on this dataset. This could be due to the fact that on the MNIST dataset, the attack samples exhibit very distinctive patterns at the logit and pre-logit DNN layers, which are the focus of the methods \odds and \trust. We note that \odds and \trust do not carry over this good performance to all datasets and attacks. Also, both variants of \proposed perform well in the low perturbation norm regime.

\subsection{Running Time Comparison}
\label{sec:app_running_time}
\begin{table*}[htb]
\captionsetup{font=small,skip=10pt}
\centering
\caption{Average wall-clock running time per-fold (in minutes) for the different detection methods.}
\vspace{-1mm}
\begin{tabular}{@{}llllllll@{}}
\toprule
Dataset & \proposed, Fisher & \proposed, LPE & \dm & \odds & \lid & \dknn & \trust \\ 
\midrule
CIFAR-10 & 2.73 & 2.18 & 15.08 & 142.94 & 49.74 & 11.99 & 0.53 \\
SVHN & 6.37 & 5.01 & 4.19 & 33.60 & 100.80 & 23.54 & 0.60 \\
MNIST & 0.92 & 0.85 & 1.18 & 6.79 & 6.96 & 1.73 & 0.24 \\ 
\bottomrule
\end{tabular}%
\vspace{-2mm}
\label{tab:running_time}
\end{table*}
Table~\ref{tab:running_time} reports the wall-clock running time (in minutes) of the different detection methods per-fold, averaged across all attack methods. \trust consistently has the least running time, while both variants of \proposed have low running time as well. The running time of \dm is comparable to \proposed on MNIST and SVHN, but is higher on CIFAR-10. This is because \dm performs a search for the best noise parameter at each layer using 5-fold cross-validation, which takes a longer time on the Resnet-34 DNN for CIFAR-10. \odds and \lid have much higher running time compared to the other methods. For each test sample, \odds computes an expectation over noisy inputs (from a few different noise parameters), which increases its running time as the size of the DNN increases. The computation involved in estimating the LID features at the DNN layers increases with the sample size used for estimation and the number of layers.

\end{document}